\documentclass[runningheads]{llncs}
\PassOptionsToPackage{table}{xcolor}

\usepackage{eccv}

\usepackage{eccvabbrv}

\usepackage{graphicx}
\usepackage{booktabs}

\usepackage{verbatim}
\makeatletter
\let\c@rownum\undefined  \let\therownum\undefined
\let\c@colnum\undefined  \let\thecolnum\undefined
\let\c@rowcount\undefined \let\therowcount\undefined
\let\c@colcount\undefined \let\thecolcount\undefined
\makeatother
\usepackage{tabularray-2021}
\usepackage{array}
\usepackage{multirow}
\usepackage{ragged2e}
\usepackage{wrapfig}
\usepackage{adjustbox}
\usepackage{enumitem}
\usepackage{tikz}
\usetikzlibrary{arrows.meta}
\usepackage{algorithm}
\usepackage{algorithmic}

\usepackage{fontawesome}
\usepackage{colortbl}
\usepackage{fancyvrb}
\usepackage{tcolorbox}
\tcbuselibrary{skins,breakable}

\newenvironment{codebox}[1]{%
	\VerbatimEnvironment
	\begin{tcolorbox}[enhanced, breakable, colback=#1!15, colframe=#1!60,
		boxrule=0.8pt, arc=5pt, left=2mm, right=2mm, top=1mm, bottom=1mm]%
	\begin{Verbatim}[fontsize=\small]%
}{%
	\end{Verbatim}%
	\end{tcolorbox}%
}

\definecolor{DoublePearlLusta}{rgb}{0.988,0.898,0.803}
\definecolor{LinkWater}{rgb}{0.811,0.886,0.952}
\definecolor{Corvette}{rgb}{0.976,0.796,0.611}
\definecolor{Alto}{rgb}{0.85,0.85,0.85}

\newcommand{\red}[1]{{\color{red}#1}}

\newcommand{\subheader}[1]{\vspace{0.25\baselineskip} \noindent \textbf{#1.}}

\newcommand{\fkgprop}[1]{\texttt{#1}}

\newcommand{\xmark}{\text{\sffamily X}}

\newcommand{\mediumcaption}{\fontsize{8.75pt}{11pt}\selectfont}
\newcommand{\smallcaption}{\fontsize{8.5pt}{10.5pt}\selectfont}
\newcommand{\smaller}{\fontsize{8pt}{9pt}\selectfont}
\newcommand{\smallerr}{\fontsize{6.5pt}{7.6pt}\selectfont}

\newcommand{\pullup}{\vspace{-0.2\baselineskip}}
\newcommand{\pullupp}{\vspace{-0.4\baselineskip}}
\newcommand{\pulluppp}{\vspace{-0.8\baselineskip}}
\newcommand{\pullupppp}{\vspace{-1.6\baselineskip}}

\usepackage[accsupp]{axessibility}  % Improves PDF readability for those with disabilities.

\usepackage{hyperref}

\usepackage{orcidlink}

\begin{document}

% ---------------------------------------------------------------
\title{Trustworthy Image Authentication using Forensic Knowledge Graphs}

\titlerunning{Forensic Knowledge Graphs}

\author{Tai D. Nguyen\inst{1} \and
Matthew C. Stamm\inst{1}}

\authorrunning{T.~Nguyen et al.}

\institute{Drexel University, Philadelphia, PA 19104, USA \\}
% Institution 2 \and
% Institution 3}

\makeatletter
\let\@oldmaketitle\@maketitle
\renewcommand{\@maketitle}{%
	\@oldmaketitle
	\centering
		\includegraphics[width=0.9\linewidth]{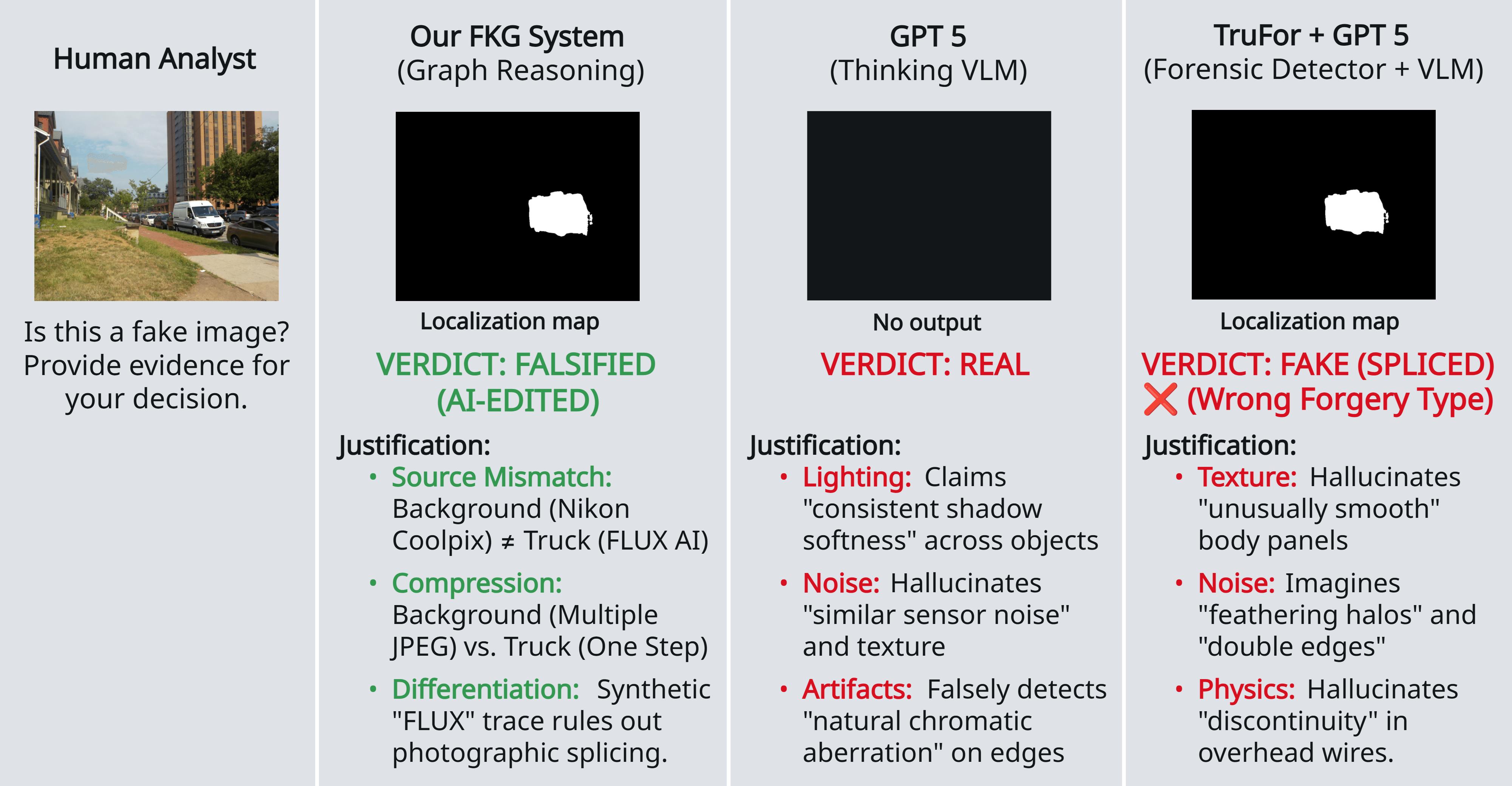}\par
		% \vspace{0.1em}
		\justifying\mediumcaption
		\noindent\textbf{Fig. 1:} Unlike existing systems that miss forgeries or hallucinate evidence, our FKG framework delivers trustworthy authentication grounded in real forensic traces.
	\vspace{-1.2em}
}
\makeatother

\maketitle

\setcounter{figure}{0}
\refstepcounter{figure}\label{fig:teaser}

\begin{abstract}
Advances in generative AI have made image falsification highly realistic, demanding trustworthy authentication systems. Existing forensic detectors can target certain forgery types but lack interpretability, while vision-language models (VLMs) provide explanations but cannot exploit forensic traces for reliable detection. We propose Forensic Knowledge Graphs (FKGs), a unified framework that integrates forensic evidence extraction, structured reasoning, and human-interpretable explanation. Our FKG structure encodes forensic traces along with their causal dependencies and links to scene content. To generate accurate FKGs, we introduce a novel forensic authentication network and an Iterative Context Refinement strategy that guides VLMs to produce faithful, grounded explanations. We also present FKG-50K, a dataset of 50,000 realistic forgeries with ground-truth FKGs. Experiments demonstrate that FKG outperforms both forensic detectors and VLMs in detection, forgery identification and localization, and forensic justification.
\pullupp
\keywords{Image Forensics \and Knowledge Graphs \and Trustworthy AI}
\end{abstract}
\pullupp

\vspace{-2.0em}
\section{Introduction}
\label{sec:intro}
\vspace{-0.8em}

Recent advances in generative AI~\cite{GANs,rombach2022high} and modern AI image editors~\cite{brooks2023instructpix2pix,flux_kontex,gptimage1,qwenimage2025,gemini2025} have dramatically expanded the ways images can be falsified. Hyper-realistic forgeries can now be produced through fully AI-generated imagery~\cite{dalle2,sdxl,stylegan2}, content splicing, localized editing, and AI-based modifications~\cite{verdoliva2020media}. As these capabilities improve, distinguishing authentic images from falsified ones has become increasingly challenging, even for expert analysts~\cite{nightingale2017can,nightingale2022ai,groh2022deepfake}.

To protect against these threats, there is a significant need for media authentication systems~\cite{verdoliva2020media}. However, for these systems to be effective in practice, users must be able to trust them~\cite{li2023trustworthy}. Specifically, users require interpretable and explainable outputs to establish trust in automated decision-making systems~\cite{doshivelez2017towards,gunning2019darpa,arrieta2020explainable}, no matter how accurate they are.
Therefore, a trustworthy authentication system must: (1) provide \textbf{accurate, general-purpose detection} regardless of forgery type, (2) produce \textbf{human-interpretable explanations} describing what was falsified and how, and (3) provide \textbf{justification of forensic reasoning} that grounds every claim in verifiable evidence.

Current authentication systems only partially satisfy these requirements. While numerous specialized systems have been developed, such as synthetic image detectors~\cite{CNNDet,NPR}, manipulation classifiers~\cite{MISLnet,guo2023hifi}, and splicing localizers~\cite{MVSS-Net,TruFor}, each method only targets a narrow family of forgeries and fails to generalize beyond it. 
For example, a splicing detector may achieve high accuracy on spliced images but miss AI-generated edits entirely~\cite{mareen2024tgif,amerini2024deepfake}, while a synthetic image detector may flag fully generated content but overlook traditional manipulations~\cite{mareen2024tgif,corvi2023detection}.
% For example, splicing detectors miss AI-generated edits while synthetic image detectors overlook traditional manipulations~\cite{mareen2024tgif,amerini2024deepfake,corvi2023detection}. 
At the moment, no single system can accurately detect a broad spectrum of forgery types~\cite{verdoliva2020media,roessler2019faceforensics}. Furthermore, their outputs are often confined to binary ``real/fake'' labels or heatmaps of suspicious areas~\cite{TruFor,MVSS-Net}, with no additional explanation of what content is falsified, how it was manipulated, or why the system believes a particular region is inauthentic~\cite{xu2025fakeshield,liu2024forgerygpt}.

At the opposite extreme, vision-language models (VLMs)~\cite{openai2023gpt4,geminiteam2023gemini,liu2023llava} can produce natural language justifications, but utilize visual or semantic inconsistencies~\cite{wang2025forensicsbench,jia2024chatgpt} such as unnatural lighting, cues rapidly disappearing as AI generators produce increasingly realistic outputs~\cite{zheng2024breakingsemantic,yan2025sanitycheck}. More critically, VLMs cannot leverage the invisible statistical traces (\ie, forensic microstructures such as sensor noise, compression artifacts, and demosaicing residuals) that reliable forensic techniques rely on~\cite{stamm2010forensic,Noiseprint}. Recent benchmarks confirm that even the best VLMs barely exceed chance-level accuracy on forgery detection tasks~\cite{wang2025forensicsbench,tariq2025llms,jia2024chatgpt}, and in real-world incidents, chatbots have confidently declared AI-generated images authentic~\cite{fullfact2025grok,edwards2025ai}, eroding public trust in digital media and AI verification~\cite{europol2022facing}.
% Citation notes for the last two sentences of this paragraph:
% "barely exceed chance-level accuracy":
%   - Wang et al., "Forensics-Bench: A Comprehensive Forgery Detection Benchmark Suite for Large VLMs" (CVPR 2025) — https://arxiv.org/abs/2503.15024
%   - Tariq et al., "LLMs Are Not Yet Ready for Deepfake Image Detection" (2025) — https://arxiv.org/abs/2506.10474
%   - Jia et al., "Can ChatGPT Detect DeepFakes?" (CVPR 2024 Workshop) — https://arxiv.org/abs/2403.14077
% "declared AI-generated images to be authentic":
%   - Full Fact, "Grok and Google Lens AI overviews claim fake imagery shows Huntingdon train attack" (Nov 2025) — https://fullfact.org/crime/grok-google-lens-ai-imagery-train-attack/
%   - Edwards et al. (Reuters Institute/WITNESS), "AI is undermining OSINT's core assumptions" (Dec 2025) — https://reutersinstitute.politics.ox.ac.uk/news/ai-undermining-osints-core-assumptions-heres-how-journalists-should-adapt
% "eroding public trust":
%   - Bloomberg, "Deepfakes and Chatbots Have Web Users Struggling to Prove Their Humanity" (2025) — https://www.bloomberg.com/features/2025-ai-deepfakes-chatbots-human/
%   - Europol Innovation Lab, "Facing Reality? Law Enforcement and the Challenge of Deepfakes" (2024)
%   - Springer AI & Society, "Deciphering authenticity in the age of AI" (2025) — https://link.springer.com/article/10.1007/s00146-025-02416-5

% These limitations create a critical need for systems that unify the reliability of forensic analysis with the interpretability and communicative power of VLMs.

To address these challenges, we introduce Forensic Knowledge Graphs (FKGs), a unified framework for trustworthy image authentication that integrates forensic evidence, structured reasoning, and interpretable explanation within a single system. FKGs capture forensic analyses as a structured graph representation, linking image regions with their sources, manipulation histories, and relevant scene content. This representation allows the system to analyze multiple forgery types simultaneously, preserve dependencies among forensic traces, link evidence to scene regions, and generate grounded natural-language explanations. We perform comprehensive experimental analysis of our framework using multiple datasets, demonstrating that FKG outperforms both existing forensic detectors and VLMs in detection, forgery identification and localization, and forensic justification.
\noindent Our contributions are summarized as follows:
\begin{enumerate}[topsep=0pt, itemsep=1.5pt, parsep=0pt, partopsep=0pt, leftmargin=0.5cm]
    \item We introduce \emph{Forensic Knowledge Graphs}, a novel information structure that encodes forensic evidence, causal dependencies among analyses, and links findings to semantic scene content.
    \item We propose a \emph{unified forensic authentication network} composed of a novel self-supervised trace extraction backbone, a novel hybrid region proposal network, and a transformer-based reasoner that can detect multiple different forgery types and construct accurate FKGs from input images.
    \item We propose \emph{Iterative Context Refinement (ICR)}, a novel prompting strategy that enables vision-language models to generate faithful, forensic-evidence-grounded summaries and justifications from FKGs.
    \item We release \emph{FKG-50K}, a dataset of 50,000 images with ground-truth FKGs covering multiple forgery types, and demonstrate state-of-the-art performance across all three dimensions of trustworthy authentication.
\end{enumerate}

\vspace{-1.25em}
\section{Background \& Related Work}
\label{sec:background}
\vspace{-0.5em}

\subheader{Forensic Microstructures}
These are subtle, content-independent pixel correlations imprinted by an image's acquisition and processing history~\cite{stamm2010forensic,stamm2013information}. They arise from sensor noise~\cite{lukas2006digital}, demosaicing~\cite{zhao2016computationally}, and compression~\cite{pevny2008detection,stamm2011antiforensics}, collectively forming a statistical ``fingerprint'' unique to each imaging pipeline~\cite{FSM}. % Unlike semantic or visual cues, these traces persist across diverse content and reveal how an image was formed, compressed, or manipulated. By modeling these microstructures, forensic systems can infer an image's origin~\cite{bayar2016deep,MISLnet}, detect inconsistencies among regions~\cite{Noiseprint,MVSS-Net}, and distinguish camera artifacts from AI-generated ones~\cite{corvi2023detection,CNNDet}.
Unlike semantic cues, these traces persist across diverse content, enabling forensic systems to infer an image's origin~\cite{bayar2016deep,MISLnet}, detect inconsistencies among regions~\cite{Noiseprint,MVSS-Net}, and distinguish camera artifacts from AI-generated ones~\cite{corvi2023detection,CNNDet}.

\subheader{VLMs for Image Authentication}
As AI-generated imagery proliferates online~\cite{europol2022facing}, users and journalists increasingly turn to VLMs such as ChatGPT and Gemini to verify image authenticity~\cite{caulfield2025chatgpt,edwards2025ai}. However, as discussed in \cref{sec:intro}, even the best VLMs barely exceed chance-level accuracy on forensic tasks~\cite{wang2025forensicsbench,jia2024chatgpt,tariq2025llms}, and fact-checking investigations confirm that VLMs and detection tools fail under simple perturbations like cropping or rescaling~\cite{anlen2024spotting,edwards2025ai}. Even forensic VLMs like FakeShield~\cite{xu2025fakeshield}, ForgeryGPT~\cite{liu2024forgerygpt}, SIDA~\cite{huang2025sida}, and recent reasoning-oriented detectors~\cite{cao2025reveal,tan2025forenx,bagaria2025insight} rely on visible cues rather than forensic microstructures, and cover only narrow forgery families (ForgeryGPT handles only faces). %These findings underscore the gap between public expectations and actual VLM authentication capabilities.

\subheader{Forensic Systems}
Modern forensic systems exploit these microstructures but vary in scope and design~\cite{verdoliva2020media}. Traditional detectors like CAT-Net~\cite{kwon2022catnet}, HiFi-IFDL~\cite{guo2023hifi}, and TruFor~\cite{TruFor} focus on specific manipulation families such as splicing and compression inconsistencies, while newer approaches target AI-generated content by modeling generator-specific artifacts (\eg, NPR~\cite{NPR}, DE-FAKE~\cite{DE-Fake}, UFD~\cite{UFD}, FSD~\cite{nguyen2025fsd}). However, these systems operate as binary detectors that cannot reason about causal relationships between multiple forensic cues~\cite{verdoliva2020media,amerini2024deepfake} struggling with forgeries that blend real and synthetic content~\cite{mareen2024tgif}.

% Our Forensic Knowledge Graph (FKG) framework addresses these limitations by integrating evidence from multiple forensic analyses into a unified structure that supports causal reasoning and interpretable decision-making.

\vspace{-1em}
\section{Forensic Knowledge Graphs}
\label{sec:fkg}
\vspace{-0.5em}

We introduce Forensic Knowledge Graphs (FKGs) - a hierarchical structure whose nodes represent forensic entities and edges encode labeled relations capturing how findings are interconnected and why they support a particular authenticity conclusion. The FKG's structure and semantics are defined by an \emph{ontology} that specifies entity classes and their permissible relationships~\cite{gruber1993translation}. An overview follows with formal definitions in Appendix A.

\begin{wrapfigure}{r}{4.6cm}
	\vspace{-2em}
	\centering
	% {\input{fig/fkg_example}}
    \includegraphics[width=4.6cm,height=6.0cm]{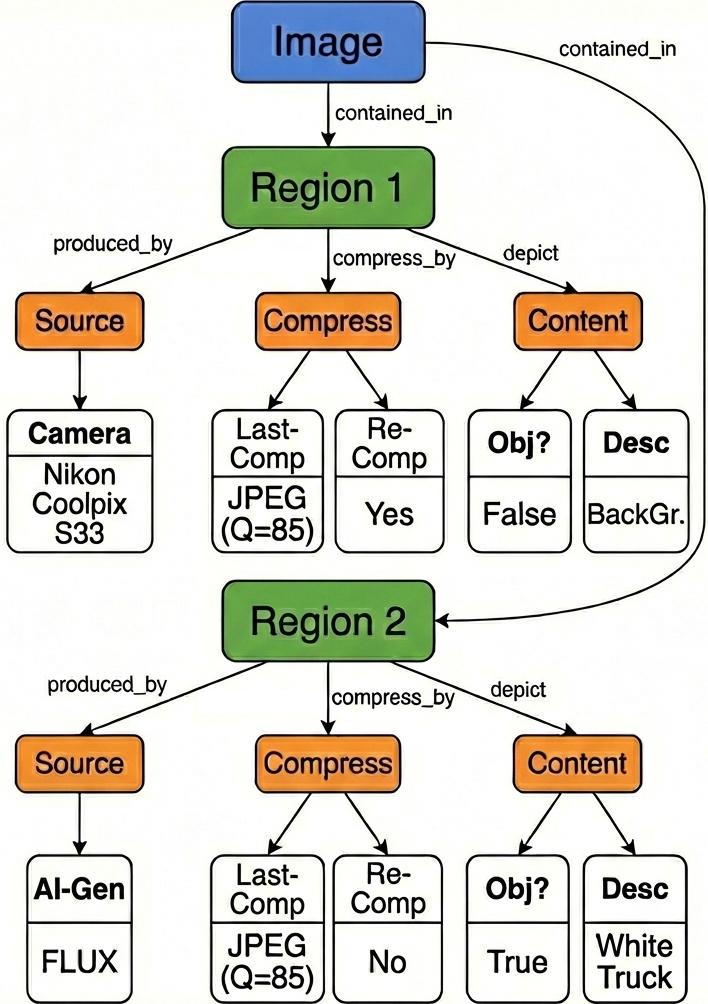}
    \vspace{-1.3em}
	\caption{\mediumcaption FKG for the AI-edited image in \cref{fig:teaser}.}
	\label{fig:fkg_example}
	\vspace{-2.1em}
\end{wrapfigure}

\subheader{Entity Classes and Relations}
The ontology defines six core entity classes and their typed relations: (1) \textbf{Image} - the root node representing the analyzed image; (2) \textbf{Region} - a pixel subset sharing common forensic traces, serving as the central entity from which all relations originate; (3) \textbf{Source} - the origin of pixels, with subclasses for \textit{Camera Model} and \textit{AI Generator}; (4) \textbf{Post-Processing} - operations applied after generation, \eg resampling, blur, noise; (5) \textbf{Compression} - compression artifacts, with subclasses tracking whether a region is uncompressed, its last compression parameters, and whether recompression occurred; and (6) \textbf{Content} - semantic scene elements characterized by objectedness and object description.

Relations encode dependencies among entity classes as ontology object properties. Each Region is linked to its parent Image via \fkgprop{contained\_in}, to its origin via \fkgprop{produced\_by}, to post-processing and compression nodes via \fkgprop{modified\_by} and \fkgprop{compress\_by}, and to semantic content via \fkgprop{depict}. Additional datatype properties (Appendix A) associate entities with descriptive literals like probabilities, model IDs, and post-processing parameters. These typed edges are instantiated deterministically from our system's per-region predictions, not inferred by a VLM, so every relation is grounded in an explicit forensic prediction.

\subheader{Illustrative FKG Example}
\cref{fig:fkg_example} shows the FKG for an AI-edited image where a truck was added to a real photo. The graph contains two Region nodes: one linked to Nikon Coolpix S33 and another to FLUX generator. Compression evidence reinforces this split: the camera region shows multiple recompressions while the FLUX region shows only one, indicating post-hoc content insertion. %Together, the FKG provides a clear reasoning path — a genuine Nikon photo was locally edited with FLUX to insert a truck — demonstrating how FKGs encode and explain forensic logic step by step.

\vspace{-1em}
\section{FKG Generation System}
\label{sec:fkg_sys}
\vspace{-1em}

\begin{wrapfigure}{r}{0.54\linewidth}
	% \pullupppp\pullupp
	\vspace{-2.4em}
	\centering
	\includegraphics[width=\linewidth]{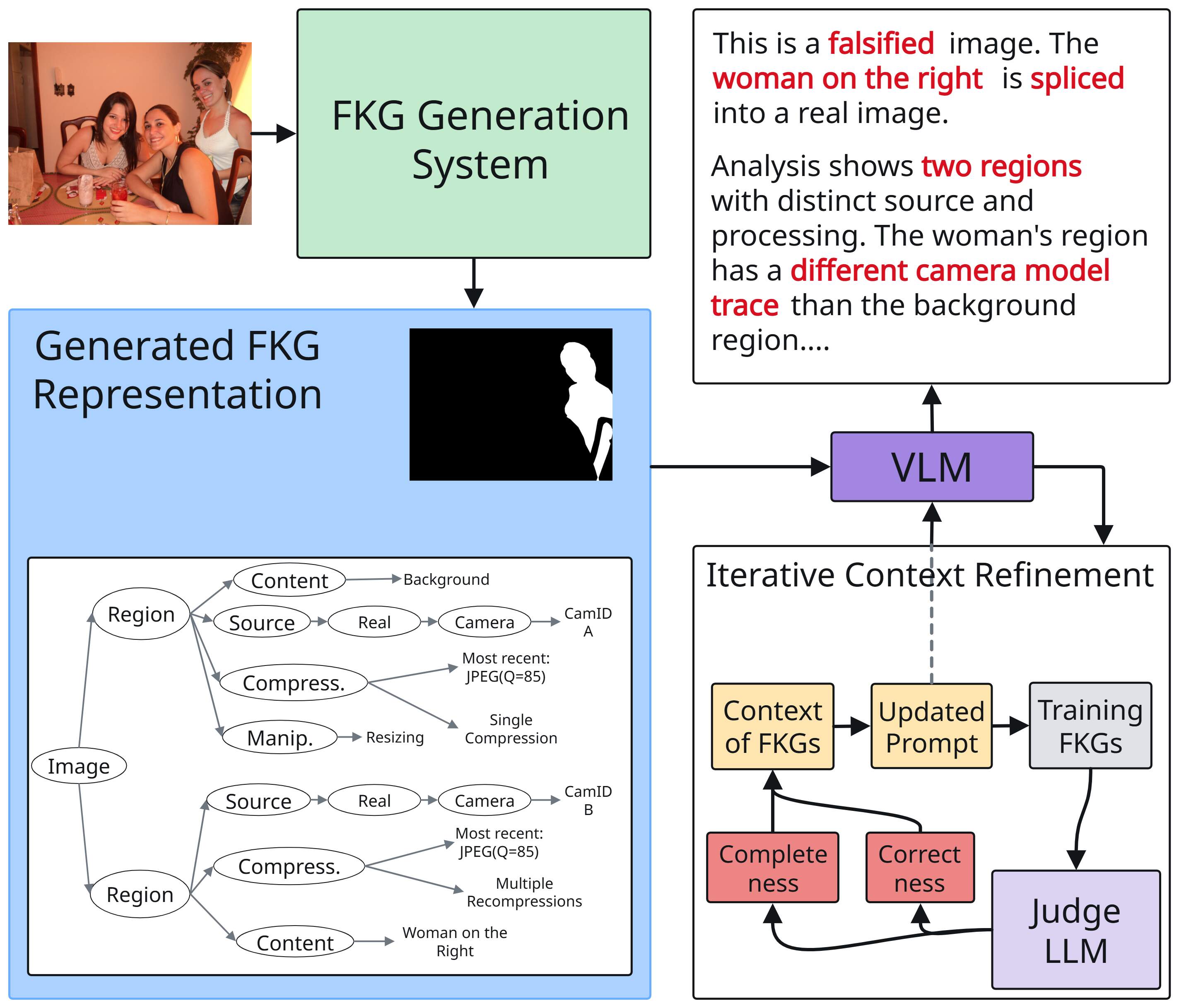}
	\vspace{-2em}
	\caption{Overview of our full FKG framework.}
	\label{fig:fkg_fw_overview}
	\pullupppp\pullupp
\end{wrapfigure}

\begin{wrapfigure}{r}{5.6cm}
	\vspace{-2.4em}
	\centering
	\includegraphics[width=5.6cm]{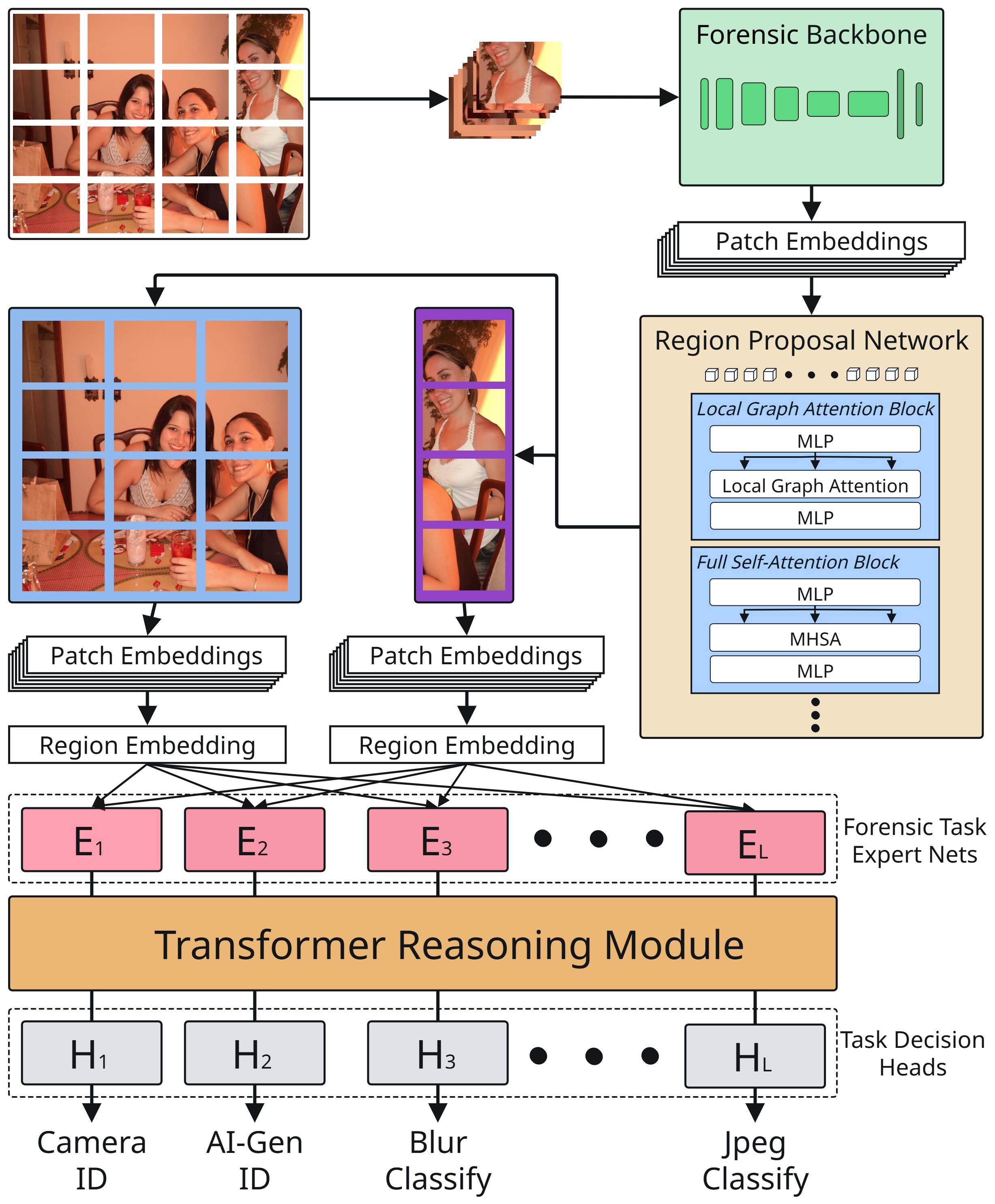}
	\vspace{-2.1em}
	\caption{Overview of our system which generates a FKG for an input image.}
	\label{fig:fkg_gen_overview}
	\vspace{-2.4em}
\end{wrapfigure}

\subheader{Overview}
%
% Our full framework (\cref{fig:fkg_fw_overview}) consists of three stages: an FKG Generation System, a VLM that translates the FKG into a forensic explanation, and an Iterative Context Refinement strategy (\cref{sec:interp_fkg}). The generation system itself operates in two stages (\cref{fig:fkg_gen_overview}): representation and inference. In the \emph{representation} stage, an image is divided into overlapping patches, each processed by a Forensic Backbone Network that extracts embeddings $z_i$, and a Forensic Region Proposal Network (FRPN) clusters patches sharing a common source into \emph{forensic regions}. In the \emph{inference} stage, Forensic Task Expert Networks analyze each region for source identity, post-processing, and compression, a Transformer Reasoning Module refines their outputs by modeling cross-task dependencies, and decision heads map the results to forensic attributes that form the final FKG. Full architectural and training details are in Appendix B–D.
Our framework (\cref{fig:fkg_fw_overview}) unifies forensic evidence extraction, structured reasoning, and interpretable explanation in a single pipeline. Given an input image, the FKG Generation System (\cref{fig:fkg_gen_overview}) divides it into patches and extracts forensic fingerprints via a Forensic Backbone Network. A Forensic Region Proposal Network (FRPN) then clusters patches with similar fingerprints into forensic regions. Finally, Task Expert Networks and a Transformer Reasoning Module infer each region's forensic attributes and map them to the FKG ontology. A VLM then translates the resulting FKG into a human-interpretable forensic explanation, guided by our Iterative Context Refinement strategy (\cref{sec:interp_fkg}). Full architectural and training details are in Appendix B–D.

\subheader{Forensic Backbone Network}
%
% Image forensic analysis relies on forensic microstructures, statistical fingerprints left by an image's source and processing history~\cite{stamm2010forensic,Noiseprint}. Prior methods learn these through task-specific supervision (\eg, camera model classification)~\cite{MISLnet}, limiting generalization~\cite{verdoliva2020media}, while standard self-supervised approaches~\cite{chen2020simclr} learn representations invariant to augmentations such as cropping and recompression that destroy the very traces we aim to capture.
Generating an FKG first requires a backbone network that captures forensic microstructures, statistical fingerprints left by an image's source and processing history~\cite{stamm2010forensic,Noiseprint}. These \emph{general-purpose} embeddings are the foundation of our system: the FRPN relies on them to partition the image into forensic regions, and the task experts rely on them to infer forensic attributes. Prior methods learn these through task-specific supervision (\eg, camera model classification)~\cite{bayar2016deep,MISLnet,kwon2022catnet,MVSS-Net,guo2023hifi}, tying the backbone to specific sources and limiting generalization~\cite{verdoliva2020media,mareen2024tgif,amerini2024deepfake}.

% We address this with a forensic-specific self-supervised pretext task grounded in a fundamental property of imaging pipelines: \textit{local patches from the same image share a common forensic fingerprint, while patches from different images do not}~\cite{FSM,Noiseprint}. This principle provides a scalable, \emph{label-free} training signal over \emph{large unlabeled image collections}.
Rather than learning from task-specific labels, we exploit a fundamental yet surprisingly simple property of imaging pipelines: \textit{local patches from the same image share a common forensic fingerprint, while patches from different images do not}. This observation alone turns every unlabeled image into a free training sample, enabling the backbone to learn general forensic fingerprints at scale without forgery labels or source annotations. To prevent the network from learning content-dependent shortcuts (\eg, mapping semantically similar patches to the same image), we first process each patch using a learned high-pass filter~\cite{bayar2016deep,MISLnet} to suppress image content. We then enforce self-supervision through two complementary objectives: a pairwise similarity term enforcing same-image alignment and cross-image separation, and a contrastive term sharpening per-patch discrimination. Given a set of patches $\mathcal{X}$, the backbone is trained with:\pullupp
\begin{equation}
	\pullupp
	\mathcal{L}_{B} = \frac{1}{|\mathcal{X}|^2}\sum_{i,j} w_{ij} (Y_{ij} - z_i^\top z_j)^2 - \lambda \sum_{i} \log \frac{e^{\cos(z_i,z_{i^+})/\tau}} {\sum_{j}e^{\cos(z_i,z_j)/\tau}},
\end{equation}
where $z_i$ are $\ell_2$-normalized patch embeddings, $Y_{ij} \in \{0,1\}$ is a same-image indicator, $z_{i^+}$ is the positive pair of $z_i$, $w_{ij}$ balance positive and negative pairs, and $\tau$ is a temperature. These embeddings encode each patch's forensic identity and serve as input to the FRPN.

\subheader{Forensic Region Proposal Network}
%
% Constructing an FKG requires partitioning the image into regions that share a common forensic trace.
Given the backbone's forensic embeddings, the next step is partitioning the image into regions that share a common forensic trace. Existing visual segmentation and region proposal methods~\cite{he2017mask,kirillov2023sam} cannot accomplish this because they operate on visual features, whereas forensic regions are defined by \emph{invisible} 
statistical properties: a spliced region may appear visually seamless yet carry an entirely different forensic fingerprint.

% To address this, we introduce a Forensic Region Proposal Network (FRPN) based on the following observation:
% \emph{partitioning an image into forensically homogeneous regions requires comparing every patch against every other - the all-pairs operation that self-attention naturally computes}. However, standard global self-
% attention suffers from \emph{attention dilution}~\cite{vaswani2017attention}: softmax normalization assigns every token a non-zero weight, and these small irrelevant contributions accumulate, diluting focus from forensically meaningful neighbors. To mitigate this, we propose a novel \emph{Hybrid Graph Attention Transformer} that takes the backbone embeddings $Z = [z_1, \ldots, z_{|\mathcal{P}|}]$ as input tokens and alternates global self-attention with local graph attention. Global self-attention captures long-range dependencies (\eg, two distant regions originating from the same camera), while local graph attention restricts aggregation to each token's $k$-nearest neighbors using an additive scoring mechanism~\cite{velickovic2018graph}.
To address this, we introduce a Forensic Region Proposal Network (FRPN) based on the following observation: \emph{partitioning an image into forensically homogeneous regions requires comparing every patch against every other — the all-pairs operation that transformer self-attention naturally computes}. However, global self-attention suffers from \emph{attention dilution}~\cite{vaswani2017attention}: accumulated contributions from distant, unrelated patches cause forensically distinct regions to appear connected, leading to false alarms and misclassified clusters. To mitigate this, we propose a novel \emph{Hybrid Graph Attention Transformer} that takes the backbone embeddings $Z = [z_1, \ldots, z_{|\mathcal{P}|}]$ as input tokens and alternates global self-attention with local graph attention. Local graph attention suppresses false alarms by restricting aggregation to each token's $k$-nearest neighbors using an additive scoring mechanism~\cite{velickovic2018graph}.
%
% Additive attention cite: Veličković et al., "Graph Attention Networks" (ICLR 2018) - https://arxiv.org/abs/1710.10903
The attention weights under each mode are:\vspace{-0.4em}
\begin{equation}
	% \fontsize{7pt}{8pt}\selectfont
	% \begin{split}
	\vspace{-0.4em}
	\smaller
	\alpha_{ij}^{\text{global}} = \frac{\exp(\mathbf{w}_{qi}^\top \mathbf{w}_{kj} / \sqrt{d})}{\sum_{m=1}^{|\mathcal{P}|} \exp(\mathbf{w}_{qi}^\top \mathbf{w}_{km} / \sqrt{d})}, \enspace
	\alpha_{ij}^{\text{local}} = \frac{\exp\!\big(\sigma(\mathbf{a}_s^\top \mathbf{w}_{gi} + \mathbf{a}_d^\top \mathbf{w}_{gj})\big)}{\sum_{m \in \mathcal{N}(i)} \exp\!\big(\sigma(\mathbf{a}_s^\top \mathbf{w}_{gi} + \mathbf{a}_d^\top \mathbf{w}_{gm})\big)},
	% \end{split}
\end{equation}
\noindent
where $\mathbf{w}_{ni} = W_n z_i \in \mathbb{R}^d$ are learned projections of token $i$, $\sigma$ is the LeakyReLU activation, $\mathbf{a}_s, \mathbf{a}_d$ are learned score vectors, and $\mathcal{N}(i)$ are the dynamic $k$-nearest neighbors of $i$ by cosine similarity. Stacking $L$ alternating layers yields:
% \begin{equation}
	$\tilde{Z} = \text{Attn}_{g}^{(L)} \circ \text{Attn}_{\ell}^{(L)} \circ \cdots \circ \text{Attn}_{g}^{(1)} \circ \text{Attn}_{\ell}^{(1)}(Z),$
% \end{equation}
where $\text{Attn}_{g}$ and $\text{Attn}_{\ell}$ denote global self-attention and local graph attention layers. The output $\tilde{Z} = [\tilde{z}_1, \ldots, \tilde{z}_{|\mathcal{P}|}]$ are refined embeddings from which a classification head predicts a cluster assignment $\hat{c}_i$ for each patch.

% To train the FRPN on images containing multiple ground-truth forensic regions (\eg, manipulated images with known tampering masks), we combine a pairwise similarity term that preserves embedding structure with a Hungarian-matched cross-entropy term~\cite{carion2020detr} that enforces discrete cluster assignments:
To train cluster assignments against ground-truth forensic regions (\eg, from known tampering masks), we must account for the fact that predicted cluster IDs are arbitrary and do not correspond to ground-truth region labels. We use Hungarian matching~\cite{carion2020detr} to find the optimal assignment, and define a loss by combining it with a pairwise similarity term that preserves embedding structure:\vspace{-0.75em}
% Hungarian matching cite: Carion et al., "End-to-End Object Detection with Transformers" (ECCV 2020) — https://arxiv.org/abs/2005.12872
%
% \begin{equation}
% 	\pullupp
% 	\smaller
% 	\begin{split}
% 	&\mathcal{L}_{F} = \frac{1}{|\mathcal{P}|^2}\sum_{i,j} w_{ij}(Y_{ij} - \tilde{z}_i^\top \tilde{z}_j)^2 \\
% 	&+ \alpha \sum_{i=1}^{|\mathcal{P}|} \ell_{\text{CE}}\!\big(y_i,\; \hat{\sigma}(\hat{c}_i)\big),
% 	\end{split}
% \end{equation}
\begin{equation}
	\vspace{-0.75em}
	\mathcal{L}_{F} = \frac{1}{|\mathcal{P}|^2}\sum_{i,j} w_{ij}(Y_{ij} - \tilde{z}_i^\top \tilde{z}_j)^2 
	+ \alpha \sum_{i=1}^{|\mathcal{P}|} \ell_{\text{CE}}\!\big(y_i,\; \hat{\sigma}(\hat{c}_i)\big),
\end{equation}
where $\tilde{z}_i$ are $\ell_2$-normalized, $Y_{ij} = \mathbf{1}[y_i = y_j]$ is a same-region indicator, $\ell_{\text{CE}}$ is the cross-entropy loss, $\hat{\sigma}$ is the optimal Hungarian matching from predicted to ground-truth labels~\cite{carion2020detr}, and $\alpha$ balances the two terms. Patches sharing the same predicted cluster form a forensic region, serving as Region nodes in the FKG.

\subheader{Forensic Task and Decision Making}
Given the forensic regions produced by the FRPN, this stage determines \emph{what happened} to each region and assembles the final FKG. Each region's refined patch embeddings $\{\tilde{z}_i\}$ are average-pooled into a region-level vector $\psi_k$, then processed by Forensic Task Expert Networks - MLPs specializing in source identification, post-processing classification, and compression analysis - each producing a task-specific embedding $\xi_k^{(t)}$. A Transformer Reasoning Module refines these by modeling cross-task dependencies (\eg, compression artifacts informing source identification) to produce updated features $\tilde{\xi}_k^{(t)}$, from which task-specific heads predict binary, categorical, or continuous outputs under:
$\mathcal{L}_{T} = \sum_{t \in \mathcal{T}} \lambda_t \mathcal{L}_t,$
where $\mathcal{T}$ is the set of forensic tasks, $\lambda_t$ are task-specific weights, and $\mathcal{L}_t$ is cross-entropy or mean-squared error between predictions from $\tilde{\xi}_k^{(t)}$ and ground-truth labels.

These predictions are mapped to the FKG ontology: each region becomes a Region node whose predicted attributes instantiate entity nodes and edges via \fkgprop{produced\_by}, \fkgprop{modified\_by}, and \fkgprop{compress\_by} relations. When different regions link to different sources, the graph directly reveals manipulation. The backbone, FRPN, and task networks are trained sequentially (details in Appendix D).

\vspace{-0.8em}
\section{Interpreting and Reasoning over FKGs}
\label{sec:interp_fkg}
\vspace{-0.8em}

Given an FKG, we use a vision-language model (VLM) to produce forensic decisions from its region-level evidence, including image authenticity, manipulation localization, and whether content is spliced, AI-edited, or fully synthetic, along with justifications grounding each decision in FKG evidence.

This requires addressing two challenges. First, the FKG combines symbolic attributes with references to spatial visual content, making it inherently multimodal, and standard VLMs cannot consume graph structures directly. Second, even with a serialized FKG, current VLMs lack forensic domain knowledge and tend to fabricate or omit evidence, yielding untrustworthy justifications.

\begin{wrapfigure}{r}{5.2cm}
	\vspace{-2.45em}
	\centering
	\includegraphics[width=5.2cm,height=6.3cm]{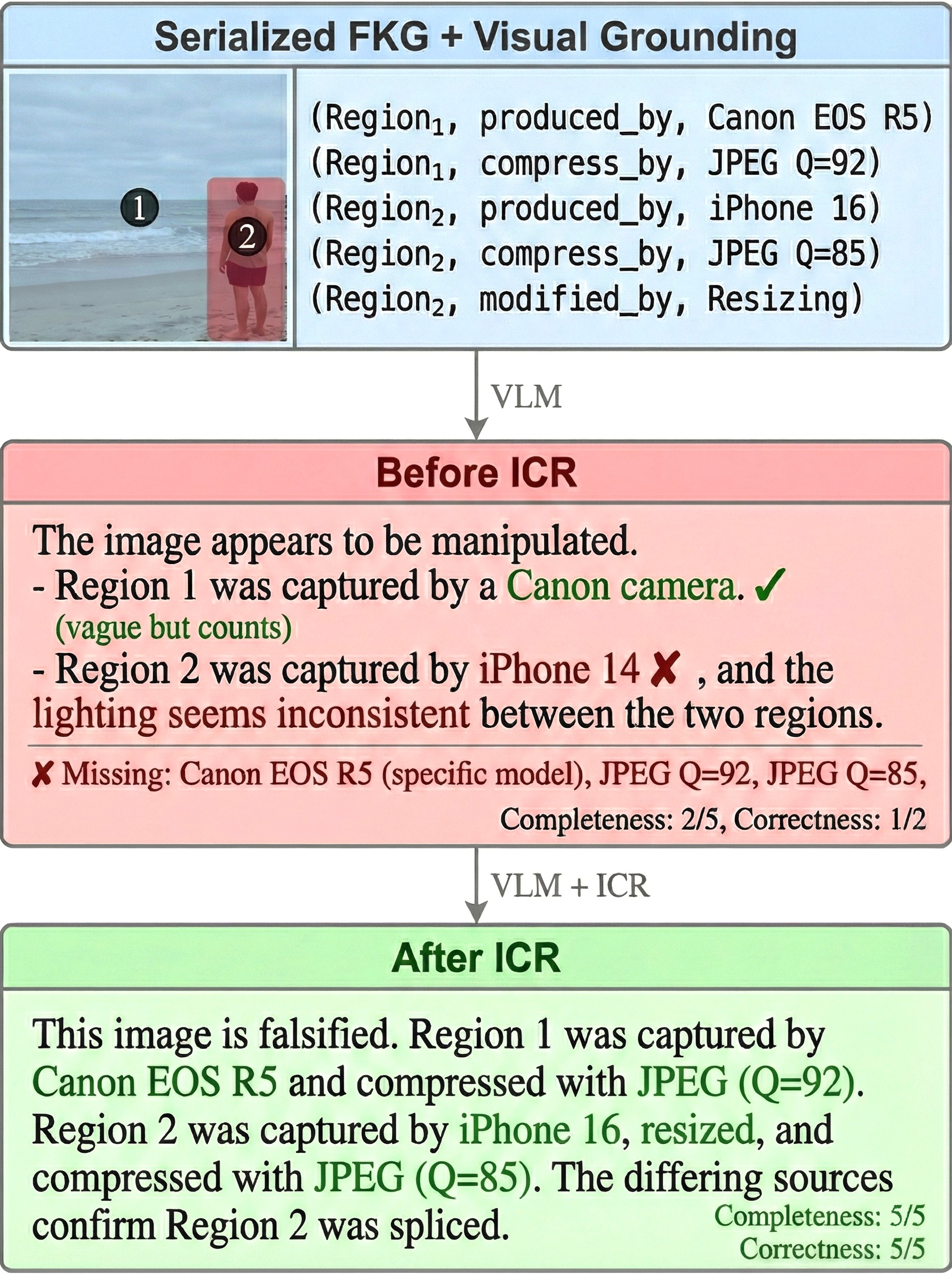}
	\vspace{-2em}
	\caption{ICR improves forensic justifications by refining VLM's context.}
	\label{fig:icr_example}
	\vspace{-2.4em}
\end{wrapfigure}

We serialize the FKG $F$ into subject-predicate-object triplets $F = \{(s_i, p_i, o_i)\}$. However, triplets alone cannot convey what each region depicts. We therefore use Set-of-Mark (SoM) prompting~\cite{yang2023setofmark}, overlaying each Region with a numbered mark on the image and referencing it symbolically in the triplets (\cref{fig:icr_example}, top). This visual grounding enables the VLM to recognize the scene content (\eg, a spliced object or background) and link it to forensic evidence. These, along with task instructions and a response schema (Appendix F), form the input prompt to the VLM.

\subheader{Iterative Context Refinement}
A naive approach to the second challenge is in-context learning~\cite{brown2020language}. However, manual curation of good FKG-report pairs is infeasible given FKG diversity, and existing prompt optimization methods~\cite{li2023unified,ye2023ceil,khattab2024dspy,pryzant2023protegi} optimize for producing correct answers (\eg, to a math problem), not for faithfulness to structured evidence. We therefore propose Iterative Context Refinement (ICR), a novel optimization strategy that automatically curates in-context examples to maximize report faithfulness and completeness. To measure these quantities, we treat each triplet $m_j \in F$ as an atomic fact and define two indicators: $\tau(m_j, R) \in \{0,1\}$ for whether $m_j$ is mentioned in report $R$, and $Q(m_j, R) \in \{0,1\}$ for whether it is correctly stated. Then, we define completeness and correctness, as:\vspace{-0.4em}
\begin{equation}
	\vspace{-0.4em}
	\mediumcaption
	\label{eq:metrics}
	M_{\text{comp}} = \frac{1}{|F|} \sum_{j} \tau(m_j, R), \enspace
	M_{\text{corr}} = \frac{1}{\sum_{j} \tau(m_j, R)} \sum_{j} Q(m_j, R)\tau(m_j, R).
\end{equation}

% DONE: Judge LM specification and convergence analysis in Appendix G.
The key idea behind ICR is to identify and correct the VLM's failure modes. Starting from $\mathcal{C} = \emptyset$ (\cref{fig:fkg_fw_overview}), a judge LM~\cite{zheng2023judging} evaluates $M_{\text{comp}}$ and $M_{\text{corr}}$ for each FKG in a training set of $N$ authentic and manipulated images. The FKGs where the VLM most severely fabricates or omits evidence are paired with their errors as corrective demonstrations and appended to $\mathcal{C}$, forming an updated context set $\mathcal{C}'$ that is prepended to the VLM's prompt in the next round. By focusing on the VLM's blind spots, ICR progressively drives faithfulness toward convergence:\vspace{-0.4em}
\begin{equation}
	\vspace{-0.6em}
	\frac{1}{N} \sum_{i}
	\big(
	| M_{\text{corr},i}^{(t)} - M_{\text{corr},i}^{(t-1)} |
	+
	| M_{\text{comp},i}^{(t)} - M_{\text{comp},i}^{(t-1)} |
	\big) < \epsilon.
\end{equation}
As shown in \cref{fig:icr_example} (bottom), this ensures every forensic claim is backed by verifiable graph evidence and every piece of evidence is accounted for.

% \vspace{-0.6em}
\section{Datasets}
\label{sec:datasets}
% \vspace{-0.6em}

Evaluating an image authentication system requires ground-truth data whose provenance is \textit{\textbf{fully known, accurately labeled, and completely captured}}. Existing forensic datasets provide at most binary real/fake labels or tampering masks, which is insufficient for evaluating systems that reason over structured forensic evidence such as source identity, compression history, and modification lineage. To address this, we curate \textbf{FKG-50K}, a dataset of 50,000 images with complete provenance and manipulation histories, built from a corpus of pristine, camera-original photographs sourced from Unsplash~\cite{unsplash}, Pexels~\cite{pexels}, and crowdsourcing. We filtered for images with rich EXIF metadata (camera model, processing parameters, color pipeline), as these define the source-level forensic fingerprint the FKG ontology captures. Details are provided in Appendix E.

From these originals, we generated \textit{realistic} manipulated counterparts spanning splicing, local retouching, and AI-based editing (FLUX-Kontex-1~\cite{flux_kontex}, SDXL-Inpaint~\cite{sdxl}), recording each manipulation as a structured operation encoded into its corresponding FKG. We also generated \textit{photo-realistic} fully synthetic counterparts from originals using SDXL~\cite{sdxl}, FLUX~\cite{flux}, Midjourney v6~\cite{Midjourney}, \& GPT-Image-1~\cite{gptimage1}. The dataset includes 40K training and 10K evaluation samples. We also benchmark on out-of-distribution public datasets such as DSO-1~\cite{decarvalho2013dso}, CASIAv2~\cite{dong2013casia}, GenImage~\cite{zhu2024genimage}, Synthbuster~\cite{Synthbuster}.

\section{Experiments}
\label{sec:experiments}

We evaluate our FKG framework \& others on three dimensions of trustworthy image authentication: (1) detecting falsified images, (2) identifying how \& where manipulations occur, (3) providing interpretable, evidence-based justifications.

\begin{table}[t]
	\centering
	\caption{\smallcaption Fake \& manipulated image detection on FKG-50K and other datasets. Each cell shows ACC / AUC. [I] = Instruct, [T] = Thinking.}
	\vspace{-0.5em}
	\resizebox{\linewidth}{!}{%
		\begin{tblr}{
				width=\linewidth,
				colspec={l *{5}{c} || *{4}{c}},
				hlines, vlines,
				row{2-3}={c},
				row{4-11}={bg=DoublePearlLusta},
				row{12-18}={bg=LinkWater},
				row{19}={bg=Corvette},
				hline{4,12,19}={1.5pt},
				stretch=0.4,
			}
			\multirow{3}{*}{\textbf{Method}}
			& \multicolumn{5}{c}{\multirow{2}{*}{\textbf{FKG-50K}}}
			& \multicolumn{4}{c}{\textbf{Out of Distribution Datasets}} \\

			& \multicolumn{5}{c}{}
			& \multicolumn{2}{c}{\textbf{Splice \& Edit}}
			& \multicolumn{2}{c}{\textbf{Fully AI-Gen}} \\

			& \textbf{AI-Edit}
			& \textbf{Full AI}
			& \textbf{Splicing}
			& \textbf{Trad-Edit}
			& \textbf{Overall}
			& \textbf{DSO-1}
			& \textbf{CASIAv2}
			& \textbf{GenImage}
			& \textbf{Synthbuster} \\

			GPT-5 & 0.59 / 0.60 & 0.68 / 0.82 & 0.81 / 0.84 & 0.55 / 0.61 & 0.66 / 0.72 & 0.59 / 0.61 & 0.74 / 0.75 & \textbf{0.93} / 0.88 & 0.90 / 0.86 \\
			Sonnet 4 & 0.55 / 0.52 & 0.72 / 0.75 & 0.54 / 0.53 & 0.51 / 0.51 & 0.58 / 0.58 & 0.50 / 0.53 & 0.56 / 0.62 & 0.85 / 0.66 & 0.91 / 0.78 \\
			Gemini 2.0 & 0.53 / 0.51 & 0.64 / 0.62 & 0.71 / 0.72 & 0.55 / 0.59 & 0.60 / 0.61 & 0.52 / 0.53 & 0.66 / 0.65 & 0.90 / 0.75 & 0.91 / 0.78 \\
			Qwen3 VL [I] & 0.53 / 0.53 & 0.72 / 0.71 & 0.63 / 0.63 & 0.54 / 0.54 & 0.60 / 0.60 & 0.50 / 0.51 & 0.59 / 0.59 & 0.89 / 0.73 & 0.92 / 0.81 \\
			Qwen3 VL [T] & 0.57 / 0.57 & 0.69 / 0.80 & 0.68 / 0.72 & 0.54 / 0.53 & 0.62 / 0.66 & 0.52 / 0.53 & 0.57 / 0.61 & 0.61 / 0.84 & 0.60 / 0.84 \\
			Gemma 3 & 0.56 / 0.58 & 0.69 / 0.76 & 0.65 / 0.68 & 0.50 / 0.52 & 0.60 / 0.64 & 0.57 / 0.57 & 0.52 / 0.55 & 0.69 / 0.75 & 0.69 / 0.79 \\
			Llama 4 S. & 0.52 / 0.52 & 0.64 / 0.65 & 0.52 / 0.50 & 0.50 / 0.52 & 0.54 / 0.55 & 0.48 / 0.51 & 0.56 / 0.57 & 0.73 / 0.68 & 0.71 / 0.58 \\
			Nemotron N. & 0.52 / 0.52 & 0.48 / 0.48 & 0.58 / 0.56 & 0.53 / 0.50 & 0.53 / 0.51 & 0.37 / 0.48 & 0.45 / 0.53 & 0.72 / 0.61 & 0.64 / 0.59 \\

			TruFor & 0.53 / 0.76 & 0.74 / 0.81 & 0.88 / 0.96 & 0.65 / 0.72 & 0.70 / 0.81 & 0.94 / 0.99 & 0.89 / 0.95 & 0.84 / 0.82 & 0.81 / 0.73 \\
			MVSS-Net & 0.48 / 0.48 & 0.47 / 0.46 & 0.56 / 0.65 & 0.53 / 0.55 & 0.51 / 0.54 & 0.46 / 0.47 & \textbf{0.88} / \textbf{0.97} & 0.26 / 0.33 & 0.29 / 0.45 \\
			CAT-Net & 0.57 / 0.52 & 0.46 / 0.37 & 0.85 / 0.91 & 0.85 / 0.91 & 0.68 / 0.68 & \textbf{0.99} / \textbf{1.00} & 0.72 / 0.83 & 0.76 / 0.49 & 0.73 / 0.41 \\
			HiFi-IFDL & 0.50 / 0.50 & 0.47 / 0.35 & 0.51 / 0.52 & 0.51 / 0.53 & 0.49 / 0.47 & 0.50 / 0.51 & 0.50 / 0.55 & 0.21 / 0.33 & 0.21 / 0.54 \\
			NPR & 0.50 / 0.51 & 0.90 / 0.98 & 0.50 / 0.50 & 0.52 / 0.52 & 0.60 / 0.63 & 0.50 / 0.49 & 0.51 / 0.52 & 0.92 / \textbf{0.97} & \textbf{0.94} / \textbf{0.98} \\
			UFD & 0.51 / 0.54 & 0.71 / 0.73 & 0.56 / 0.65 & 0.51 / 0.53 & 0.57 / 0.61 & 0.47 / 0.48 & 0.49 / 0.50 & 0.79 / 0.74 & 0.79 / 0.57 \\
			DE-FAKE & 0.51 / 0.52 & 0.88 / 0.95 & 0.51 / 0.53 & 0.51 / 0.57 & 0.60 / 0.64 & 0.52 / 0.56 & 0.49 / 0.55 & 0.86 / 0.98 & 0.83 / 0.92 \\

			\textbf{Ours} & \textbf{0.86} / \textbf{0.88} & \textbf{0.93} / \textbf{0.94} & \textbf{0.95} / \textbf{0.99} & \textbf{0.93} / \textbf{0.97} & \textbf{0.92} / \textbf{0.94} & 0.92 / 0.96 & 0.85 / 0.93 & 0.92 / 0.96 & \textbf{0.94} / 0.96 \\
	\end{tblr}}
	\label{tab:exp1}
	\vspace{-1.7em}
\end{table}

\subsection{Fake \& Manipulated Image Detection}
\label{subsec:exp1}

\subheader{Setup}
We benchmark our FKG system and competing approaches, including both forensic systems and modern vision-language models (VLMs), on our FKG-50K test set and four public datasets: DSO-1~\cite{decarvalho2013dso}, CASIA-v2~\cite{dong2013casia}, GenImage~\cite{zhu2024genimage}, and Synthbuster~\cite{Synthbuster}. We report overall and category-wise performance across forgery types: splicing, traditional editing, AI-editing, and fully AI-generated. VLM prompt templates and protocols are detailed in the appendix.

\subheader{Metrics}
We report detection accuracy and area under the ROC curve (AUC).

\subheader{Competing Methods}
We compare against forensic systems purpose-built for forgery detection (TruFor~\cite{TruFor}, MVSS-Net~\cite{MVSS-Net}, CAT-Net~\cite{kwon2022catnet}, HiFi-IFDL~\cite{guo2023hifi} for traditional manipulations; NPR~\cite{NPR}, UFD~\cite{UFD}, DE-FAKE~\cite{DE-Fake} for AI-generated detection) and powerful SOTA vision-language models (VLMs) including GPT-5~\cite{openai2025gpt5}, Sonnet 4~\cite{anthropic2025claude4}, Gemini 2.0 Flash~\cite{gemini2flash2025}, Qwen3 VL~\cite{bai2025qwen3vl}, Gemma 3~\cite{gemmateam2025gemma3}, Llama 4 Scout~\cite{meta2025llama4}, and Nemotron Nano~\cite{deshmukh2025nemotron}.

\subheader{Results}
\cref{tab:exp1} reports detection accuracy and AUC across forgery types and datasets. On FKG-50K, we achieve the best overall accuracy (0.92) and AUC (0.94) with consistently strong per-category performance ($\ge$ 0.94 AUC except AI-edits at 0.88). On OOD datasets, we generalize well and remain competitive with the best specialized systems (DSO-1: 0.96, Synthbuster: 0.96).

Forensic systems only perform well on the specific forgery type they were trained for. Splicing detectors such as CAT-Net excel on their target domain (1.00 AUC on DSO-1) but fail on AI-generated content (0.37 AUC on Full AI), while AI-generation detectors like NPR show the opposite pattern (0.98 AUC on Synthbuster but 0.50 AUC on splicing). No single forensic system covers all forgery types. Conversely, VLMs can detect fully AI-generated images (GPT-5: 0.93 accuracy on GenImage, Qwen3 VL: 0.92 on Synthbuster) but fail on local manipulations (GPT-5: 0.59 on AI-edits, 0.55 on traditional edits). This suggests VLMs identify wholesale generation via image-wide artifacts such as texture regularity or lighting inconsistencies~\cite{wang2025forensicsbench,tariq2025llms}, but cannot detect localized manipulations that preserve overall visual coherence.
% Citation note for VLM detection strategy:
%   - Wang et al., "Forensics-Bench: A Comprehensive Forgery Detection Benchmark Suite for Large VLMs" (CVPR 2025) — https://arxiv.org/abs/2503.15024
%   - Tariq et al., "LLMs Are Not Yet Ready for Deepfake Image Detection" (2025) — https://arxiv.org/abs/2506.10474

% Table: Forgery Type Identification Accuracy
\begin{table}[t]
	\centering
	\caption{\smallcaption Forgery type identification accuracy on FKG-50K and OOD datasets. VLMs show both real-world and oracle-assisted (in parentheses) scenarios. [I] = Instruct, [T] = Thinking.}
	\vspace{-0.5em}
	\resizebox{\linewidth}{!}{%
		\begin{tblr}{
				width=\linewidth,
				colspec={l l *{5}{c} || *{4}{c}},
				hlines, vlines,
				column{1}={c},
				row{2-3}={c},
				row{4-11}={bg=DoublePearlLusta},
				row{12}={bg=Corvette},
				hline{4,12}={1.5pt},
				stretch=0.5,
			}
			\multirow{3}{*}
			& \multirow{3}{*}{\textbf{Method}}
			& \multicolumn{5}{c}{\multirow{2}{*}{\textbf{FKG-50K}}}
			& \multicolumn{4}{c}{\textbf{Out of Distribution Datasets}} \\

			&
			& \multicolumn{5}{c}{}
			& \multicolumn{2}{c}{\textbf{Splice \& Edit}}
			& \multicolumn{2}{c}{\textbf{Fully AI-Gen}} \\

			&
			& \textbf{AI-Edit}
			& \textbf{Full AI}
			& \textbf{Splicing}
			& \textbf{Trad-Edit}
			& \textbf{Overall}
			& \textbf{DSO-1}
			& \textbf{CASIAv2}
			& \textbf{GenImage}
			& \textbf{Synthbuster} \\

			\multirow{8}{*}{\textbf{VLM}}
			& GPT-5 & 0.03 (0.02) & 0.29 (0.39) & 0.49 (0.75) & 0.12 (0.56) & 0.23 (0.43) & 0.00 (0.79) & 0.39 (0.81) & 0.62 (0.54) & 0.56 (0.57) \\
			& Sonnet 4 & 0.04 (0.10) & 0.44 (0.80) & 0.04 (0.32) & 0.02 (0.71) & 0.14 (0.48) & 0.00 (0.26) & 0.03 (0.32) & 0.26 (0.72) & 0.55 (0.86) \\
			& Gemini 2.0 & 0.03 (0.01) & 0.23 (0.40) & 0.40 (0.54) & 0.04 (0.91) & 0.18 (0.47) & 0.04 (0.09) & 0.22 (0.34) & 0.52 (0.66) & 0.54 (0.61) \\
			& Qwen3 VL [I] & 0.03 (0.37) & 0.19 (0.13) & 0.10 (0.16) & 0.02 (0.77) & 0.09 (0.36) & 0.00 (0.00) & 0.01 (0.00) & 0.24 (0.09) & 0.51 (0.39) \\
			& Qwen3 VL [T] & 0.45 (0.49) & 0.45 (0.40) & 0.29 (0.34) & 0.05 (0.28) & 0.31 (0.38) & 0.00 (0.70) & 0.06 (0.23) & 0.38 (0.37) & 0.65 (0.63) \\
			& Gemma 3 & 0.13 (0.95) & 0.28 (0.35) & 0.25 (0.10) & 0.21 (0.05) & 0.22 (0.36) & 0.07 (0.14) & 0.25 (0.38) & 0.38 (0.51) & 0.50 (0.72) \\
			& Llama 4 S. & 0.04 (0.06) & 0.15 (0.91) & 0.02 (0.01) & 0.16 (0.01) & 0.09 (0.25) & 0.00 (0.00) & 0.05 (0.20) & 0.16 (0.09) & 0.21 (0.08) \\
			& Nemotron N. & 0.17 (0.25) & 0.06 (0.33) & 0.12 (0.40) & 0.11 (0.43) & 0.12 (0.35) & 0.14 (0.56) & 0.30 (0.31) & 0.06 (0.20) & 0.05 (0.21) \\

			\textbf{FKG} & \textbf{Ours} & \textbf{0.81} & \textbf{0.84} & \textbf{0.92} & \textbf{0.89} & \textbf{0.87} & 0.67 & 0.71 & \textbf{0.91} & \textbf{0.94} \\
	\end{tblr}}
	\label{tab:exp2_type}
	\vspace{-1.6em}
\end{table}

% In contrast, our FKG generalizes across all forgery types by reasoning over structured, per-region forensic evidence rather than learning a single discriminative boundary. The FKG encodes source identity, compression, and post-processing for each region: differing camera sources reveal splicing, absent camera sources reveal full synthesis, and mixed real/synthetic traces reveal AI-editing. AI-edits are the hardest category because they replace content locally while preserving surrounding forensic structure, producing subtle boundaries. Even so, our FKG achieves 0.88 AUC, substantially outperforming the best forensic system (TruFor, 0.76) and best VLM (GPT-5, 0.60).
In contrast, our FKG generalizes across all forgery types because our backbone's novel self-supervised learning approach learns intrinsic forensic fingerprints of any imaging pipeline, AI generators included. This empowers the FRPN to detect any manipulation by identifying regions with differing forensic fingerprints, and task experts to detect fully AI-generated images via fingerprints distinct from any camera source. AI-edits are the hardest because AI editors condition on real pixels, generating content that partially inherits the original's forensic properties. Despite this, our FKG achieves 0.88 AUC, substantially outperforming the best forensic system (TruFor, 0.76) and best VLM (GPT-5, 0.60).

\vspace{-0.5em}
\subsection{Forgery Type \& Location Identification}
\label{subsec:exp2}
\vspace{-0.5em}

\subheader{Setup}
This experiment tests forgery type identification and localization under two settings. Scenario 1 simulates real-world use, requiring simultaneous detection, classification, and localization, with missed detections penalized. Scenario 2 gives competitors a perfect oracle detector, while \textit{our FKG operates identically in both}. All methods are evaluated on FKG-50K and public OOD datasets.

\subheader{Metrics}
We report forgery-type classification accuracy and localization F1 (predicted vs.\ ground-truth mask), following prior practices~\cite{TruFor,MVSS-Net}. VLM descriptions are converted to masks using Grounding Dino~\cite{liu2023grounding}.

\subheader{Competing Methods}
Same as~\cref{subsec:exp1}, excluding synthetic image detectors from \cref{tab:exp2_type,tab:exp2_loc} (not designed to classify or localize forgeries), and forensic systems from \cref{tab:exp2_type} (not designed to classify forgery types).

\subheader{Forgery Type Classification Results}
\cref{tab:exp2_type} reports forgery type classification accuracy under real-world and oracle-assisted scenarios, where our FKG uses no oracle. On FKG-50K, we achieve the best overall type acc. (0.87) with strong per-category performance (0.81--0.92 across all forgery types). On OOD datasets, we maintain strong generalization (Synthbuster: 0.94, CASIAv2: 0.71).

VLMs fail at this task even under favorable conditions. On FKG-50K without oracle, the best VLM achieves only 0.31 type acc. (Qwen3 VL). With oracle, the best improves to just 0.48 (Sonnet 4), far below our 0.87. In rare instances, VLMs achieved strong performance with an oracle (GPT-5: 0.75 on splicing), though these gains are inconsistent and oracle access can even hurt (GPT-5: 0.62 $\rightarrow$ 0.54 on GenImage). On OOD datasets, oracle-assisted VLMs show competitive numbers (GPT-5: 0.79 on DSO-1, 0.81 on CASIAv2). Without oracle, they score near zero while our FKG achieves its OOD performance unaided.

\subheader{Localization Results}
\cref{tab:exp2_loc} reports forgery localization F1. On FKG-50K, we achieve the best overall F1 (0.94) with consistent per-category performance (0.91--0.98). On OOD data, we achieve 0.95 F1 on DSO-1.

% Table: Forgery Localization F1
\begin{table}[t]
	\centering
	\caption{\smallcaption Forgery localization F1 on FKG-50K and OOD datasets. VLMs show both real-world and oracle-assisted (in parentheses) scenarios. [I] = Instruct, [T] = Thinking.}
	\vspace{-0.5em}
	\resizebox{0.75\linewidth}{!}{%
		\begin{tblr}{
				colspec={l l *{4}{c} || *{2}{c}},
				hlines, vlines,
				column{1}={c},
				row{2-3}={c},
				row{4-11}={bg=DoublePearlLusta},
				row{12-15}={bg=LinkWater},
				row{16}={bg=Corvette},
				hline{4,12,16}={1.5pt},
				stretch=0.5,
			}
			\multirow{3}{*}
			& \multirow{3}{*}{\textbf{Method}}
			& \multicolumn{4}{c}{\multirow{2}{*}{\textbf{FKG-50K}}}
			& \multicolumn{2}{c}{\textbf{OOD}} \\

			&
			& \multicolumn{4}{c}{}
			& \multicolumn{2}{c}{\textbf{Splice \& Edit}} \\

			&
			& \textbf{AI-Edit}
			& \textbf{Splicing}
			& \textbf{Trad-Edit}
			& \textbf{Overall}
			& \textbf{DSO-1}
			& \textbf{CASIAv2} \\

			\multirow{8}{*}{\textbf{VLM}}
			& GPT-5 & 0.17 (0.66) & 0.59 (0.84) & 0.10 (0.56) & 0.29 (0.69) & 0.00 (0.63) & 0.43 (0.77) \\
			& Sonnet 4 & 0.09 (0.46) & 0.08 (0.47) & 0.02 (0.53) & 0.06 (0.49) & 0.00 (0.43) & 0.07 (0.47) \\
			& Gemini 2.0 & 0.05 (0.50) & 0.40 (0.72) & 0.07 (0.45) & 0.17 (0.56) & 0.02 (0.34) & 0.28 (0.65) \\
			& Qwen3 VL [I] & 0.05 (0.65) & 0.25 (0.71) & 0.05 (0.56) & 0.12 (0.64) & 0.00 (0.39) & 0.25 (0.78) \\
			& Qwen3 VL [T] & 0.30 (0.50) & 0.58 (0.69) & 0.30 (0.50) & 0.39 (0.56) & 0.00 (0.34) & 0.40 (0.61) \\
			& Gemma 3 & 0.20 (0.36) & 0.28 (0.48) & 0.22 (0.48) & 0.23 (0.44) & 0.12 (0.22) & 0.22 (0.40) \\
			& Llama 4 S. & 0.03 (0.02) & 0.01 (0.03) & 0.01 (0.03) & 0.02 (0.03) & 0.00 (0.10) & 0.00 (0.05) \\
			& Nemotron N. & 0.17 (0.39) & 0.18 (0.40) & 0.13 (0.34) & 0.16 (0.38) & 0.15 (0.60) & 0.22 (0.49) \\

			\multirow{4}{*}{\shortstack{\textbf{Forensic}\\\textbf{Systems}}}
			& TruFor & 0.27 & 0.81 & 0.29 & 0.46 & 0.93 & \textbf{0.83} \\
			& MVSS-Net & 0.32 & 0.59 & 0.22 & 0.38 & 0.25 & 0.63 \\
			& CAT-Net & 0.38 & 0.64 & 0.40 & 0.47 & 0.59 & 0.66 \\
			& HiFi-IFDL & 0.10 & 0.18 & 0.12 & 0.13 & 0.27 & 0.26 \\

			\textbf{FKG} & \textbf{Ours} & \textbf{0.93} & \textbf{0.98} & \textbf{0.91} & \textbf{0.94} & \textbf{0.95} & 0.73 \\
	\end{tblr}}
	\label{tab:exp2_loc}
	\vspace{-1.6em}
\end{table}

Similar to~\cref{subsec:exp1}, forensic systems only perform well on the forgery types they were trained for. For instance, TruFor excels on splicing (CASIAv2: 0.83 vs.\ our 0.73, DSO-1: 0.93 vs.\ our 0.95) but fails on AI-edits (0.27) and traditional edits (0.29). Similarly, VLMs without oracle cannot localize forgeries (best: 0.39). With oracle, some show reasonable results on specific categories (GPT-5: 0.84 on splicing, Qwen3 VL: 0.78 on CASIAv2). However, these gains are inconsistent and oracle access is unavailable in practice.

% \subheader{Discussion}
%
% These results reveal a clear pattern. Forensic systems can localize anomalies within their training scope but cannot classify what caused them (absent from \cref{tab:exp2_type}). VLMs can sometimes describe forgeries, but only when given unrealistic oracle access, and their performance is inconsistent across categories and datasets. Only our FKG delivers consistent type classification and localization end-to-end under real-world conditions.
% Our FKG's consistent performance across both tasks stems from its generation pipeline: the FRPN identifies forensic region boundaries through all-pairs comparison of patch embeddings via hybrid graph attention, and the task expert networks classify each region's forensic attributes. The FKG ontology structures these outputs into typed cross-region relations that jointly ground both tasks: regional boundaries provide localization ($\ge$ 0.91 F1 across all forgery types), while forensic relations across regions enable accurate type classification: 0.92 for splicing (differing camera sources), 0.89 for traditional edits (localized anomalies), 0.84 for full synthesis (absent camera sources), and 0.81 for AI-edits (mixed real/synthetic traces).
Two factors drive our strong performance across both tasks. First, accurate localization is enabled by the FRPN's novel Hybrid Graph Attention Transformer, which combines global self-attention to capture long-range forensic similarities with local graph attention to suppress false alarms from distant patches, yielding precise region boundaries. Second, accurate type classification is enabled by the task expert networks and the FKG ontology, which enables reasoning over cross-region relations: differing camera sources reveal splicing, absent sources reveal full synthesis, and mixed real/synthetic traces reveal AI-editing.

\vspace{-0.5em}
\subsection{Forensic Decision Justification}
\label{subsec:exp3}
\vspace{-0.5em}

\subheader{Setup}
This experiment measures each system's ability to generate forensic justifications that accurately and completely reference supporting evidence. Our FKG operates under real-world conditions, performing all tasks end-to-end, while competing VLMs are tested with oracle access to perfect ground-truth labels, isolating their reasoning abilities. All methods are evaluated on FKG-50K, which provides reference FKGs with full provenance histories.

\subheader{Metrics}
We assess justification quality with two metrics: correctness (whether forensic facts are described accurately) and completeness (whether all facts from the ground-truth FKG appear in the justification), averaged per image (\cref{eq:metrics}).

\subheader{Competing Methods}
Because classical forensic models do not produce textual explanations, we compare only against the list of advanced VLMs capable of multimodal reasoning in~\cref{subsec:exp1,subsec:exp2}.

\subheader{Results}
\cref{tab:exp3} reports forensic justification correctness and completeness on FKG-50K. Our FKG achieves the best overall correctness (0.75) and completeness (0.85), outperforming all VLMs despite operating without oracle access. Even with oracle access, VLMs reach only 0.16--0.24 correctness and 0.06--0.10 completeness. Per-category, we are strongest on full synthesis (0.86/0.96) and splicing (0.81/0.92), and lowest on traditional edits (0.65/0.75).

\begin{wraptable}{r}{0.63\linewidth}
	\pullupppp\pulluppp\pullupp
	\centering
	\caption{Forensic justification correctness (COR) and completeness (COM) on FKG-50K.}
	\resizebox{\linewidth}{!}{%
		\begin{tblr}{
				width=\linewidth,
				colspec={l l *{10}{c}},
				hlines, vlines,
				column{1}={c},
				row{2}={c},
				row{3-10}={bg=DoublePearlLusta},
				row{11}={bg=Corvette},
			}
			\multirow{2}{*}                                   &
			\multirow{2}{*}{\textbf{Method}}                  &
			\multicolumn{2}{c}{\shortstack{\textbf{AI-Edit}}} &
			\multicolumn{2}{c}{\shortstack{\textbf{Full AI}}} &
			\multicolumn{2}{c}{\shortstack{\textbf{Splice}}}  &
			\multicolumn{2}{c}{\shortstack{\textbf{Trad}}}    &
			\multicolumn{2}{c}{\shortstack{\textbf{Overall}}}                                                                                                                                                                                 \\

			&               & COR           & COM           & COR           & COM           & COR           & COM           & COR           & COM           & COR           & COM           \\

			\multirow{8}{*}{\textbf{VLM}}
			& GPT-5         & 0.15          & 0.05          & 0.28          & 0.16          & 0.33          & 0.12          & 0.08          & 0.03          & 0.21          & 0.09          \\
			& Sonnet 4      & 0.17          & 0.05          & 0.37          & 0.27          & 0.16          & 0.05          & 0.05          & 0.03          & 0.19          & 0.10          \\
			& Gemini 2.0    & 0.17          & 0.02          & 0.30          & 0.14          & 0.38          & 0.06          & 0.10          & 0.03          & 0.24          & 0.06          \\
			& Qwen3 VL [I]  & 0.25          & 0.08          & 0.05          & 0.05          & 0.26          & 0.08          & 0.15          & 0.06          & 0.18          & 0.07          \\
			& Qwen3 VL [T]  & 0.28          & 0.09          & 0.20          & 0.14          & 0.30          & 0.09          & 0.11          & 0.04          & 0.22          & 0.09          \\
			& Gemma 3       & 0.36          & 0.10          & 0.28          & 0.14          & 0.24          & 0.07          & 0.06          & 0.02          & 0.23          & 0.08          \\
			& Llama 4 S.    & 0.05          & 0.01          & 0.62          & 0.28          & 0.03          & 0.01          & 0.00          & 0.00          & 0.17          & 0.08          \\
			& Nemotron N.   & 0.21          & 0.08          & 0.18          & 0.13          & 0.18          & 0.08          & 0.07          & 0.03          & 0.16          & 0.08          \\

			\textbf{FKG}
			& \textbf{Ours} & \textbf{0.69} & \textbf{0.77} & \textbf{0.86} & \textbf{0.96} & \textbf{0.81} & \textbf{0.92} & \textbf{0.65} & \textbf{0.75} & \textbf{0.75} & \textbf{0.85} \\
		\end{tblr}}
	\label{tab:exp3}
	\pullupppp\pullup
\end{wraptable}

These results extend the findings from \cref{subsec:exp1,subsec:exp2}: even when given the forgery type and exact location, VLMs cannot explain the forensic evidence behind a manipulation. The completeness gap is particularly revealing: VLMs cover at most 10\% of forensic facts (Sonnet 4: 0.10), compared to our 85\%. Their justifications rely on visual speculation (``the lighting looks inconsistent'') rather than verifiable forensic traces such as camera model discrepancies, compression inconsistencies, or modification lineage.

Our FKG produces faithful justifications because the FKG ontology encodes each region's source identity, compression history, and modification lineage as structured triplets, giving the VLM (\cref{sec:interp_fkg}) concrete facts to report rather than requiring it to infer forensic conclusions from pixels. Iterative Context Refinement then optimizes the VLM's in-context examples to maximize both correctness and completeness without fine-tuning.

\vspace{-0.7em}
\section{Discussion}
\label{sec:discussion}
\vspace{-0.7em}

\begin{figure}[t]
	\pullupp
	\centering
	\includegraphics[width=1.0\linewidth]{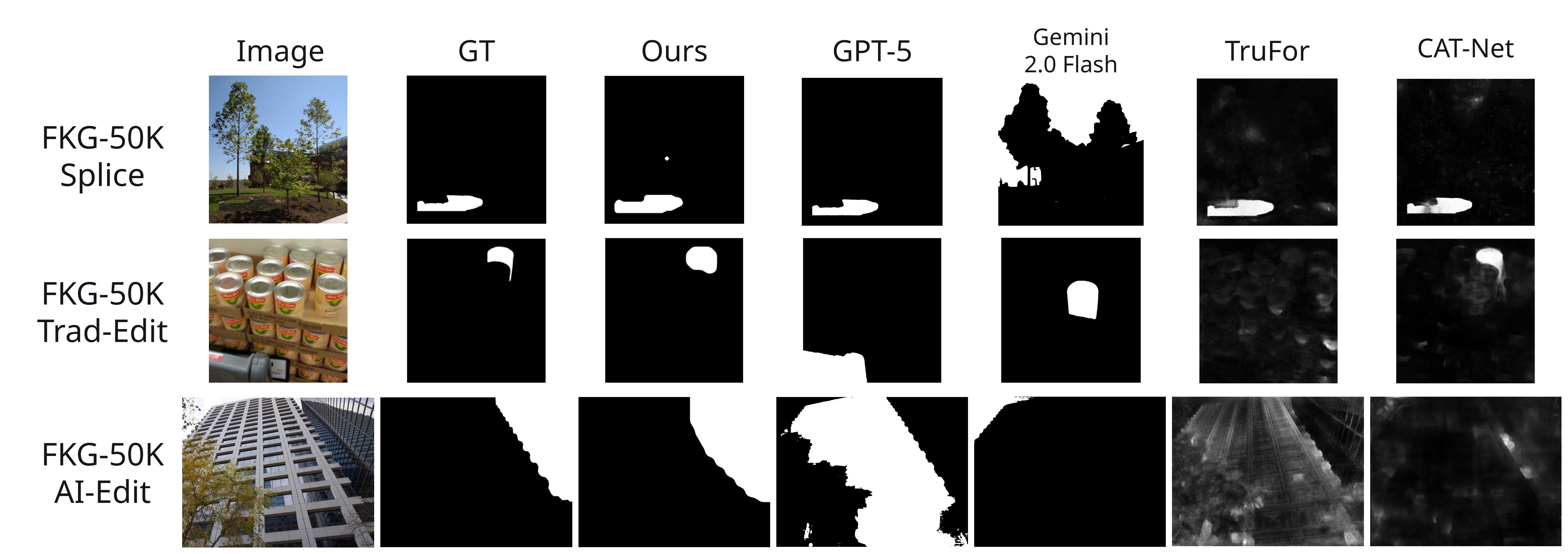}
	\caption{Qualitative comparison of image forgery localization results.}
	\label{fig:results_comparison}
	\vspace{-2.2em}
\end{figure}

\subheader{Effects of Iterative Context Refinement}
We compare the forensic justification quality of different prompting strategies (\cref{tab:disc_prompt_opt}) on identical FKG inputs from FKG-50K.
The results show that our ICR protocol achieves the highest correctness (0.75) and completeness (0.85). Without any context, the VLM covers only 32\% of forensic facts. Random in-context examples improve completeness to 0.51 by demonstrating the expected output format, and adding CoT~\cite{wei2022chain} further improves correctness to 0.70. However, the largest gain comes from ICR, which nearly doubles completeness from 0.55 to 0.85 by iteratively selecting demonstrations that target the VLM's specific failure patterns. Crucially, ICR achieves this without any VLM fine-tuning, operating purely at the prompt level.

\subheader{FKG Accuracy and Failure Modes}
Beyond the correctness and completeness reported in \cref{subsec:exp3}, we evaluate FKG generation accuracy using \emph{perfect match}, which requires every fact in the predicted FKG to exactly match the ground truth. Our system achieves 0.65 perfect match overall, with the highest accuracy on full synthesis (0.81) and splicing (0.68), and the lowest on AI-edits (0.58) and traditional edits (0.52).

\begin{wraptable}{r}{0.49\linewidth}
	\vspace{-3.45em}
	\centering
	\caption{Effect of optimization strategies on forensic justification quality.}
	\label{tab:disc_prompt_opt}
	\resizebox{\linewidth}{!}{
		\begin{tblr}{
				width = \linewidth,
				colspec = {m{29mm}m{10mm}m{25mm}m{25mm}},
				hlines = {1pt,solid},
				vlines,
				column{1-4} = {c},
				colsep = 3pt,
				stretch = 0.6,
				row{5} = {bg=gray!35}
			}

			\textbf{Opt. Strategy} & \textbf{CoT} & \textbf{Correctness} & \textbf{Completeness} \\

			 None & \xmark & 0.64 & 0.32 \\
			 Random & \xmark & 0.63 & 0.51 \\
			 Random & \checkmark & 0.70 & 0.55 \\
			 ICR protocol & \checkmark & \textbf{0.75} & \textbf{0.85} \\

		\end{tblr}
	}
	\vspace{-2.1em}
\end{wraptable}

The gap between correctness (0.75) and perfect match (0.65) reveals that most errors are in fine-grained parameters (\eg, generator model, compression quality) rather than structural relationships, which the system consistently recovers. Full synthesis achieves the highest perfect match (0.81) because its FKGs are structurally simple: one region, one AI source, with fewer parameters to predict. Splicing follows (0.68) due to sharp boundaries between clearly distinct sources. Traditional edits are hardest (0.52) because operations like blur and resampling leave subtle traces that make exact parameter recovery difficult. AI-edits present a different challenge (0.58): some AI generators produce similar forensic fingerprints, making exact source attribution difficult for the current task expert. \cref{fig:results_comparison} shows representative success and failure cases.
% Additional qualitative examples are provided at the end of the supplementary material.

Despite these parameter-level errors, authentication performance remains strong (\cref{sec:experiments}) because correct graph topology suffices for detection, localization, and justification. This structural robustness also explains out-of-distribution generalization: the self-supervised backbone learns forensic fingerprints from imaging pipeline properties rather than image content, and the ontology structures evidence as universal forensic properties rather than forgery-specific features.

\subheader{Human Trust Evaluation}
A core premise of our framework is that grounding authentication in structured forensic evidence makes its decisions more trustworthy to users. To test this directly, 30 participants rated their trust (1 to 5) in our system's output across 5 images, each shown both as a bare verdict and with its accompanying forensic graph explanation. Presenting the graph raised average trust from 3.39 to 4.53, confirming that auditable, evidence-grounded explanations substantially increase user trust over a verdict alone.

\subheader{Limitations}
While our framework provides a strong foundation for trustworthy image authentication, fine-grained parameter prediction and error propagation from imperfect region proposals remain the primary bottlenecks for perfect FKG reconstruction. Importantly, both are addressable through the framework's modular design: more accurate task experts or region proposal methods can be integrated without modifying the ontology, the reasoning pipeline, or ICR. Extended analysis, including computational costs, is provided in the appendix.

\vspace{-0.8em}
\section{Ablation}
\vspace{-0.8em}

We conduct comprehensive ablation studies of each major system component on FKG-50K, shown in \cref{tab:ablation_fkg}.

\subheader{No self-supervision}
Without self-supervised fingerprint learning, the backbone relies only on task labels and loses its ability to encode generalizable forensic microstructures, causing all metrics to drop sharply.

\begin{wraptable}{r}{0.54\linewidth}
	\vspace{-3.5em}
	\centering
	\caption{Ablation study of system components on FKG-50K.}
	\label{tab:ablation_fkg}
	\resizebox{\linewidth}{!}{
		\begin{tblr}{
				width = \linewidth,
				colspec = {m{19mm}m{42mm}m{22mm}m{22mm}m{22mm}},
				hlines = {1pt,solid},
				vlines,
				column{1,3-5} = {c},
				colsep = 3pt,
				stretch = 0.95,
				row{2} = {bg=gray!15}
			}

			\textbf{Stage} & \textbf{Variant} & \textbf{Detect AUC} & \textbf{Type ACC} & \textbf{Loc F1} \\

			– & Ours & \textbf{0.94} & \textbf{0.87} & \textbf{0.94} \\

			\SetCell[r=2]{c}\textbf{Backbone} & No self-supervision & 0.72 (-0.22) & 0.66 (-0.21) & 0.44 (-0.50) \\
			& No pairwise similarity & 0.90 (-0.04) & 0.63 (-0.24) & 0.86 (-0.08) \\

			\SetCell[r=4]{c}\textbf{FRPN} & Only local graph-attention & 0.85 (-0.09) & 0.57 (-0.30) & 0.93 (-0.01) \\
			& Only full self-attention & 0.92 (-0.02) & 0.55 (-0.32) & 0.93 (-0.01) \\
			& Only contrastive loss & 0.89 (-0.05) & 0.52 (-0.35) & 0.72 (-0.22) \\
			& Only Hungarian loss & 0.81 (-0.13) & 0.45 (-0.42) & 0.59 (-0.35) \\

			\SetCell[r=1]{c}\textbf{Attr. Infer} & No transformer reasoner & 0.94 ($\pm$0.00) & 0.69 (-0.18) & 0.94 ($\pm$0.00) \\

		\end{tblr}
	}
	\vspace{-2.1em}
\end{wraptable}

\subheader{No pairwise similarity}
Keeping only the contrastive loss removes the pairwise similarity constraint, weakening structural alignment, reducing type and localization accuracy. Contrastive learning alone fails to capture relational consistency among patches.

\subheader{Only local graph-attention}
Disabling global self-attention limits reasoning to local neighborhoods, causing large drops in type accuracy. Long-range dependencies are essential to model global forgery context.

\subheader{Only full self-attention}
Removing local graph attention breaks spatial coherence, sharply degrading type accuracy. Local connectivity is necessary to preserve region homogeneity and prevent attention dilution.

\subheader{Only contrastive loss}
Training with only contrastive loss omits Hungarian matching, yielding unstable and inconsistent region assignments. Contrastive learning alone cannot form discrete, semantically aligned regions.

\subheader{Only Hungarian loss}
Eliminating the contrastive term leaves only discrete alignment, reducing detection and classification accuracy. Continuous similarity constraints are crucial for coherent region grouping.

\subheader{No transformer reasoner}
Replacing the transformer reasoner with linear classifiers drops type accuracy by 0.18. The reasoning module is essential for integrating cross-region evidence to infer manipulation semantics.

\section{Conclusion}
\label{sec:conclusion}

We introduced Forensic Knowledge Graphs, a unified framework for trustworthy image authentication that integrates forensic evidence extraction, structured reasoning, and interpretable explanation. By modeling images as structured graphs linking regions, sources, and manipulations, our system delivers accurate, localized, and explainable authenticity judgments across diverse forgery types.

{
    \small
    \bibliographystyle{splncs04}
    \bibliography{main}
}

% ---------------------------------------------------------------
% Supplementary Material
% ---------------------------------------------------------------
% Supplementary Material
% ---------------------------------------------------------------

\clearpage
\setcounter{page}{1}
\appendix
% Make hyperref destinations unique so they don't collide with the main paper
% Use \def because newer hyperref (v7+) may not pre-define \theH* as user-level commands
\def\theHpage{suppl.\thepage}
\def\theHsection{suppl.\thesection}
\def\theHsubsection{suppl.\thesubsection}
\def\theHfigure{suppl.\thefigure}
\def\theHtable{suppl.\thetable}
\def\theHequation{suppl.\theequation}

\begin{center}
\Large\textbf{Trustworthy Image Authentication using Forensic Knowledge Graphs}\\[0.5em]
\large Supplementary Material
\end{center}
\vspace{0.5em}

\noindent
\begin{tabular}{@{}ccl@{}}
\toprule
\textbf{Page} & \textbf{Appendix} & \textbf{Title} \\
\midrule
\hyperref[sec:taxonomy]{\pageref*{sec:taxonomy}}  & -- & \hyperref[sec:taxonomy]{Taxonomy of Competing Approaches} \\
\hyperref[app:ontology]{\pageref*{app:ontology}}  & A & \hyperref[app:ontology]{Formal Ontology Definitions} \\
\hyperref[app:backbone]{\pageref*{app:backbone}}  & B & \hyperref[app:backbone]{Backbone Architecture Details} \\
\hyperref[app:frpn]{\pageref*{app:frpn}}          & C & \hyperref[app:frpn]{FRPN Architecture Details} \\
\hyperref[app:training]{\pageref*{app:training}}   & D & \hyperref[app:training]{Forensic Task Expert Reasoner} \\
\hyperref[app:dataset]{\pageref*{app:dataset}}     & E & \hyperref[app:dataset]{FKG-50K Dataset Details} \\
\hyperref[app:prompts]{\pageref*{app:prompts}}     & F & \hyperref[app:prompts]{VLM Prompts \& Response Schema} \\
\hyperref[app:compute]{\pageref*{app:compute}}     & G & \hyperref[app:compute]{Computational Costs \& ICR Convergence} \\
\hyperref[app:examples]{\pageref*{app:examples}}   & H & \hyperref[app:examples]{Qualitative Examples} \\
\bottomrule
\end{tabular}
\vspace{1.5em}

% Symbols and colors for capability taxonomy table
\newcommand{\newcheckmark}{\textrm{\faCheck}}
\newcommand{\newcrossmark}{\textrm{\faTimes}}
\newcommand{\newhalfcirc}{\textrm{\faAdjust}}
\definecolor{Ebb}{rgb}{0.909,0.89,0.89}

\section*{Taxonomy of Competing Approaches}
\label{sec:taxonomy}

\cref{tab:taxonomy} categorizes the competing methods evaluated in the main paper by their authentication capabilities, forgery-type coverage, and whether they leverage signal-level forensic evidence. Our FKG is the only system that combines all four authentication capabilities, generalizes across all forgery types, and grounds its decisions in forensic microstructures rather than visual heuristics.

\begin{table}[h!]
	\centering
	\caption{Categorization of image authentication approaches by capabilities, forgery coverage, and evidence type. \newcrossmark~= no ability or poor performance, \newhalfcirc~= partial or limited, \newcheckmark~= capable.}
	\label{tab:taxonomy}
	\resizebox{\linewidth}{!}{%
		\begin{tblr}{
			colspec = {l cccc cc m{50mm}},
			row{1,2} = {font=\bfseries, c, m},
			row{5,7,10,14,22} = {Ebb},
			column{8} = {font=\mediumcaption},
			cell{1}{1} = {r=2}{l},
			cell{1}{2} = {c=4}{c},
			cell{1}{6} = {c=2}{c},
			cell{1}{8} = {r=2}{l},
			cell{3}{1} = {c=8}{l, font=\normalsize\itshape},
			cell{8}{1} = {c=8}{l, font=\normalsize\itshape},
			cell{12}{1} = {c=8}{l, font=\normalsize\itshape},
			cell{15}{1} = {c=8}{l, font=\normalsize\itshape},
			cell{16}{8} = {r=7}{l, font=\mediumcaption},
			hline{1} = {-}{0.08em},
			hline{3} = {-}{0.05em},
			hline{8,12,15,23} = {-}{},
			hline{24} = {-}{0.08em},
			hline{2} = {2-7}{},
		}
			Method                   & Authentication Capabilities & & & & Properties & & {\normalsize \textbf{Approach}} \\
			                         & Detect & {Forgery\\Type ID} & Localize & Explain & {All Forg.\\Types} & {Forensic\\Evidence} & \\
			Splicing/Editing Detection \& Localization Systems & & & & & & & \\
			CAT-Net~\cite{kwon2022catnet}    & \newcheckmark & \newcrossmark & \newcheckmark & \newcrossmark & \newcrossmark & \newcheckmark & JPEG artifact (DCT) dual-stream. No type classification, no explanation, splicing only \\
			TruFor~\cite{TruFor}             & \newcheckmark & \newcrossmark & \newcheckmark & \newcrossmark & \newcrossmark & \newcheckmark & Noiseprint++ noise-RGB transformer. No type classification, no explanation, no AI-generated \\
			MVSS-Net~\cite{MVSS-Net}         & \newcheckmark & \newcrossmark & \newcheckmark & \newcrossmark & \newcrossmark & \newcheckmark & Multi-scale noise + boundary supervision. No type classification, no explanation, no AI-generated \\
			HiFi-IFDL~\cite{guo2023hifi}     & \newcheckmark & \newcrossmark & \newcheckmark & \newcrossmark & \newcrossmark & \newcheckmark & Hierarchical multi-branch learning. No type classification, no explanation, no AI-generated \\
			AI-Generated Image Detection Systems & & & & & & & \\
			NPR~\cite{NPR}                   & \newcheckmark & \newcrossmark & \newcrossmark & \newcrossmark & \newcrossmark & \newcheckmark & Neighboring pixel relationships. Binary only, no localization, no explanation, AI-generated only \\
			UFD~\cite{UFD}                    & \newcheckmark & \newcrossmark & \newcrossmark & \newcrossmark & \newcrossmark & \newcrossmark & CLIP embedding distances. Binary only, no localization, no explanation, no forensic features \\
			DE-FAKE~\cite{DE-Fake}            & \newcheckmark & \newcrossmark & \newcrossmark & \newcrossmark & \newcrossmark & \newcrossmark & CLIP + BLIP embeddings. Binary only, no localization, no explanation, no forensic features \\
			Fine-tuned Vision-Language Models & & & & & & & \\
			FakeShield~\cite{xu2025fakeshield}  & \newcheckmark & \newcrossmark & \newhalfcirc & \newhalfcirc & \newcrossmark & \newcrossmark & Forgery-tuned VLM. Visual heuristics, not forensic microstructures \\
			ForgeryGPT~\cite{liu2024forgerygpt} & \newcheckmark & \newcrossmark & \newhalfcirc & \newhalfcirc & \newcrossmark & \newcrossmark & Face-forgery-tuned VLM. Visual cues only, limited to face domain \\
			General-Purpose Vision-Language Models & & & & & & & \\
			GPT-5~\cite{openai2025gpt5}               & \newhalfcirc & \newhalfcirc & \newhalfcirc & \newhalfcirc & \newhalfcirc & \newcrossmark & Visual/semantic reasoning only. No forensic training, no signal-level evidence. Justifications are visual speculation, not verifiable forensic facts \\
			Sonnet 4~\cite{anthropic2025claude4}       & \newhalfcirc & \newhalfcirc & \newhalfcirc & \newhalfcirc & \newhalfcirc & \newcrossmark & \\
			Gemini 2.0~\cite{gemini2flash2025}         & \newhalfcirc & \newhalfcirc & \newhalfcirc & \newhalfcirc & \newhalfcirc & \newcrossmark & \\
			Qwen3 VL~\cite{bai2025qwen3vl}             & \newhalfcirc & \newhalfcirc & \newhalfcirc & \newhalfcirc & \newhalfcirc & \newcrossmark & \\
			Gemma 3~\cite{gemmateam2025gemma3}          & \newhalfcirc & \newhalfcirc & \newhalfcirc & \newhalfcirc & \newhalfcirc & \newcrossmark & \\
			Llama 4 Scout~\cite{meta2025llama4}         & \newhalfcirc & \newhalfcirc & \newhalfcirc & \newhalfcirc & \newhalfcirc & \newcrossmark & \\
			Nemotron Nano~\cite{deshmukh2025nemotron}   & \newhalfcirc & \newhalfcirc & \newhalfcirc & \newhalfcirc & \newhalfcirc & \newcrossmark & \\
			\SetCell{Ebb} \textbf{FKG (Ours)} & \SetCell{Ebb} \newcheckmark & \SetCell{Ebb} \newcheckmark & \SetCell{Ebb} \newcheckmark & \SetCell{Ebb} \newcheckmark & \SetCell{Ebb} \newcheckmark & \SetCell{Ebb} \newcheckmark & \SetCell{Ebb} Self-supervised backbone, FRPN, MoE task experts, ontology reasoning, ICR-optimized VLM justification \\
		\end{tblr}%
	}
\end{table}

\section{Formal Ontology Definitions}
\label{app:ontology}

This appendix provides the complete formal specification of the Forensic Knowledge Graph (FKG) ontology introduced in \cref{sec:fkg}. The ontology follows standard knowledge-graph design principles~\cite{hogan2021knowledge} and specifies the classes, object properties, datatype properties, and structural axioms governing valid forensic reasoning structures. The ontology adopts open-world semantics (OWA) and a non-unique-name assumption (NUNA): unobserved relations are treated as unknown rather than false, and distinct nodes may refer to the same real-world entity.

\subsection{Class Hierarchy}

The FKG defines a set of disjoint and hierarchically organized classes. Each class may have subclasses and individuals (entities).

\begin{table}[h]
	\centering
	\caption{FKG ontology class hierarchy.}
	\label{tab:app_class_hierarchy}
	\resizebox{\linewidth}{!}{
	\begin{tabular}{@{}lp{50mm}p{55mm}@{}}
		\toprule
		\textbf{Superclass} & \textbf{Subclass} & \textbf{Description} \\
		\midrule
		\rowcolor{gray!30} Image & -- & Top-level container for the image under analysis. \\
		Region & -- & Non-overlapping pixel subset; atomic unit of forensic reasoning. \\
		\rowcolor{gray!30} Source & RealSource, SyntheticSource & Origin of pixel generation. \\
		\quad RealSource & CameraModel & Physical capture device (\eg, DSLR, phone). \\
		\rowcolor{gray!30} \quad SyntheticSource & AIGenerator & AI model producing the region (\eg, diffusion, GAN). \\
		Post-Processing & NoPostProc, Resampling, GaussianBlur, GaussianNoise, Sharpening & Non-semantic transformations applied after generation. \\
		\rowcolor{gray!30} \quad Parameter & -- & Stores an operation's estimated parameter(s) as key-value nodes. \\
		Compression & Uncompressed, LastCompressed, ReCompressed & Compression or recompression artifacts. \\
		\rowcolor{gray!30} Content & -- & Semantic scene elements: objectedness and object description. \\
		\bottomrule
	\end{tabular}
	}
\end{table}

Each Image must contain at least one Region, and all regions belonging to an image must together cover its entire pixel domain without overlap.

\subsection{Object Properties}

Object properties define typed relations between classes. Each property has a domain, range, and optional cardinality constraint. All object properties are functional in the forward direction unless otherwise stated.

\begin{table}[h]
	\centering
	\caption{FKG ontology object properties.}
	\label{tab:app_object_props}
	\resizebox{\linewidth}{!}{
	\begin{tabular}{@{}lllp{55mm}@{}}
		\toprule
		\textbf{Property} & \textbf{Domain} & \textbf{Range} & \textbf{Cardinality / Semantics} \\
		\midrule
		\rowcolor{gray!30} \fkgprop{contained\_in} & Region & Image & Each region belongs to exactly one image. \\
		\fkgprop{produced\_by} & Region & Source & Each region is generated by exactly one source. \\
		\rowcolor{gray!30} \fkgprop{modified\_by} & Region & Post-Processing & Optional; a region may have 0 or $\geq$1 post-processing ops. \\
		\fkgprop{compress\_by} & Region & Compression & Optional; a region may have 0 or 1 compression node. \\
		\rowcolor{gray!30} \fkgprop{depict} & Region & Content & Each region depicts exactly one content node. \\
		\fkgprop{hasParameter} & Post-Processing & Parameter & One-to-many; connects an operation to its parameters. \\
		\rowcolor{gray!30} \fkgprop{hasNext} & Post-Processing & Post-Processing & Optional; defines ordering for sequential operations. \\
		\bottomrule
	\end{tabular}
	}
\end{table}

\subsection{Datatype Properties}

Datatype properties assign scalar, string, or structured values to class instances.

\begin{table}[h]
	\centering
	\caption{FKG ontology datatype properties.}
	\label{tab:app_datatype_props}
	\resizebox{\linewidth}{!}{
	\begin{tabular}{@{}lllp{45mm}@{}}
		\toprule
		\textbf{Property} & \textbf{Domain} & \textbf{Type} & \textbf{Example / Semantics} \\
		\midrule
		\rowcolor{gray!30} \fkgprop{hasConfidence} & Source & float $\in [0,1]$ & Confidence of source attribution \\
		\fkgprop{hasCameraModel} & CameraModel & string & \eg, ``Canon EOS R6'' \\
		\rowcolor{gray!30} \fkgprop{hasGeneratorModel} & AIGenerator & string & \eg, ``Stable Diffusion XL'' \\
		\fkgprop{hasParameterName} & Parameter & string & \eg, ``sigma'', ``scaleFactor'' \\
		\rowcolor{gray!30} \fkgprop{hasParameterValue} & Parameter & float / structured & Numeric or tuple value \\
		\fkgprop{hasCompressionAlgorithm} & LastCompressed & string & \eg, ``JPEG'', ``WebP'' \\
		\rowcolor{gray!30} \fkgprop{hasCompressionFactor} & LastCompressed & float & \eg, JPEG quality factor \\
		\fkgprop{hasReCompressed} & ReCompressed & boolean & True if recompression detected \\
		\rowcolor{gray!30} \fkgprop{hasObjectedness} & Content & boolean & True if region contains a salient object \\
		\fkgprop{hasObjectDescription} & Content & string & \eg, ``truck'', ``person'' \\
		\bottomrule
	\end{tabular}
	}
\end{table}

\subsection{Structural Axioms}

The ontology enforces structural axioms that guarantee internal consistency and encode known causal relationships in image formation.

\subheader{Axiom 1 --- Region Partition Constraint}
Every image is decomposed into a finite, non-overlapping set of regions whose union covers the entire pixel lattice. This enforces that forensic reasoning operates on complete, non-redundant spatial partitions.

\begin{equation*}
	\forall I,\quad
	\bigcup_{r\in \text{Regions}(I)} \text{Pixels}(r)=\text{Pixels}(I),\qquad
	\forall i\neq j,\ \text{Pixels}(r_i)\cap \text{Pixels}(r_j)=\varnothing.
\end{equation*}

\subheader{Axiom 2 --- Functional Source Attribution}
Each region is generated by exactly one source, ensuring that every pixel subset has a unique origin hypothesis.

\begin{equation*}
	\forall r:\text{Region},\ \exists!\, s:\text{Source},\ \texttt{produced\_by}(r,s).
\end{equation*}

\subheader{Axiom 3 --- Mutual Exclusivity of Source Types}
A region cannot simultaneously originate from both a real and synthetic source.

\begin{equation*}
	\forall s:\text{Source},\ \neg\!\big(\text{instanceOf}(s,\text{RealSource}) \land \text{instanceOf}(s,\text{SyntheticSource})\big).
\end{equation*}

\subheader{Axiom 4 --- Causal Ordering}
Post-processing and compression follow the causal direction of real imaging processes. Compression is modeled as the final operation in the chain.

\begin{equation*}
	\begin{aligned}
	\forall r,\, s,\, m,\, c:\
	&\texttt{produced\_by}(r,s) \land \texttt{modified\_by}(r,m) \\
	&\land\ \texttt{compress\_by}(r,c)
	\ \Rightarrow\ s \prec m \prec c.
	\end{aligned}
\end{equation*}

\subsection{Example Subgraph Schemas}

We present three canonical instantiations of the ontology corresponding to distinct forensic scenarios.

\subheader{1. Authentic (Camera-Captured) Image}

\begin{codebox}{green}
  r1 : Region
  |-- produced_by   -> s1 : RealSource
  |   \-- hasCameraModel = "Canon EOS R6"
  |-- modified_by   -> m1 : NoPostProc
  |-- compress_by   -> c1 : LastCompressed
  |   |-- hasCompressionAlgorithm = "JPEG"
  |   \-- hasCompressionFactor = 0.9
  |-- depict        -> d1 : Content
  |   |-- hasObjectedness = True
  |   \-- hasObjectDescription = "landscape"
  \-- contained_in  -> I : Image
\end{codebox}

\subheader{2. Fully Synthetic Image}

\begin{codebox}{green}
  r1 : Region
  |-- produced_by   -> s1 : SyntheticSource
  |   \-- hasGeneratorModel = "SDXL"
  |-- modified_by   -> m1 : NoPostProc
  |-- compress_by   -> c1 : LastCompressed
  |   |-- hasCompressionAlgorithm = "WebP"
  |   \-- hasCompressionFactor = 0.8
  |-- depict        -> d1 : Content
  |   |-- hasObjectedness = True
  |   \-- hasObjectDescription = "person"
  \-- contained_in  -> I : Image
\end{codebox}

\subheader{3. Locally Manipulated Image (AI-editing + Blur)}

\begin{codebox}{green}
  r1 : Region  (authentic background)
  |-- produced_by   -> s1 : RealSource
  |   \-- hasCameraModel = "Nikon Coolpix S33"
  |-- modified_by   -> m1 : NoPostProc
  |-- compress_by   -> c1 : ReCompressed
  |   \-- hasReCompressed = True
  |-- depict        -> d1 : Content
  |   |-- hasObjectedness = False
  |   \-- hasObjectDescription = "background"
  \-- contained_in  -> I : Image

  r2 : Region  (AI-edited object)
  |-- produced_by   -> s2 : SyntheticSource
  |   \-- hasGeneratorModel = "FLUX"
  |-- modified_by   -> m2 : GaussianBlur
  |   \-- hasParameter -> p1 : Parameter
  |       |-- hasParameterName = "sigma"
  |       \-- hasParameterValue = 2.0
  |-- compress_by   -> c2 : LastCompressed
  |   \-- hasCompressionAlgorithm = "JPEG"
  |-- depict        -> d2 : Content
  |   |-- hasObjectedness = True
  |   \-- hasObjectDescription = "truck"
  \-- contained_in  -> I : Image
\end{codebox}

\section{Backbone Architecture Details}
\label{app:backbone}

\begin{figure}[t]
	\centering
	\includegraphics[width=\linewidth]{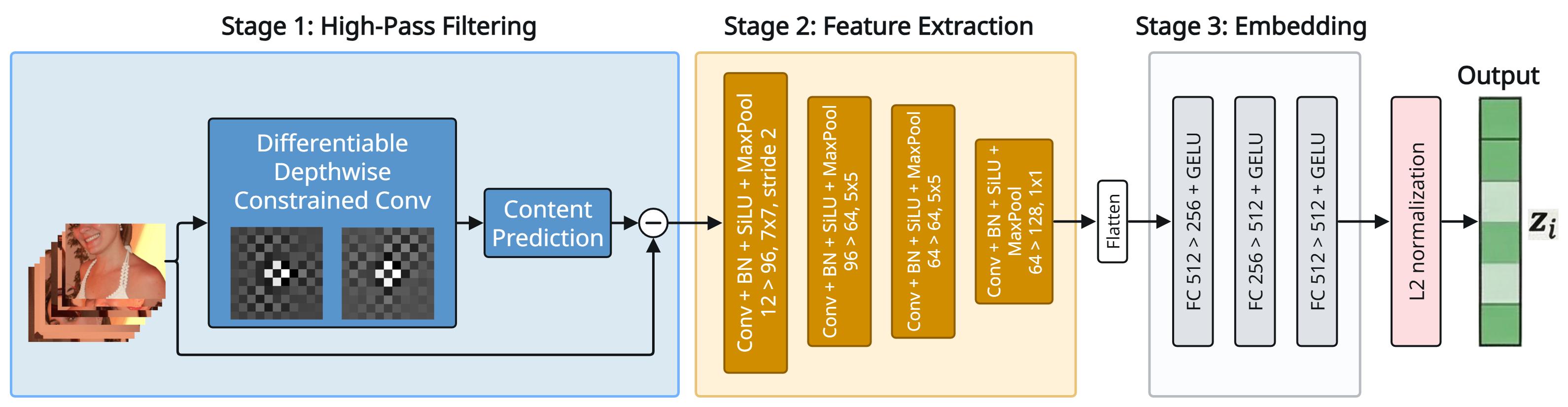}
	\caption{Lightweight Forensic Backbone Network architecture.}
	\label{fig:backbone_arch}
\end{figure}

This appendix provides the full architecture and training details of the Forensic Backbone Network described in \cref{sec:fkg_sys}. The backbone introduces two key contributions over prior forensic CNNs: (1) a fully differentiable constrained convolutional layer with depthwise grouped convolution, and (2) auxiliary regularization losses that improve forensic residual quality and filter diversity.

\subsection{Differentiable Constrained Convolution}

Prior forensic CNNs~\cite{bayar2016deep,MISLnet} use a constrained convolutional layer to suppress image content and isolate forensic residuals. The original formulation applies the constraint as a post-hoc weight projection after each gradient step, breaking the computational graph and preventing gradient flow through the constraint itself. We reformulate this as a fully differentiable operation applied within the forward pass, enabling end-to-end training with proper gradient propagation through the constraint.

Our constrained layer uses \emph{depthwise grouped convolution} with $M$ filters per input channel ($3M$ total for RGB), each with a $K \times K$ kernel subject to two constraints: (1) the center weight is always zero, and (2) all non-center weights sum to one. These constraints force each filter to predict a pixel from its neighbors, and the forensic residual is the difference between the input and this prediction. The full procedure is described in \cref{alg:constrained_conv}.

\begin{figure}[t]
	\begin{minipage}{\linewidth}
		\begin{algorithm}[H]
			\caption{Differentiable Constrained Convolution (Forward Pass)}
			\label{alg:constrained_conv}
			\begin{algorithmic}[1]
				\REQUIRE Input image $\mathbf{x} \in \mathbb{R}^{C \times H \times W}$, learnable weights $\mathbf{W} \in \mathbb{R}^{CM \times 1 \times K \times K}$
				\ENSURE Forensic residual $\mathbf{r} \in \mathbb{R}^{CM \times H \times W}$
				\STATE \textbf{// Apply constraints (differentiable)}
				\STATE Reshape $\mathbf{W}$ to $\mathbf{W}' \in \mathbb{R}^{CM \times K^2}$
				\STATE Center and normalize: $w'_{ij} \leftarrow w'_{ij} - \bar{w}'_i + \frac{1}{K^2 - 1}$
				\STATE Compute scaling: $s_i \leftarrow \sum_{j \neq \text{center}} w'_{ij}$
				\STATE Normalize non-center: $w'_{ij} \leftarrow w'_{ij} / s_i$
				\STATE Zero center: $w'_{i,\text{center}} \leftarrow 0$
				\STATE Reshape back to $\hat{\mathbf{W}} \in \mathbb{R}^{CM \times 1 \times K \times K}$
				\STATE \textbf{// Depthwise grouped convolution}
				\STATE $\mathbf{x}_{\text{pad}} \leftarrow \text{ReflectPad}(\mathbf{x}, K//2)$
				\STATE $\hat{\mathbf{x}} \leftarrow \text{Conv2d}(\mathbf{x}_{\text{pad}},\, \hat{\mathbf{W}},\, \text{groups}=C)$ \hfill \textit{// content prediction}
				\STATE \textbf{// Compute forensic residual}
				\STATE $\mathbf{x}_{\text{rep}} \leftarrow \text{RepeatInterleave}(\mathbf{x}, M, \text{dim}=1)$
				\STATE $\mathbf{r} \leftarrow \mathbf{x}_{\text{rep}} - \hat{\mathbf{x}}$ \hfill \textit{// residual = input $-$ prediction}
				\RETURN $\mathbf{r}$
			\end{algorithmic}
		\end{algorithm}
	\end{minipage}
\end{figure}

Because all operations in \cref{alg:constrained_conv} (centering, scaling, zeroing) are algebraic and differentiable, gradients flow through the constraint during backpropagation. The depthwise grouped structure is also forensically motivated: different color channels carry different forensic traces (\eg, the green channel in Bayer-pattern sensors has twice the spatial sampling density, producing distinct noise characteristics from red and blue), and depthwise convolution preserves these channel-specific signatures rather than mixing them prematurely. With $M=4$ and $K=5$, this produces 12 residual feature maps from an RGB input.

\subsection{Full Architecture}

\begin{wraptable}{r}{0.55\linewidth}
	\vspace{-1.2em}
	\centering
	\caption{Backbone architecture (128$\times$128 RGB input).}
	\label{tab:app_backbone_arch}
	\resizebox{\linewidth}{!}{
		\begin{tabular}{@{}llccl@{}}
			\toprule
			\textbf{Stage} & \textbf{Layer}    & \textbf{Channels}   & \textbf{Kernel} & \textbf{Notes}   \\
			\midrule
			1              & ConstrainedConv2d & 3$\rightarrow$12    & 5$\times$5      & depthwise, $M$=4 \\
			\midrule
			2a             & Conv+BN+SiLU+Pool & 12$\rightarrow$96   & 7$\times$7      & stride 2, valid  \\
			2b             & Conv+BN+SiLU+Pool & 96$\rightarrow$64   & 5$\times$5      & stride 1, same   \\
			2c             & Conv+BN+SiLU+Pool & 64$\rightarrow$64   & 5$\times$5      & stride 1, same   \\
			2d             & Conv+BN+SiLU+Pool & 64$\rightarrow$128  & 1$\times$1      & stride 1, same   \\
			\midrule
			3a             & Linear+GELU       & 512$\rightarrow$256 & --              & --               \\
			3b             & Linear+GELU       & 256$\rightarrow$512 & --              & --               \\
			3c             & Linear+GELU       & 512$\rightarrow$512 & --              & --               \\
			\midrule
			4              & $\ell_2$ norm     & 512                 & --              & unit embedding   \\
			\bottomrule
		\end{tabular}
	}
	\vspace{-2em}
\end{wraptable}

The backbone takes 128$\times$128 RGB patches and produces $\ell_2$-normalized 512-dimensional embeddings (\cref{tab:app_backbone_arch}). The constrained layer produces 12 forensic residual maps (Stage 1), which are processed by four convolutional blocks each consisting of Conv2d, BatchNorm2d, SiLU activation, and MaxPool2d(3, 2) (Stage 2). After the final block, features are flattened (2$\times$2$\times$128 = 512) and passed through three FC layers with GELU activations (Stage 3). The output is $\ell_2$-normalized to produce $z_i \in \mathbb{R}^{512}$.

\subsection{Auxiliary Training Losses}

In addition to the pairwise similarity and contrastive losses described in \cref{sec:fkg_sys}, we introduce two optional auxiliary regularization terms that improve training convergence and forensic residual quality. The backbone can be trained without these terms, but we find they accelerate convergence and produce higher-quality forensic residuals.

\subheader{Residual RMSE Loss}
This term regularizes the constrained convolutional layer by encouraging each filter to accurately predict image content from its neighbors. Minimizing the RMSE between the input and the constrained prediction pushes the filters toward better content estimation, so that the resulting residual isolates forensic microstructures rather than retaining image content. Without this term, the constrained filters can converge to poor predictions whose residuals are dominated by edges and texture rather than forensic traces. The loss is:
\begin{equation*}
	\mathcal{L}_{\text{res}} = \gamma_{\text{res}} \cdot \text{RMSE}(\mathbf{x}_{\text{rep}},\, \hat{\mathbf{x}})
\end{equation*}
where $\gamma_{\text{res}}$ controls the regularization strength.

\begin{wraptable}{r}{0.4\linewidth}
	\vspace{-3.4em}
	\centering
	\caption{Backbone training hyperparameters.}
	\label{tab:app_backbone_training}
	\adjustbox{max width=0.8\linewidth}{
	\begin{tabular}{@{}lp{17mm}@{}}
		\toprule
		\textbf{Hyperparameter} & \textbf{Value} \\
		\midrule
		Optimizer & AdamW \\
		Learning rate & $1 \times 10^{-3}$ \\
		Weight decay & $1 \times 10^{-2}$ \\
		LR schedule & 0.8 / 2 ep \\
		Minimum LR & $1 \times 10^{-6}$ \\
		Batch size & 64 images \\
		Epochs & 30 \\
		\midrule
		\multicolumn{2}{@{}l@{}}{\textit{Loss weights}} \\
		Pairwise $\alpha, \beta$ & 10.0, 10.0 \\
		Contrastive $\lambda$ & 0.1 \\
		Temperature $\tau$ & 0.07 \\
		Residual $\gamma_{\text{res}}$ & 25.5 \\
		Spectral $\gamma_{\text{det}}$ & 0.1 \\
		Spectral $\gamma_{\text{sv}}$ & 0.05 \\
		\bottomrule
	\end{tabular}
	}
	\vspace{-2em}
\end{wraptable}

\subheader{Spectral Diversity Loss}
This term encourages the learned constrained filters to capture diverse forensic features by penalizing singular value concentration in the filter weight matrix. Forensic fingerprints comprise multiple independent components (sensor noise patterns, demosaicing artifacts, compression traces), and each constrained filter should specialize in a different component. Without this term, multiple filters can converge to similar solutions, reducing the effective number of independent forensic features. The loss combines a log-determinant term (encouraging spread across all singular values) with a singular value penalty (discouraging any single dominant direction):
\begin{equation*}
	\mathcal{L}_{\text{spec}} = -\gamma_{\text{det}} \cdot \log\det(\mathbf{W}^{\top}\mathbf{W} + \epsilon\mathbf{I}) + \gamma_{\text{sv}} \cdot \|\sigma_1(\mathbf{W})\|
\end{equation*}
where $\sigma_1$ is the largest singular value.

The full backbone training loss is:
\begin{equation*}
	\mathcal{L}_{B} = \mathcal{L}_{\text{pairwise}} + \lambda \mathcal{L}_{\text{contrastive}} + \mathcal{L}_{\text{res}} + \mathcal{L}_{\text{spec}}
\end{equation*}

\subsection{Pretraining Dataset}

The backbone is pretrained exclusively on authentic, camera-captured photographs with no synthetic or manipulated images. The training corpus, MIDB--Unsplash--Pexels (MUP), is assembled from three sources:

\begin{wraptable}{r}{0.40\linewidth}
	\vspace{-3.4em}
	\centering
	\caption{Pretraining corpus (authentic images only).}
	\label{tab:app_backbone_data}
	\adjustbox{max width=\linewidth}{
	\begin{tabular}{@{}lc@{}}
		\toprule
		\textbf{Source}          & \textbf{Images}  \\
		\midrule
		MIDB                     & 21,605           \\
		Pexels~\cite{pexels}     & 86,309           \\
		Unsplash~\cite{unsplash} & 202,312          \\
		\midrule
		\textbf{Total}           & \textbf{310,226} \\
		\bottomrule
	\end{tabular}
	}
	\vspace{-2em}
\end{wraptable}

All images are filtered to ensure they are camera-original with reliable EXIF metadata (camera make/model, color pipeline, compression parameters). The final corpus spans over 350 unique camera models across 20 manufacturers. FKG-50K (\cref{app:dataset}) is built from a curated 72-model subset of this corpus with fully trusted provenance, while the backbone trains on the full corpus to maximize forensic fingerprint diversity. No forgery labels, source annotations, or data augmentation are used during pretraining, as common augmentations (\eg, cropping, resizing, recompression) can destroy or alter the forensic traces the backbone aims to capture.

\subsection{Patch Sampling}

Each training image is cropped to at most 1920$\times$1920 pixels (aligned to 16px boundaries) and sampled into 128$\times$128 patches using a center-neighbor strategy: 56 center locations are randomly selected per image, and for each center, 3 neighboring patches are sampled from the 8 adjacent positions (up, down, left, right, and four diagonals) at a one-patch-size offset. This yields 168 patches per image. Patches from the same image share a forensic fingerprint ($Y_{ij}=1$), while patches from different images do not ($Y_{ij}=0$), providing the self-supervised training signal. The pairwise weights $w_{ij}$ in the backbone loss (\cref{sec:fkg_sys}) are set using $\alpha$ and $\beta$: positive pairs are weighted by $\alpha$ and negative pairs by $\beta$. All training hyperparameters are listed in \cref{tab:app_backbone_training}.

\section{FRPN Architecture Details}
\label{app:frpn}

\begin{figure}[t]
	\centering
	\includegraphics[width=\linewidth]{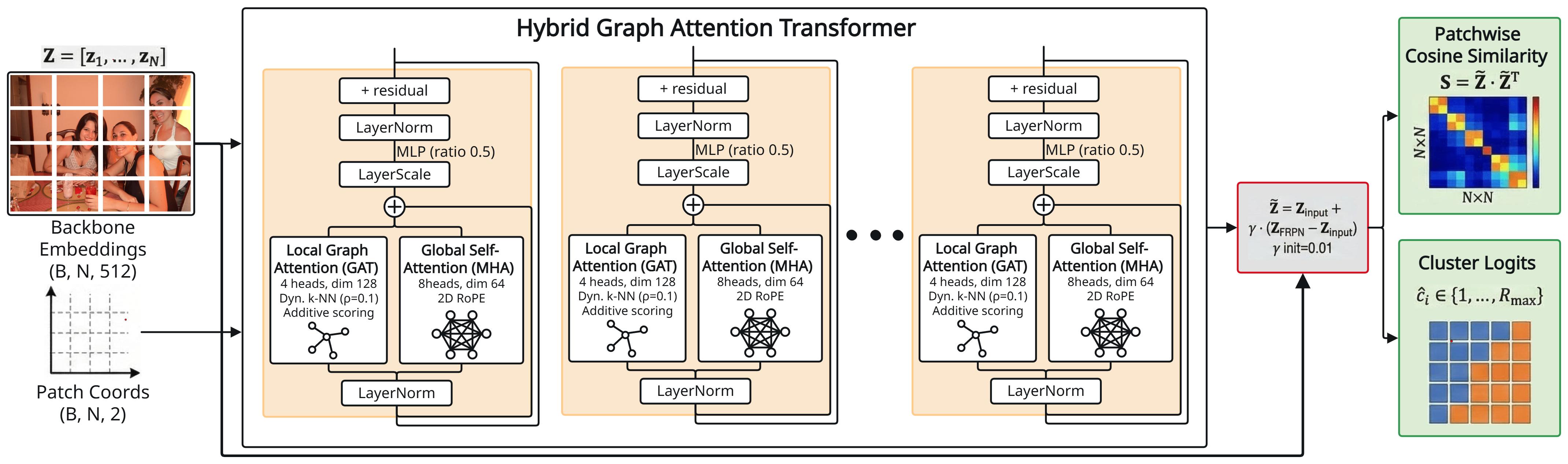}
	\caption{Forensic Region Proposal Network (FRPN) architecture.}
	\label{fig:frpn_arch}
\end{figure}

This appendix provides the full architecture and training details of the Forensic Region Proposal Network (FRPN) described in \cref{sec:fkg_sys}. The FRPN addresses a fundamental challenge unique to image forensics: partitioning an image into regions defined not by visual boundaries but by invisible statistical properties. Existing region proposal and segmentation methods~\cite{he2017mask,kirillov2023sam} operate on visual features and cannot detect forensic region boundaries, because a spliced region may appear visually seamless yet carry an entirely different forensic fingerprint. We introduce three novel design choices to solve this: (1) a Hybrid Graph Attention Transformer that alternates global self-attention with local graph attention for robust forensic clustering, (2) a suite of training stability mechanisms (LayerScale, output-level residual blend, near-identity initialization) that preserve the pretrained backbone signal while the FRPN learns to refine it, and (3) a multi-objective loss combining focal weighting, residual regularization, and hard boundary cropping for precise forensic region discovery.

\begin{wraptable}{r}{0.50\linewidth}
	\vspace{-3.5em}
	\centering
	\caption{FRPN architecture summary.}
	\label{tab:app_frpn_arch}
	\adjustbox{max width=\linewidth}{
	\begin{tabular}{@{}lp{30mm}@{}}
		\toprule
		\textbf{Parameter} & \textbf{Value} \\
		\midrule
		Embedding dim & 512 \\
		Alternating layers $L$ & 8 (16 blocks total) \\
		Global attention heads & 8 \\
		Local attention heads & 4 \\
		MLP ratio & 0.5 (hidden = 256) \\
		$k$-NN ratio $\rho$ & 0.1 \\
		LayerScale init & $1 \times 10^{-2}$ \\
		Output blend $\gamma_0$ & 0.01 \\
		Max regions & 2 \\
		Patch size / stride & 128 / 64 \\
		\bottomrule
	\end{tabular}
	}
	\vspace{-2.2em}
\end{wraptable}

\subsection{Hybrid Graph Attention Transformer}

The core insight behind the FRPN is that partitioning an image into forensically homogeneous regions requires comparing every patch against every other, the all-pairs operation that self-attention naturally computes. However, as discussed in \cref{sec:fkg_sys}, global self-attention suffers from attention dilution: accumulated contributions from distant, unrelated patches cause forensically distinct regions to appear connected. We address this with a novel hybrid architecture that alternates two complementary attention mechanisms in $L$ layer pairs, each serving a distinct forensic purpose. The full forward pass is described in \cref{alg:frpn_forward}.

\begin{figure}[t]
	\begin{minipage}{\linewidth}
		\begin{algorithm}[H]
			\caption{Hybrid Graph Attention Transformer (FRPN Forward Pass)}
			\label{alg:frpn_forward}
			\begin{algorithmic}[1]
				\REQUIRE Backbone embeddings $Z \in \mathbb{R}^{N \times d}$, patch coordinates $\mathbf{p} \in [0,1]^{N \times 2}$
				\ENSURE Refined embeddings $\tilde{Z}$, similarity $S$, cluster logits $C$
				\STATE $Z^{(0)} \leftarrow \ell_2\text{-normalize}(Z)$; \enspace $Z_{\text{input}} \leftarrow Z^{(0)}$
				\FOR{$l = 1$ \TO $L$}
				\STATE \textbf{// Local graph attention --- suppress false alarms}
				\STATE $\mathcal{N}_i \leftarrow k\text{-NN}(Z^{(2l-2)})$ for each patch $i$ \hfill \textit{// dynamic neighborhood}
				\STATE $Z^{(2l-1)} \leftarrow \text{GAT}(Z^{(2l-2)},\, \mathcal{N})$ \hfill \textit{// additive-score, 4 heads}
				\STATE \textbf{// Global self-attention --- capture long-range forensic consistency}
				\STATE $Z^{(2l)} \leftarrow \text{MHA}(Z^{(2l-1)},\, \mathbf{p})$ \hfill \textit{// 2D RoPE on Q, K}
				\ENDFOR
				\STATE \textbf{// Output-level residual blend --- preserve backbone signal}
				\STATE $\tilde{Z} \leftarrow Z_{\text{input}} + \gamma \cdot (Z^{(2L)} - Z_{\text{input}})$ \hfill \textit{// $\gamma$ learnable, init 0.01}
				\STATE $\tilde{Z} \leftarrow \ell_2\text{-normalize}(\tilde{Z})$
				\STATE $S \leftarrow \tilde{Z}\, \tilde{Z}^{\top}$ \hfill \textit{// pairwise cosine similarity}
				\STATE $C \leftarrow \text{Linear}(\tilde{Z})$ \hfill \textit{// cluster assignment logits}
				\RETURN $\tilde{Z},\, S,\, C$
			\end{algorithmic}
		\end{algorithm}
	\end{minipage}
\end{figure}

\subheader{Global Self-Attention with 2D RoPE}
The global attention layers use standard multi-head self-attention (8 heads, head dimension 64) augmented with 2D Rotary Position Embeddings (RoPE). Given normalized patch coordinates $(y, x) \in [0, 1]^2$, the RoPE module computes sinusoidal frequencies along each spatial axis and applies rotary transforms to the query and key vectors. Unlike learned position embeddings, RoPE naturally generalizes to variable-sized images at inference time without retraining. This is forensically motivated: manipulation boundaries are inherently spatial, and encoding spatial relationships directly into attention allows the model to distinguish between ``same fingerprint, same location'' (strong evidence of a common region) and ``same fingerprint, distant location'' (weaker but still informative).

\subheader{Local Graph Attention with Dynamic $k$-NN}
The local attention layers implement additive-score GAT~\cite{velickovic2018graph} with dynamic $k$-nearest neighbors (4 heads, head dimension 128). This mechanism directly addresses the attention dilution problem: by restricting each patch's attention to its $k$ most forensically similar neighbors, irrelevant contributions from distant patches are eliminated rather than merely downweighted. The neighborhood size is $k = \max(8, \lfloor \rho \cdot N \rfloor)$ with $\rho = 0.1$, ensuring a minimum of 8 neighbors regardless of image size. Critically, neighbors are recomputed at each forward pass based on cosine similarity of the current embeddings, making the graph structure adaptive: as the embeddings evolve through the layers and forensic clusters become more distinct, the neighborhoods tighten around true forensic neighbors. Self-loops are excluded to prevent trivial self-reinforcement.

\subheader{Why Alternating Attention}
The alternation of local and global layers is not arbitrary but follows a specific forensic reasoning pattern. Local graph attention first refines each patch's representation using only its most forensically similar neighbors, suppressing noise from unrelated patches and sharpening cluster boundaries. Global self-attention then propagates this refined information across the entire image, connecting distant patches that belong to the same forensic region (\eg, two halves of an authentic background separated by a spliced object). This local-then-global pattern repeats $L=8$ times, progressively building forensic consensus from local evidence to global structure.

\subheader{Cluster Assignment Head}
The refined, $\ell_2$-normalized embeddings are passed through a single linear layer that outputs logits over $R_{\max}$ clusters. Currently $R_{\max} = 2$ (authentic region vs.\ manipulated region), though the architecture supports any value. Pairwise cosine similarity is also computed from the refined embeddings for the pairwise similarity loss.

\subsection{Training Stability}

The FRPN operates on top of pretrained backbone embeddings that already carry forensic signal. A naive transformer could overwhelm this signal in early training, before meaningful attention patterns have been learned. We introduce three mechanisms to ensure the FRPN begins as a near-identity function and gradually learns to refine the backbone's output.

\subheader{LayerScale Attention Block}
Each attention block follows the pre-norm transformer design with LayerScale~\cite{touvron2021going}. Given input $x$, one block computes:
\begin{equation*}
	x \leftarrow x + \boldsymbol{\gamma}_{\text{attn}} \odot \text{Attn}(\text{LN}(x)), \quad
	x \leftarrow x + \boldsymbol{\gamma}_{\text{mlp}} \odot \text{MLP}(\text{LN}(x)),
\end{equation*}
where $\boldsymbol{\gamma}_{\text{attn}}, \boldsymbol{\gamma}_{\text{mlp}} \in \mathbb{R}^{d}$ are learnable per-channel scaling vectors initialized to $10^{-2}$, and LN is LayerNorm. This prevents the skip connection from being overwhelmed early in training, when attention weights are essentially random. The MLP uses a single hidden layer with GELU activation and a hidden dimension of $0.5 \times 512 = 256$. The last MLP layer is initialized with small weights (Xavier gain $= 0.01$), further ensuring each block is near-identity at initialization.

\subheader{Output-Level Residual Blend}
After all attention blocks, the FRPN blends its output with the original backbone embeddings via a learnable scalar $\gamma$ (initialized to 0.01, clamped to $[0, 1]$):
\begin{equation*}
	\tilde{Z} = \ell_2\text{-norm}\!\big(Z_{\text{input}} + \gamma \cdot (Z_{\text{FRPN}} - Z_{\text{input}})\big)
\end{equation*}
At initialization $\gamma = 0.01$, so downstream components operate on essentially the raw backbone embeddings. As training progresses, $\gamma$ increases to allow the FRPN to contribute more strongly. Unlike standard per-block residual connections, this operates at the output level of the entire network, providing a global safety valve that preserves the backbone's forensic signal.

\subsection{Training Losses}

The FRPN loss extends the pairwise + Hungarian formulation from \cref{sec:fkg_sys} with two optional auxiliary terms designed for forensic region discovery. The core focal MSE and Hungarian losses are sufficient for training, but we find these additional terms improve boundary precision and training stability:

\begin{equation*}
	\mathcal{L}_F = \mathcal{L}_{\text{MSE}} + \lambda_{\text{H}} \mathcal{L}_{\text{Hungarian}} + \lambda_{\text{res}} \mathcal{L}_{\text{res}} + \lambda_{\text{crop}} \mathcal{L}_{\text{crop}}
\end{equation*}

\begin{itemize}[leftmargin=5mm, noitemsep, topsep=0pt]
	\item \textbf{Focal MSE} ($\mathcal{L}_{\text{MSE}}$): The pairwise similarity loss from the main paper, enhanced with focal weighting:
	\begin{equation*}
		\mathcal{L}_{\text{MSE}} = \frac{1}{|\mathcal{P}|^2}\sum_{i,j} w_{ij} \cdot |Y_{ij} - \tilde{z}_i^\top \tilde{z}_j|^{\gamma_{ij}} \cdot (Y_{ij} - \tilde{z}_i^\top \tilde{z}_j)^2
	\end{equation*}
	where $\gamma_{ij} = \gamma_{\text{pos}}$ if $Y_{ij} = 1$ and $\gamma_{\text{neg}}$ otherwise. We use $\gamma_{\text{pos}} = 1.0$ and $\gamma_{\text{neg}} = 2.0$, assigning higher weight to negative pairs. This addresses class imbalance in forensic clustering, where manipulated regions are typically small relative to the authentic background. Focal weighting forces the model to focus on hard negative pairs where forensically distinct patches appear similar.
	\item \textbf{Hungarian matching} ($\mathcal{L}_{\text{Hungarian}}$): Combines dice and binary cross-entropy losses between predicted cluster assignments and ground-truth region labels after optimal Hungarian matching, as described in \cref{sec:fkg_sys}. Hungarian matching is necessary because predicted cluster IDs are arbitrary and do not correspond to ground-truth labels.
	\item \textbf{Residual regularization} ($\mathcal{L}_{\text{res}}$): Constrains the FRPN's output embeddings to remain close to the backbone embeddings, providing a continuous gradient signal that complements the output-level residual blend and prevents catastrophic drift from the pretrained representations.
	\item \textbf{Hard boundary crop} ($\mathcal{L}_{\text{crop}}$): Applies an additional loss on 512$\times$512 crops centered on manipulation boundaries, forcing the model to resolve fine-grained boundary transitions at a scale where they are not diluted by the overwhelming majority of authentic patches.
\end{itemize}

\subsection{Training Configuration}

\begin{wraptable}{r}{0.47\linewidth}
	\vspace{-3.4em}
	\centering
	\caption{FRPN training hyperparams.}
	\label{tab:app_frpn_training}
	\adjustbox{max width=\linewidth}{
	\begin{tabular}{@{}lp{20mm}@{}}
		\toprule
		\textbf{Hyperparameter} & \textbf{Value} \\
		\midrule
		Optimizer & AdamW \\
		Learning rate & $2 \times 10^{-4}$ \\
		Weight decay & $1 \times 10^{-2}$ \\
		LR schedule & 0.8 / 2 ep \\
		Batch size & 16 images \\
		Epochs & 30 \\
		Backbone & Frozen \\
		\midrule
		\multicolumn{2}{@{}l@{}}{\textit{Loss weights}} \\
		Focal MSE & 1.0 \\
		Hungarian $\lambda_{\text{H}}$ & 0.05 \\
		Residual $\lambda_{\text{res}}$ & 0.1 \\
		Crop $\lambda_{\text{crop}}$ & 1.0 \\
		\bottomrule
	\end{tabular}
	}
	\vspace{-2em}
\end{wraptable}

The FRPN is trained on manipulated images from FKG-50K with known ground-truth tampering masks. The backbone is frozen during FRPN training, ensuring the pretrained forensic embeddings remain fixed while the FRPN learns to cluster them. This sequential training strategy (backbone first, then FRPN) is deliberate: the backbone's self-supervised forensic fingerprints provide a stable foundation, and the FRPN's role is to discover structure within these embeddings rather than to alter them. Images are divided into 128$\times$128 patches with stride 64 (50\% overlap), and all patches from a single image form one training sample. All hyperparameters are listed in \cref{tab:app_frpn_training}.

\section{Forensic Task Expert Reasoner}
\label{app:training}

\begin{figure}[t]
	\centering
	\includegraphics[width=\linewidth]{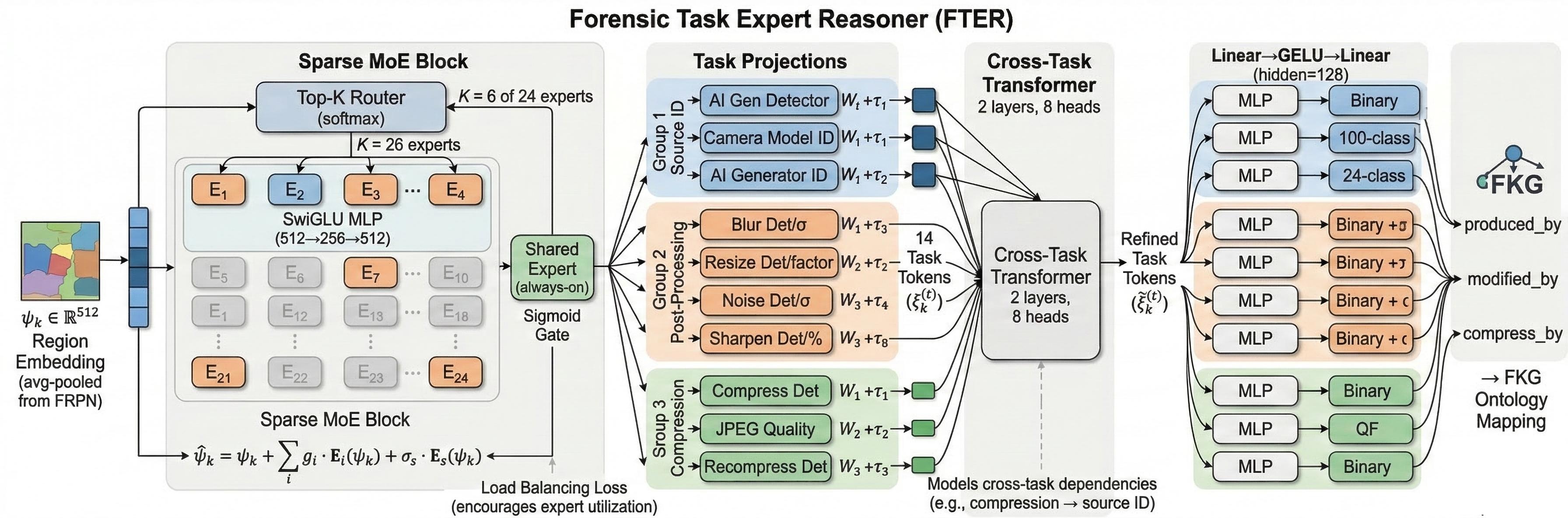}
	\caption{Forensic Task Expert Reasoner (FTER) architecture.}
	\label{fig:fter_arch}
\end{figure}

This appendix provides the full architecture and training details of the component referred to as ``Forensic Task Expert Networks'' and ``Transformer Reasoning Module'' in \cref{sec:fkg_sys}. We collectively name this component the Forensic Task Expert Reasoner (FTER). Given the FRPN's forensic regions, the FTER determines \emph{what happened} to each region by jointly predicting source identity, post-processing operations, and compression history. Rather than using independent MLPs per task, we introduce a Sparse Mixture of Experts (MoE) architecture that allows the model to dynamically allocate capacity across forensic tasks, combined with a transformer reasoning module that models cross-task dependencies.

\subsection{Architecture}

\begin{wraptable}{r}{0.50\linewidth}
	\vspace{-3.5em}
	\centering
	\caption{FTER architecture summary.}
	\label{tab:app_fter_arch}
	\adjustbox{max width=\linewidth}{
	\begin{tabular}{@{}lp{28mm}@{}}
		\toprule
		\textbf{Parameter} & \textbf{Value} \\
		\midrule
		Input dim & 512 \\
		Routed experts & 24 \\
		Active experts (top-$K$) & 6 \\
		Expert hidden dim & 256 \\
		Shared expert & 1 (always-on) \\
		Reasoning layers & 2 \\
		Reasoning heads & 8 \\
		Task head hidden dim & 128 \\
		Task heads & 14 \\
		\bottomrule
	\end{tabular}
	}
	\vspace{-2em}
\end{wraptable}

The FTER takes region-level embeddings $\psi_k$ (average-pooled from the FRPN's refined patch embeddings for each region $k$) and produces forensic attribute predictions that populate the FKG. The architecture has four stages (\cref{tab:app_fter_arch}).

\subheader{Sparse Mixture of Experts}
The core of the FTER is a Sparse MoE block with 24 routed experts and one shared always-on expert. Each expert is a SwiGLU MLP (gate projection, up projection, then down projection with SiLU gating). A learned router selects the top-$K = 6$ experts per region embedding via softmax gating, and routing weights are renormalized to sum to one. The shared expert processes every region regardless of routing decisions, with its contribution modulated by a learned sigmoid gate. The final output combines the routed expert outputs, the shared expert output, and a residual connection from the input.

Given a region embedding $\psi_k$, the router produces softmax probabilities over all $M$ experts and selects the top-$K$. Let $g_i$ denote the renormalized routing weight for selected expert $E_i$, $E_s$ the shared expert, and $\sigma_s$ its learned sigmoid gate. The MoE output is:
\begin{equation*}
	\hat{\psi}_k = \psi_k + \sum_{i \in \text{top-}K} g_i \cdot E_i(\psi_k) + \sigma_s(\psi_k) \cdot E_s(\psi_k)
\end{equation*}

This design is forensically motivated: different forensic tasks require different feature transformations (detecting camera sensor noise is fundamentally different from identifying AI generation artifacts), and sparse routing allows each region to activate only the experts relevant to its forensic profile. The shared expert captures common forensic knowledge that applies across all tasks, such as general compression artifact patterns.

\subheader{Task Projection}
The shared MoE output $\hat{\psi}_k$ is projected into $T = 14$ task-specific embeddings $\xi_k^{(t)} = W_t \hat{\psi}_k$, one per forensic task. This projection allows each task to extract the features most relevant to its prediction from the shared MoE representation.

\subheader{Transformer Reasoning Module}
The $T$ task-specific embeddings are processed by 2 transformer self-attention layers (8 heads each) that model cross-task dependencies. Each task embedding attends to all other task embeddings, allowing forensic attributes to inform one another: compression artifacts inform source identification (recompression suggests manipulation), post-processing traces constrain the set of plausible sources, and the presence of AI-generated content changes the interpretation of all other attributes. The reasoning module allows these dependencies to be learned jointly rather than requiring explicit hand-crafted rules, producing refined task embeddings $\tilde{\xi}_k^{(t)}$.

\subheader{Task Heads}
The refined task embeddings $\tilde{\xi}_k^{(t)}$ are fed to task-specific heads, each a two-layer MLP (hidden dimension 128) producing either classification logits or regression outputs. The 14 tasks are organized by the FKG ontology relations they populate:

\begin{itemize}[leftmargin=5mm, noitemsep, topsep=0pt]
	\item \fkgprop{produced\_by}: AI generation detector (binary), camera model ID (classification over ${\sim}100$ models), AI generator ID (classification over 24 generators)
	\item \fkgprop{modified\_by}: blur, resize, Gaussian noise, and sharpening detectors (binary each) with corresponding parameter regressors (blur $\sigma$, scale factor, noise $\sigma$, sharpening percent)
	\item \fkgprop{compress\_by}: compression detector (binary), JPEG quality regressor, recompression detector (binary)
\end{itemize}

\subsection{Training}

\begin{wraptable}{r}{0.46\linewidth}
	\vspace{-3.5em}
	\centering
	\caption{FTER training hyperparameters.}
	\label{tab:app_fter_training}
	\adjustbox{max width=\linewidth}{
	\begin{tabular}{@{}lp{20mm}@{}}
		\toprule
		\textbf{Hyperparameter} & \textbf{Value} \\
		\midrule
		Optimizer & AdamW \\
		Learning rate & $1 \times 10^{-3}$ \\
		Weight decay & $1 \times 10^{-2}$ \\
		LR schedule & 0.8 / 4 ep \\
		Minimum LR & $1 \times 10^{-6}$ \\
		Batch size & 256 patches \\
		Epochs & 100 \\
		Backbone, FRPN & Frozen \\
		\midrule
		\multicolumn{2}{@{}l@{}}{\textit{Loss weights}} \\
		Load balancing & 0.001 \\
		\bottomrule
	\end{tabular}
	}
	\vspace{-2em}
\end{wraptable}

The FTER is trained on the full MUP corpus (real images) combined with synthetic images, with the backbone and FRPN frozen. Each task head is trained with cross-entropy (classification) or MSE (regression). A key design choice handles the asymmetry between real and synthetic samples: real images have camera model labels but no generator labels, while synthetic images have generator labels but no camera labels. Rather than training separate models, we use NaN-padding: missing labels are set to NaN, and the per-task loss ignores NaN entries. This allows mixed batches of real and synthetic samples, enabling the shared MoE experts and reasoning module to learn joint representations across both domains.

An auxiliary load balancing loss~\cite{fedus2022switch} encourages balanced expert utilization, preventing expert collapse. Let $f_j$ be the fraction of tokens routed to expert $j$ and $p_j$ the mean router probability assigned to expert $j$, both computed over a batch. The loss is:
\begin{equation*}
	\mathcal{L}_{\text{bal}} = M \cdot \sum_{j=1}^{M} f_j \cdot p_j
\end{equation*}
where $M$ is the number of routed experts. This product is minimized when both token assignments and routing probabilities are uniform across experts. All hyperparameters are listed in \cref{tab:app_fter_training}.

The complete training pipeline is sequential: backbone (\cref{app:backbone}) $\rightarrow$ FRPN (\cref{app:frpn}) $\rightarrow$ FTER. Each stage freezes the preceding components, ensuring that downstream learning does not corrupt the representations learned upstream.

\section{FKG-50K Dataset Details}
\label{app:dataset}

This appendix provides the full construction details of the FKG-50K dataset introduced in \cref{sec:datasets}. FKG-50K contains 50,000 images with complete provenance and manipulation histories encoded as ground-truth Forensic Knowledge Graphs. The dataset is constructed through an automated data engine that generates four major classes of inauthentic imagery, each paired with its pristine camera-original counterpart and a fully annotated FKG.

\subsection{Source Corpus and Metadata Extraction}

The data engine begins with a media pool of pristine, unedited photographs captured by real cameras. Images are sourced from Unsplash~\cite{unsplash}, Pexels~\cite{pexels}, and crowdsourced collections, filtered to ensure they are camera-original (\ie, not previously edited or recompressed by social media platforms). Each image is required to contain reliable EXIF fields describing:

\begin{itemize}[leftmargin=5mm, noitemsep, topsep=0pt]
	\item Camera make and model (\eg, Nikon Coolpix S33, Canon EOS R6)
	\item Color pipeline and white balance parameters
	\item Compression format and quality factor
	\item Image dimensions and orientation
\end{itemize}

Images lacking any of these fields are discarded, as they cannot provide the source-level forensic fingerprint the FKG ontology requires. FKG-50K is built from a curated subset of a larger pretraining corpus (described in \cref{app:backbone}) that contains over 350 camera models. For FKG-50K, we select images from 72 camera models across 17 manufacturers whose provenance is fully trusted, including smartphones (Apple, Samsung, LG, HTC, Huawei, Motorola, Nokia, OnePlus, Xiaomi, BlackBerry), DSLRs and mirrorless cameras (Canon, Nikon, Sony, Fujifilm, Olympus), and compact cameras (Kodak, Panasonic, Pentax). Each retained image is additionally processed to extract a detailed scene caption using a VLM, specifying all objects, attributes, and spatial relationships present. This caption guides subsequent manipulation proposals to ensure semantic coherence.

\subsection{Edit Proposal Generation}

The core of the data engine is a proposal module that, given a target forgery type, samples a structured set of attributes describing how the forgery should be created. Proposals are generated in XML format by a VLM so they can be deterministically parsed and validated. Below we describe the proposal structure for each forgery type.

\subheader{Splicing Proposals}
For spliced images, the proposal specifies: (1) a host image, (2) a donor image from a different camera source, (3) the donor object to extract, and (4) the target location in the host image. The VLM is prompted with both images to select plausible objects and placements consistent with scene geometry. Using donors from different cameras ensures that the spliced region carries a distinct forensic fingerprint from the host.

\subheader{Traditional Editing Proposals}
For traditional editing operations (blur, noise addition, sharpening, resampling), the proposal specifies: (1) the source image, (2) the object or region to edit, and (3) the editing operation type with parameters. Semantic regions are selected by prompting the VLM with queries about salient objects, and operation types and parameters are sampled from a predefined library covering GaussianBlur ($\sigma \in [0.5, 5.0]$), GaussianNoise ($\sigma \in [1, 25]$), Sharpening (strength $\in [0.5, 3.0]$), and Resampling (scale $\in [0.5, 2.0]$).

\subheader{Fully Synthetic Proposals}
For fully synthetic images, the engine generates: (1) the original real image as reference, (2) the generation method (SDXL~\cite{sdxl}, FLUX~\cite{flux}, Midjourney v6~\cite{Midjourney}, or GPT-Image-1~\cite{gptimage1}), and (3) positive and negative prompts derived from the detailed caption of the original image. This ensures the synthetic variant is semantically aligned with its real counterpart, making the detection task realistic.

\subheader{AI-Editing Proposals}
For AI-based editing, the proposal specifies: (1) the source image, (2) the editing model (FLUX-Kontex-1~\cite{flux_kontex} or SDXL-Inpaint~\cite{sdxl}), (3) the object or region to edit, and (4) positive and negative editing prompts describing the intended modification (object replacement, insertion, or removal).

\subsection{Mask Generation}

For splicing, traditional editing, and AI-editing, the proposal is accompanied by a pixel-level mask specifying the edited region. The mask is generated by feeding the VLM-supplied region description into SAM 3~\cite{carion2025sam}, which accepts text prompts directly and produces a segmentation mask aligned with the described object or region. Fully synthetic images do not require a mask because their entire content originates from a non-camera source.

\subsection{Edit Execution}

Given the proposal and editing mask, the engine constructs a complete edit specification and executes it:

\begin{itemize}[leftmargin=5mm, noitemsep, topsep=0pt]
	\item \textbf{Splicing}: The donor region is extracted from the donor image using the mask and composited into the host image at the specified location using Poisson blending for seamless boundaries.
	\item \textbf{Traditional editing}: The specified operation is applied to the masked region with the sampled parameters.
	\item \textbf{Fully synthetic generation}: The AI model generates a full-frame image conditioned on the prompts derived from the real image.
	\item \textbf{AI editing}: The masked region is modified via the specified AI inpainting/editing model conditioned on the proposed prompts.
\end{itemize}

Each output image is saved alongside its proposal metadata, mask, and compression parameters.

\subsection{Automated Verification}

The engine performs a verification pass using a VLM to ensure that each generated image is consistent with its proposal. For splicing, traditional editing, and AI-editing, the VLM checks that the edited region corresponds to the intended object and that the visual effect matches the specified operation. For fully synthetic images, the VLM verifies that the generated content is semantically aligned with the prompts. If verification fails, the proposal is discarded and a new one is automatically sampled.

\subsection{Ground-Truth FKG Construction}

For each image, the engine emits a complete ground-truth FKG conforming to the ontology defined in \cref{sec:fkg} and \cref{app:ontology}. The FKG is constructed deterministically from the pipeline metadata:

\begin{itemize}[leftmargin=5mm, noitemsep, topsep=0pt]
	\item \textbf{Region nodes}: derived from the editing mask (manipulated vs.\ authentic regions). Authentic images have a single region.
	\item \textbf{Source attribution}: each region's \fkgprop{produced\_by} is set to the camera model (from EXIF) or AI generator (from the proposal).
	\item \textbf{Post-processing}: each region's \fkgprop{modified\_by} records the applied operation and parameters, or NoPostProc if unmodified.
	\item \textbf{Compression}: each region's \fkgprop{compress\_by} is set based on the image's compression history. Authentic regions from previously compressed sources are marked as ReCompressed after manipulation.
	\item \textbf{Content}: each region's \fkgprop{depict} records objectedness and object description from the VLM caption.
\end{itemize}

This deterministic construction ensures that every ground-truth FKG is complete, machine-verifiable, and directly usable for evaluating FKG prediction accuracy (\eg, the perfect match metric in \cref{sec:discussion}).

\subsection{Dataset Composition}

\begin{wraptable}{r}{0.4\linewidth}
	\vspace{-3.4em}
	\centering
	\caption{FKG-50K dataset composition.}
	\label{tab:app_dataset_composition}
	\adjustbox{max width=\linewidth}{
	\begin{tabular}{@{}lcc@{}}
		\toprule
		\textbf{Category} & \textbf{Train} & \textbf{Test} \\
		\midrule
		Authentic          & 8,000  & 2,000  \\
		Splicing           & 8,000  & 2,000  \\
		Traditional Editing & 8,000  & 2,000  \\
		AI Editing          & 8,000  & 2,000  \\
		Full Synthesis      & 8,000  & 2,000  \\
		\midrule
		\textbf{Total}     & \textbf{40,000} & \textbf{10,000} \\
		\bottomrule
	\end{tabular}
	}
	\vspace{-2em}
\end{wraptable}

The dataset is split into 40,000 training and 10,000 evaluation samples (\cref{tab:app_dataset_composition}), with each forgery type equally represented. The authentic subset serves as the negative class across all experiments. Each sample includes the image, its ground-truth FKG, and the pixel-level manipulation mask (for non-authentic images).

\subheader{Evaluation Protocol}
Our system emits an FKG for any input image, so FKG-style ground-truth annotations are never required at test time. On out-of-distribution datasets, which provide only conventional labels, we simply score the predicted FKG attributes against whatever each dataset provides, such as tampering masks (for localization F1) or real/AI labels (for detection ACC/AUC).

The fully synthetic subset draws from 24 AI generators spanning GANs (ProGAN, StyleGAN, StyleGAN2, StyleGAN3, BigGAN, CycleGAN, StarGAN, ProjectedGAN, GigaGAN, EG3D), diffusion models (Guided Diffusion, Latent Diffusion, Stable Diffusion 1.5, Stable Diffusion 3 Medium, SDXL, FLUX, GLIDE), and commercial systems (DALL-E 2, DALL-E 3, DALL-E Mini, Midjourney v6, GPT-Image-1). The AI-editing subset uses FLUX-Kontex-1 and SDXL-Inpaint. This diversity ensures that the dataset covers a broad range of generation architectures and forensic fingerprint characteristics.

\section{VLM Prompts \& Response Schema}
\label{app:prompts}

This appendix provides the full prompt templates and response schema referenced in \cref{sec:interp_fkg} and \cref{sec:experiments}.

\subsection{FKG Interpretation Prompt}

The following prompt is used to instruct the VLM to interpret a serialized FKG and produce a forensic report. The prompt consists of three components: task instructions, the serialized FKG triplets with Set-of-Mark references, and the response schema.

\begin{codebox}{blue}
You are a forensic image analyst. You are given an image with
numbered region marks overlaid, along with a set of forensic
analysis results encoded as subject-predicate-object triplets.

Your task is to:
1. Determine whether the image is AUTHENTIC or MANIPULATED.
2. If manipulated, identify the type of manipulation
   (SPLICING, AI_EDITING, TRADITIONAL_EDITING, or
   FULL_SYNTHESIS).
3. Identify which regions are manipulated and what they
   depict.
4. Provide a forensic justification grounded ONLY in the
   provided triplets. Do NOT speculate beyond the evidence.

Forensic evidence triplets:
{serialized_fkg_triplets}

Respond using the following XML schema EXACTLY.
\end{codebox}

\subsection{Response Schema}

The VLM must produce its output in the following structured XML format, ensuring consistency and machine readability.

\begin{codebox}{blue}
<report>
  <summary>
    [1-2 sentence summary of the forensic findings.]
  </summary>
  <decision>AUTHENTIC | MANIPULATED</decision>
  <manipulation>
    <!-- Omitted if decision = AUTHENTIC -->
    <type>SPLICING | AI_EDITING | TRADITIONAL_EDITING
          | FULL_SYNTHESIS</type>
    <instances>
      <instance>
        <region>[Region mark number]</region>
        <object>[What the region depicts]</object>
        <source>[Source attribution, e.g., camera model
                 or AI generator]</source>
      </instance>
      <!-- Additional instances as needed -->
    </instances>
  </manipulation>
  <justification>
    [Detailed forensic justification referencing specific
     triplets from the provided evidence. Each claim must
     cite the supporting triplet.]
  </justification>
</report>
\end{codebox}

\subsection{VLM Evaluation Prompts}

For the competing VLM evaluations in \cref{subsec:exp1,subsec:exp2,subsec:exp3}, we use standardized prompts for each experiment to ensure fair comparison. All VLMs receive the same image and prompt. No forensic evidence or FKG triplets are provided to competing VLMs.

\subheader{Experiment 1 --- Detection}
\begin{codebox}{blue}
Examine this image carefully. Is this image authentic
(unaltered) or has it been manipulated/falsified in any way?
This includes AI-generated content, splicing, editing, or
any form of digital manipulation.

Respond using the following XML schema EXACTLY:
<report>
  <decision>AUTHENTIC | MANIPULATED</decision>
</report>
\end{codebox}

\subheader{Experiment 2 --- Type Classification \& Localization}

\noindent
Without oracle access:
\begin{codebox}{blue}
Examine this image carefully. Determine:
1. Whether this image is authentic or manipulated.
2. If manipulated, classify the type as one of:
   SPLICING, AI_EDITING, TRADITIONAL_EDITING, FULL_SYNTHESIS.
3. Describe the location and content of any manipulated
   regions.

Respond using the following XML schema EXACTLY:
<report>
  <decision>AUTHENTIC | MANIPULATED</decision>
  <manipulation>
    <!-- Omit if AUTHENTIC -->
    <type>SPLICING | AI_EDITING | TRADITIONAL_EDITING
          | FULL_SYNTHESIS</type>
    <regions>[description of manipulated regions]</regions>
  </manipulation>
</report>
\end{codebox}

\noindent
With oracle access:
\begin{codebox}{blue}
This image has been confirmed as MANIPULATED. Determine:
1. The type of manipulation:
   SPLICING, AI_EDITING, TRADITIONAL_EDITING, FULL_SYNTHESIS.
2. The location and content of the manipulated regions.

Respond using the following XML schema EXACTLY:
<report>
  <type>SPLICING | AI_EDITING | TRADITIONAL_EDITING
        | FULL_SYNTHESIS</type>
  <regions>[description of manipulated regions]</regions>
</report>
\end{codebox}

\subheader{Experiment 3 --- Forensic Justification}

\noindent
With oracle access (ground-truth labels provided):
\begin{codebox}{blue}
This image has been confirmed as MANIPULATED.
Manipulation type: {ground_truth_type}
Manipulated regions: {ground_truth_regions}

Provide a detailed forensic justification explaining what
forensic evidence supports this conclusion, how the
manipulated regions differ from authentic regions, and any
compression, source, or processing inconsistencies.
Be specific and reference concrete forensic traces.

Respond using the following XML schema EXACTLY:
<report>
  <justification>
    [Detailed forensic justification referencing specific
     evidence. Each claim must cite supporting traces.]
  </justification>
</report>
\end{codebox}

\section{Computational Costs \& ICR Convergence}
\label{app:compute}

This appendix details the computational requirements for training and inference, and provides a convergence analysis for Iterative Context Refinement (ICR).

\subsection{Model Sizes}

\begin{wraptable}{r}{0.38\linewidth}
	\vspace{-3.4em}
	\centering
	\caption{Parameter counts by component.}
	\label{tab:param_counts}
	\adjustbox{max width=\linewidth}{
	\begin{tabular}{@{}lr@{}}
		\toprule
		\textbf{Component} & \textbf{Parameters} \\
		\midrule
		Backbone (MISLNet)         & 0.8M \\
		FRPN                       & 7.4M \\
		FTER (MoE + task heads)    & 16.1M \\
		\quad of which MoE         & 9.8M \\
		\midrule
		\textbf{Total (trainable)} & \textbf{24.3M} \\
		\bottomrule
	\end{tabular}
	}
	\vspace{-2em}
\end{wraptable}

\cref{tab:param_counts} summarizes the parameter counts for each component of the FKG generation system. The backbone is deliberately lightweight (848K parameters), consisting of a constrained convolutional layer followed by four convolutional blocks and three fully connected layers. Despite its small size, it produces expressive 512-dimensional forensic embeddings because it learns from imaging pipeline properties rather than semantic content.

The FRPN accounts for 7.4M parameters from 16 alternating attention blocks (8 global self-attention + 8 local graph attention), each with pre-norm residual connections, LayerScale, and a compact MLP (ratio 0.5). The FTER contains 16.1M parameters, of which 9.8M reside in the sparse MoE block (24 routed experts + 1 shared expert). The remainder covers 14 task projection layers, a 2-layer Transformer reasoning module, and 14 task-specific prediction heads.

The VLM used for justification is Qwen3-VL-8B~\cite{bai2025qwen3vl} in a 4-bit quantized configuration, adding approximately 4GB of memory. This compact VLM can run on consumer GPUs and edge devices, including smartphones, making the full FKG pipeline deployable without cloud infrastructure. The total trainable parameter count of 24.3M (excluding the frozen VLM) remains compact compared to end-to-end forensic models that rely on large pretrained vision encoders (\eg, HiFi-IFDL uses a multi-branch HRNet with over 60M parameters).

\subsection{Inference Costs}

\begin{wraptable}{r}{0.42\linewidth}
	\vspace{-3.4em}
	\centering
	\caption{Per-image inference latency.}
	\label{tab:inference_costs}
	\adjustbox{max width=\linewidth}{
	\begin{tabular}{@{}lr@{}}
		\toprule
		\textbf{Stage} & \textbf{Latency} \\
		\midrule
		Patch extraction      & $\sim$30ms  \\ % TODO: verify
		Backbone embedding    & $\sim$120ms \\ % TODO: verify
		FRPN clustering       & $\sim$220ms \\ % TODO: verify
		FTER task prediction  & $\sim$70ms  \\ % TODO: verify
		FKG construction      & $\sim$10ms  \\
		\midrule
		\textbf{FKG generation total} & $\sim$\textbf{0.45s} \\
		\midrule
		VLM justification & $\sim$0.7--1.2s \\
		\midrule
		\textbf{End-to-end total}     & $\sim$\textbf{1.2--1.7s} \\
		\bottomrule
	\end{tabular}
	}
	\vspace{-2em}
\end{wraptable}

% TODO: verify per-stage breakdown against profiler output; total ~450ms on 4K image is confirmed
\cref{tab:inference_costs} reports per-image inference latency on a single NVIDIA A100 GPU for a $3840 \times 2160$ (4K) input image. The FKG generation pipeline (backbone through FKG construction) completes in approximately 0.45 seconds, with FRPN clustering as the primary bottleneck due to the $O(N^2)$ pairwise similarity computation over $N$ patches.

The VLM justification step adds 0.7--1.2 seconds using Qwen3-VL-8B (4-bit), requiring a forward pass with the SoM-annotated image and serialized FKG triplets. The total end-to-end pipeline, from raw image to justified forensic report, completes in under 2 seconds on a single GPU. Because the quantized VLM requires only 4GB of memory, the entire pipeline can run on consumer hardware or edge devices without cloud dependencies. For applications that need only detection, localization, or type classification, the sub-second FKG generation is sufficient without invoking the VLM.

\subsection{ICR Convergence Analysis}

% TODO: verify judge LM choice, agreement %, N, epsilon, iteration counts, and convergence table values
\subheader{Judge LM Specification}
ICR requires a judge LM to evaluate the completeness and correctness metrics defined in \cref{eq:metrics}. The judge receives the ground-truth FKG triplets and the VLM-generated report, then evaluates each triplet for mention ($\tau$) and correctness ($Q$). We evaluated four judge LMs: GPT-4o~\cite{openai2024gpt4o}, GPT-5~\cite{openai2025gpt5}, Qwen3-VL-8B~\cite{bai2025qwen3vl}, and Gemma 3~\cite{gemmateam2025gemma3}. All four produce highly similar sets of corrective examples, indicating that the ICR outcome is robust to judge choice and not an artifact of any particular model's biases. We default to GPT-4o for its balance of cost and reliability.

\subheader{Convergence Behavior}
% TODO: verify iteration-level numbers against logs
We run ICR on a training set of $N = 1{,}000$ FKG examples (100 authentic + 900 manipulated). \cref{tab:icr_convergence} shows how the metrics evolve across iterations on this training set. ICR converges to near-perfect correctness and completeness within 5 iterations, accumulating 50 corrective demonstrations in $\mathcal{C}$ (8 authentic + 42 manipulated). The largest gains occur in the first two iterations, where ICR identifies the VLM's most severe failure modes: omitting compression and post-processing evidence (iteration 1) and fabricating source attributions for AI-edited content (iteration 2). Subsequent iterations address progressively rarer edge cases until both metrics saturate.

\begin{table}[h]
	\centering
	\caption{ICR convergence across iterations on the training set ($N = 1{,}000$, $\epsilon = 0.01$).}
	\label{tab:icr_convergence}
	\begin{tabular}{@{}ccccc@{}}
		\toprule
		\textbf{Iteration} & \textbf{Correctness} & \textbf{Completeness} & \textbf{$\Delta$ (avg)} & \textbf{Examples in $\mathcal{C}$} \\
		\midrule
		0 (no context) & 0.64 & 0.32 & --    & 0  \\
		1              & 0.78 & 0.65 & 0.235 & 12 \\
		2              & 0.89 & 0.84 & 0.150 & 25 \\
		3              & 0.96 & 0.95 & 0.090 & 38 \\
		4              & 0.99 & 0.99 & 0.035 & 47 \\
		5              & 1.00 & 1.00 & 0.005 & 50 \\
		\bottomrule
	\end{tabular}
\end{table}

At convergence, $\mathcal{C}$ contains 50 corrective demonstrations (8 authentic, 42 manipulated) covering the full range of VLM failure modes encountered on the training set. Note that these metrics measure how faithfully the VLM reports ground-truth FKG evidence, whereas the correctness and completeness in \cref{tab:disc_prompt_opt} are evaluated on predicted FKGs, making the two sets of numbers not directly comparable. Completeness improves more dramatically than correctness because the VLM's primary failure mode is omission rather than fabrication: without corrective examples, it reports only the most visually salient evidence and ignores subtle forensic properties like compression history and post-processing traces.
\section{Qualitative Examples}
\label{app:examples}

This appendix presents qualitative examples from each manipulation category in FKG-50K, as well as examples from two external datasets (DSO-1 and CASIAv2). Each table shows the input image, predicted localization mask, the forensic decision produced by the FKG, and the VLM-generated justification. For splicing, traditional editing, and AI editing, the ground-truth mask is also shown. Incorrect decisions are highlighted in \red{red}. All predictions are from our full system with ICR-optimized prompts.

\smallskip\noindent\textbf{Contents:}
\begin{itemize}[nosep,leftmargin=*]
	\item \hyperref[tab:qual_authentic]{Authentic} (FKG-50K) — \cref{tab:qual_authentic}
	\item \hyperref[tab:qual_splicing]{Splicing} (FKG-50K) — \cref{tab:qual_splicing}
	\item \hyperref[tab:qual_traditional_editing]{Traditional Editing} (FKG-50K) — \cref{tab:qual_traditional_editing}
	\item \hyperref[tab:qual_ai_editing]{AI Editing} (FKG-50K) — \cref{tab:qual_ai_editing}
	\item \hyperref[tab:qual_full_synthesis]{Full Synthesis} (FKG-50K) — \cref{tab:qual_full_synthesis}
	\item \hyperref[tab:qual_dso-1]{DSO-1} (external) — \cref{tab:qual_dso-1}
	\item \hyperref[tab:qual_casiav2]{CASIAv2} (external) — \cref{tab:qual_casiav2}
\end{itemize}

% Helper: image thumbnail inside table cell
\newcommand{\timg}[1]{\includegraphics[width=\linewidth,height=10mm,keepaspectratio]{#1}}

% Auto-generated by prepare_qualitative.py — do not edit by hand.

\begin{table}[p]
\centering
\caption{Qualitative examples from FKG-50K: \textbf{Authentic}. These are unmanipulated camera-original photographs. The predicted mask is blank (all zeros) because the system correctly identifies a single region with consistent source and processing throughout.}
\label{tab:qual_authentic}
\smallerr
\setlength{\tabcolsep}{2pt}
\renewcommand{\arraystretch}{1.1}
\begin{tabular}{
	>{\centering\arraybackslash}m{17mm}
	>{\centering\arraybackslash}m{17mm}
	>{\noindent\justifying\arraybackslash}m{24mm}
	>{\noindent\justifying\arraybackslash}m{55mm}
}
\toprule
\textbf{Image} & \textbf{Pred Mask} & \textbf{Decision} & \textbf{Justification} \\
\midrule
\rowcolor{gray!30}
\timg{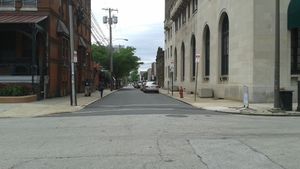} &
\timg{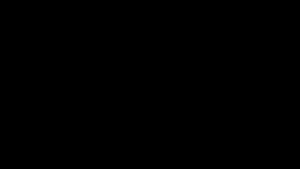} &
This image is authentic , no signs of tampering were found. &
Forensic examination finds a single, uniform region across the image, showing no signs of splicing, editing, or AI generation. The image traces back to a Canon EOS M50, the image has been JPEG compressed at quality 91. \\
\timg{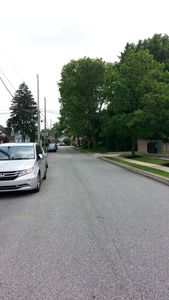} &
\timg{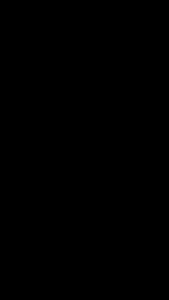} &
Forensic analysis confirms this image is genuine and has not been tampered with. &
The image presents a single coherent forensic profile throughout, ruling out splicing or partial editing. It was identified as having been taken with a iPhone X. The image has been JPEG compressed at quality 90. \\
\rowcolor{gray!30}
\timg{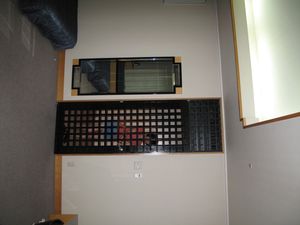} &
\timg{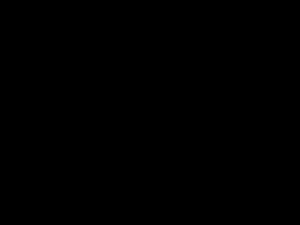} &
The image shows no evidence of forgery or manipulation. &
The entire image exhibits consistent forensic characteristics throughout, with no detectable boundaries or inconsistencies that would suggest tampering. The forensic signature is consistent with a Canon EOS 5D Mark II, and the image has been JPEG compressed at quality 92. \\
\timg{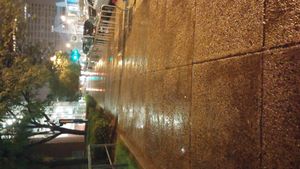} &
\timg{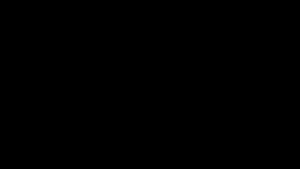} &
Analysis indicates this image is a genuine, untampered photograph. &
All parts of the image share the same forensic fingerprint, with no discontinuities that would indicate manipulation. The content is consistent with having been captured by a iPhone X. The image has been JPEG compressed at quality 94. \\
\rowcolor{gray!30}
\timg{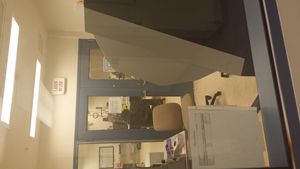} &
\timg{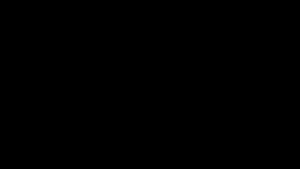} &
No manipulation detected. This image appears to be an unaltered photograph. &
No forensic anomalies were detected anywhere in the image. The uniform characteristics point to a Nikon D3300 as the source device. The image has been JPEG compressed at quality 92. \\
\timg{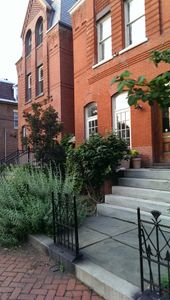} &
\timg{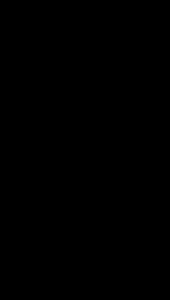} &
This is an authentic, unmodified photograph. &
Forensic examination finds a single, uniform region across the image, showing no signs of splicing, editing, or AI generation. The image traces back to a iPhone 5s, the image has been JPEG compressed at quality 94. \\
\bottomrule
\end{tabular}
\end{table}

\begin{table}[p]
\centering
\caption{Qualitative examples from FKG-50K: \textbf{Authentic}. These are unmanipulated camera-original photographs. The predicted mask is blank (all zeros) because the system correctly identifies a single region with consistent source and processing throughout. (cont.)}
\smallerr
\setlength{\tabcolsep}{2pt}
\renewcommand{\arraystretch}{1.1}
\begin{tabular}{
	>{\centering\arraybackslash}m{17mm}
	>{\centering\arraybackslash}m{17mm}
	>{\noindent\justifying\arraybackslash}m{24mm}
	>{\noindent\justifying\arraybackslash}m{55mm}
}
\toprule
\textbf{Image} & \textbf{Pred Mask} & \textbf{Decision} & \textbf{Justification} \\
\midrule
\rowcolor{gray!30}
\timg{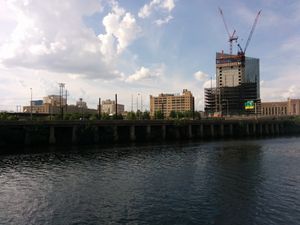} &
\timg{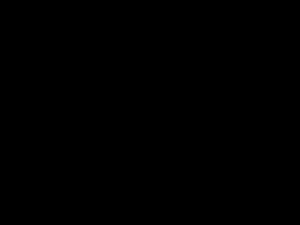} &
This photograph has not been altered , it is authentic. &
The image presents a single coherent forensic profile throughout, ruling out splicing or partial editing. It was identified as having been taken with a iPhone 5s. The image has been JPEG compressed at quality 94. \\
\timg{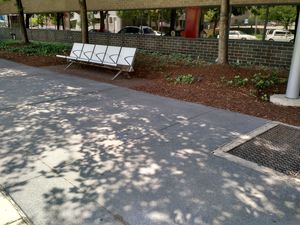} &
\timg{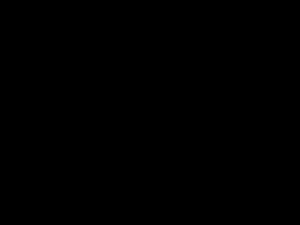} &
This image is authentic , no signs of tampering were found. &
The entire image exhibits consistent forensic characteristics throughout, with no detectable boundaries or inconsistencies that would suggest tampering. The forensic signature is consistent with a iPhone 5s, and the image has been JPEG compressed at quality 94. \\
\rowcolor{gray!30}
\timg{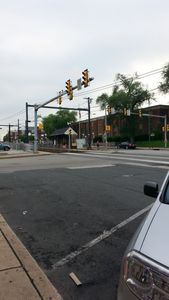} &
\timg{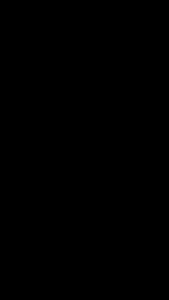} &
Forensic analysis confirms this image is genuine and has not been tampered with. &
All parts of the image share the same forensic fingerprint, with no discontinuities that would indicate manipulation. The image has been JPEG compressed at quality 94. \\
\timg{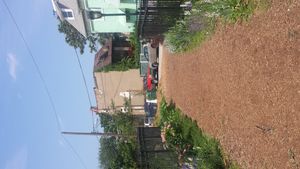} &
\timg{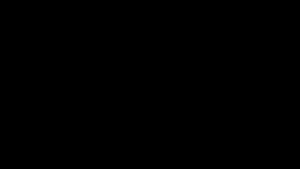} &
The image shows no evidence of forgery or manipulation. &
No forensic anomalies were detected anywhere in the image. The uniform characteristics point to a Nikon D3300 as the source device. The image has been JPEG compressed at quality 91. \\
\rowcolor{gray!30}
\timg{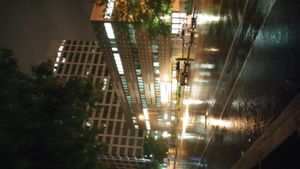} &
\timg{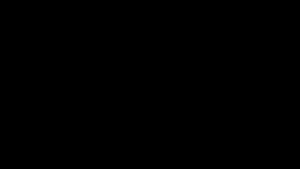} &
Analysis indicates this image is a genuine, untampered photograph. &
Forensic examination finds a single, uniform region across the image, showing no signs of splicing, editing, or AI generation. The image traces back to a Canon EOS 80D, the image has been JPEG compressed at quality 94. \\
\timg{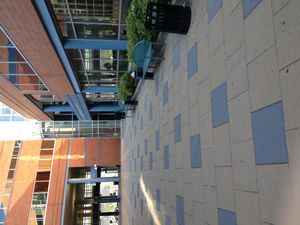} &
\timg{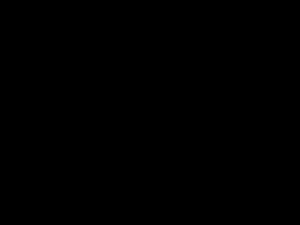} &
No manipulation detected. This image appears to be an unaltered photograph. &
The image presents a single coherent forensic profile throughout, ruling out splicing or partial editing. It was identified as having been taken with a iPhone 5. The image has been JPEG compressed at quality 94. \\
\bottomrule
\end{tabular}
\end{table}

\begin{table}[p]
\centering
\caption{Qualitative examples from FKG-50K: \textbf{Authentic}. These are unmanipulated camera-original photographs. The predicted mask is blank (all zeros) because the system correctly identifies a single region with consistent source and processing throughout. (cont.)}
\smallerr
\setlength{\tabcolsep}{2pt}
\renewcommand{\arraystretch}{1.1}
\begin{tabular}{
	>{\centering\arraybackslash}m{17mm}
	>{\centering\arraybackslash}m{17mm}
	>{\noindent\justifying\arraybackslash}m{24mm}
	>{\noindent\justifying\arraybackslash}m{55mm}
}
\toprule
\textbf{Image} & \textbf{Pred Mask} & \textbf{Decision} & \textbf{Justification} \\
\midrule
\rowcolor{gray!30}
\timg{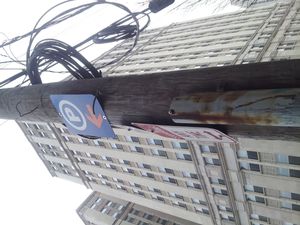} &
\timg{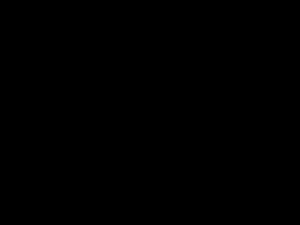} &
This is an authentic, unmodified photograph. &
The entire image exhibits consistent forensic characteristics throughout, with no detectable boundaries or inconsistencies that would suggest tampering. The forensic signature is consistent with a Canon EOS M50, and the image has been JPEG compressed at quality 87. \\
\timg{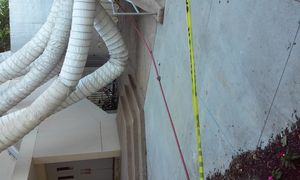} &
\timg{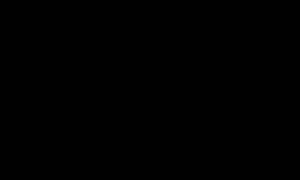} &
This photograph has not been altered , it is authentic. &
All parts of the image share the same forensic fingerprint, with no discontinuities that would indicate manipulation. The content is consistent with having been captured by a iPhone 5s. The image has been JPEG compressed at quality 94. \\
\rowcolor{gray!30}
\timg{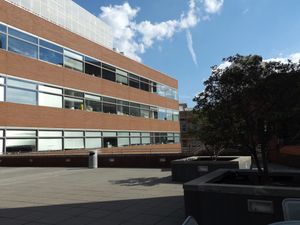} &
\timg{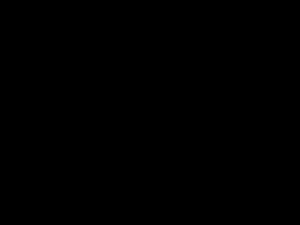} &
This image is authentic , no signs of tampering were found. &
No forensic anomalies were detected anywhere in the image. The uniform characteristics point to a iPhone 5s as the source device. The image has been JPEG compressed at quality 89. \\
\timg{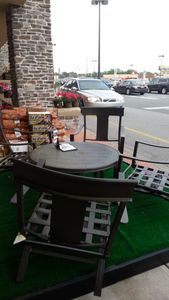} &
\timg{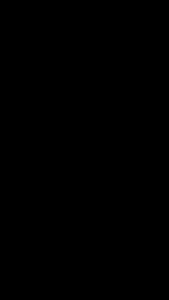} &
Forensic analysis confirms this image is genuine and has not been tampered with. &
Forensic examination finds a single, uniform region across the image, showing no signs of splicing, editing, or AI generation. The image traces back to a iPhone 5s, the image has been JPEG compressed at quality 91. \\
\rowcolor{gray!30}
\timg{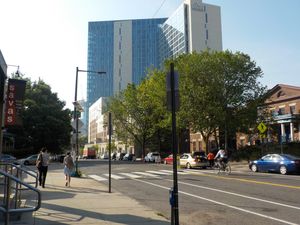} &
\timg{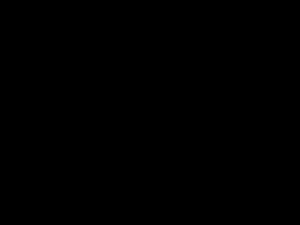} &
The image shows no evidence of forgery or manipulation. &
The image presents a single coherent forensic profile throughout, ruling out splicing or partial editing. The image has been JPEG compressed at quality 89. \\
\timg{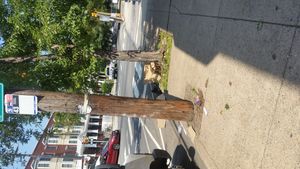} &
\timg{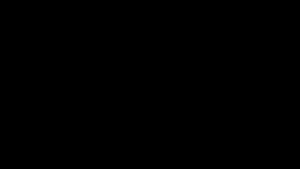} &
Analysis indicates this image is a genuine, untampered photograph. &
The entire image exhibits consistent forensic characteristics throughout, with no detectable boundaries or inconsistencies that would suggest tampering. The forensic signature is consistent with a iPhone 5s, and the image has been JPEG compressed at quality 94. \\
\bottomrule
\end{tabular}
\end{table}

\begin{table}[p]
\centering
\caption{Qualitative examples from FKG-50K: \textbf{Authentic}. These are unmanipulated camera-original photographs. The predicted mask is blank (all zeros) because the system correctly identifies a single region with consistent source and processing throughout. (cont.)}
\smallerr
\setlength{\tabcolsep}{2pt}
\renewcommand{\arraystretch}{1.1}
\begin{tabular}{
	>{\centering\arraybackslash}m{17mm}
	>{\centering\arraybackslash}m{17mm}
	>{\noindent\justifying\arraybackslash}m{24mm}
	>{\noindent\justifying\arraybackslash}m{55mm}
}
\toprule
\textbf{Image} & \textbf{Pred Mask} & \textbf{Decision} & \textbf{Justification} \\
\midrule
\rowcolor{gray!30}
\timg{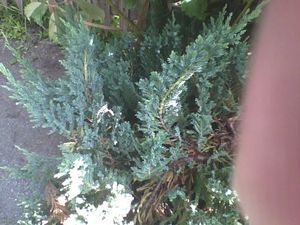} &
\timg{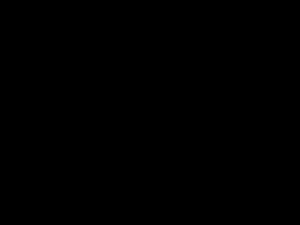} &
No manipulation detected. This image appears to be an unaltered photograph. &
All parts of the image share the same forensic fingerprint, with no discontinuities that would indicate manipulation. The content is consistent with having been captured by a Sony A6000. The image has been JPEG compressed at quality 90. \\
\timg{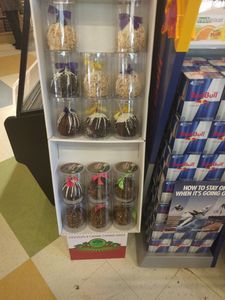} &
\timg{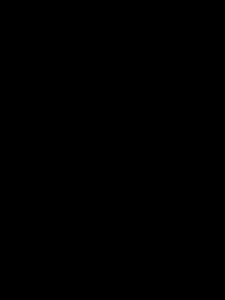} &
This is an authentic, unmodified photograph. &
No forensic anomalies were detected anywhere in the image. The uniform characteristics point to a iPhone 5s as the source device. The image has been JPEG compressed at quality 90. \\
\rowcolor{gray!30}
\timg{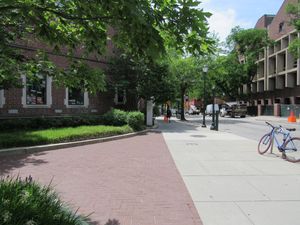} &
\timg{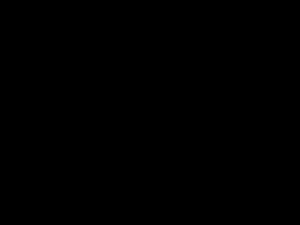} &
This photograph has not been altered , it is authentic. &
Forensic examination finds a single, uniform region across the image, showing no signs of splicing, editing, or AI generation. The image traces back to a Canon EOS M50, the image has been JPEG compressed at quality 92. \\
\timg{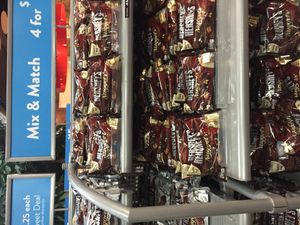} &
\timg{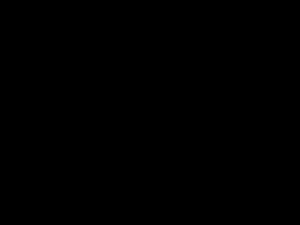} &
This image is authentic , no signs of tampering were found. &
The image presents a single coherent forensic profile throughout, ruling out splicing or partial editing. It was identified as having been taken with a iPhone 6. The image has been JPEG compressed at quality 93. \\
\rowcolor{gray!30}
\timg{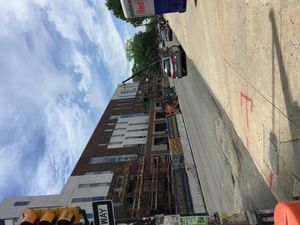} &
\timg{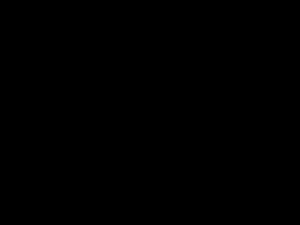} &
Forensic analysis confirms this image is genuine and has not been tampered with. &
The entire image exhibits consistent forensic characteristics throughout, with no detectable boundaries or inconsistencies that would suggest tampering. The forensic signature is consistent with a iPhone 6, and the image has been JPEG compressed at quality 93. \\
\timg{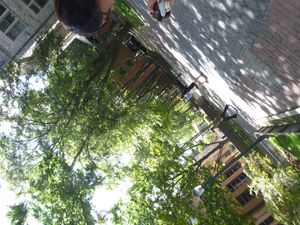} &
\timg{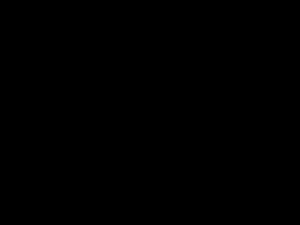} &
The image shows no evidence of forgery or manipulation. &
All parts of the image share the same forensic fingerprint, with no discontinuities that would indicate manipulation. The content is consistent with having been captured by a Nikon D3200. The image has been JPEG compressed at quality 88. \\
\bottomrule
\end{tabular}
\end{table}

\begin{table}[p]
\centering
\caption{Qualitative examples from FKG-50K: \textbf{Authentic}. These are unmanipulated camera-original photographs. The predicted mask is blank (all zeros) because the system correctly identifies a single region with consistent source and processing throughout. (cont.)}
\smallerr
\setlength{\tabcolsep}{2pt}
\renewcommand{\arraystretch}{1.1}
\begin{tabular}{
	>{\centering\arraybackslash}m{17mm}
	>{\centering\arraybackslash}m{17mm}
	>{\noindent\justifying\arraybackslash}m{24mm}
	>{\noindent\justifying\arraybackslash}m{55mm}
}
\toprule
\textbf{Image} & \textbf{Pred Mask} & \textbf{Decision} & \textbf{Justification} \\
\midrule
\rowcolor{gray!30}
\timg{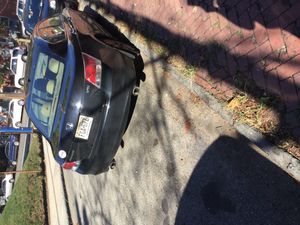} &
\timg{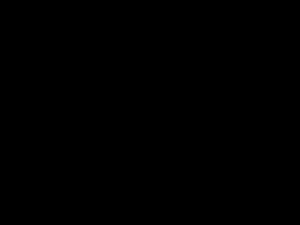} &
Analysis indicates this image is a genuine, untampered photograph. &
No forensic anomalies were detected anywhere in the image. The uniform characteristics point to a iPhone 6 as the source device. The image has been JPEG compressed at quality 93. \\
\bottomrule
\end{tabular}
\end{table}

\begin{table}[p]
\centering
\caption{Qualitative examples from FKG-50K: \textbf{Splicing}. The ground-truth mask marks the donor region composited from a different camera source.}
\label{tab:qual_splicing}
\smallerr
\setlength{\tabcolsep}{2pt}
\renewcommand{\arraystretch}{1.1}
\begin{tabular}{
	>{\centering\arraybackslash}m{17mm}
	>{\centering\arraybackslash}m{17mm}
	>{\centering\arraybackslash}m{17mm}
	>{\noindent\justifying\arraybackslash}m{20mm}
	>{\noindent\justifying\arraybackslash}m{48mm}
}
\toprule
\textbf{Image} & \textbf{GT Mask} & \textbf{Pred Mask} & \textbf{Decision} & \textbf{Justification} \\
\midrule
\rowcolor{gray!30}
\timg{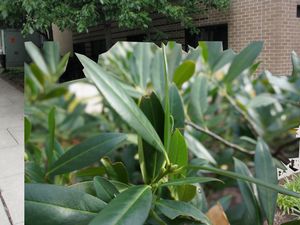} &
\timg{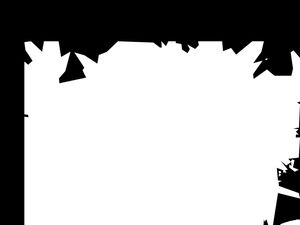} &
\timg{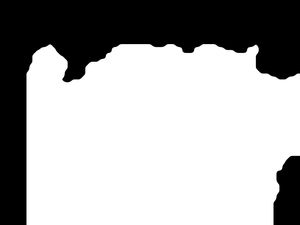} &
This image has been spliced , it contains content from two different cameras. &
Forensic analysis reveals clear evidence of tampering, with two distinct regions identified in the image. One region was captured by a Sony A6000, while the other originates from a Canon EOS 600D. The Canon EOS 600D region shows signs of downscaled to approximately 58\% of its original resolution. The fact that two different cameras produced content within the same image is strong evidence of splicing. \\
\timg{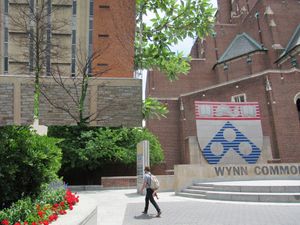} &
\timg{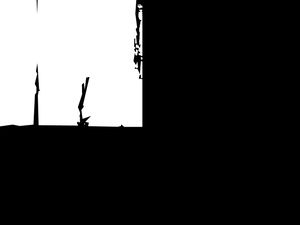} &
\timg{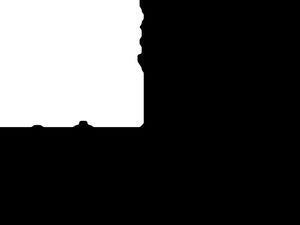} &
Image splicing detected: portions of this image originate from two separate cameras. &
The image shows strong signs of manipulation, with the content separating into two forensically different regions. The first identified region is consistent with a iPhone 6, and the second region traces back to a Sony A6000. The iPhone 6 region shows signs of downscaled to approximately 75\% of its original resolution. Finding content from two separate cameras within a single image confirms that splicing has occurred. \\
\rowcolor{gray!30}
\timg{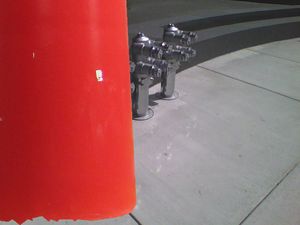} &
\timg{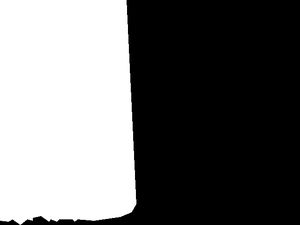} &
\timg{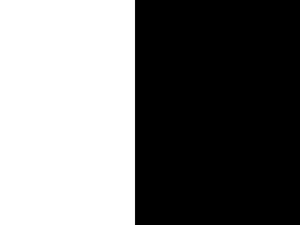} &
This image is a composite, combining photographs from two different camera sources. &
Analysis of the image reveals two regions with markedly different forensic characteristics. Part of the image matches the forensic signature of a Canon EOS 5D Mark II, whereas the remaining region was captured by a iPhone 6s. Since different camera sensors leave distinct forensic fingerprints, their co-occurrence within one image is a clear indicator of splicing. \\
\timg{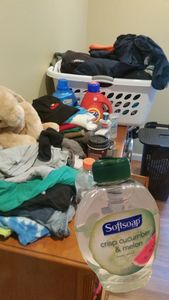} &
\timg{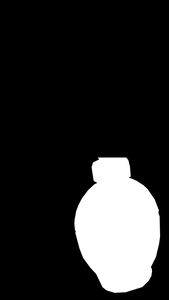} &
\timg{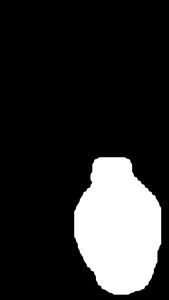} &
Splicing is present in this image , two forensically distinct camera sources were identified. &
Examination of this image uncovers two areas with distinctly different origins. One portion of the image was taken with a iPhone 6, while the rest matches a Nikon D3300. The presence of two camera-specific forensic signatures in different parts of the image is a hallmark of image splicing. \\
\rowcolor{gray!30}
\timg{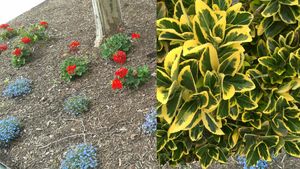} &
\timg{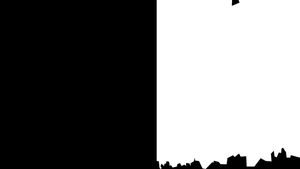} &
\timg{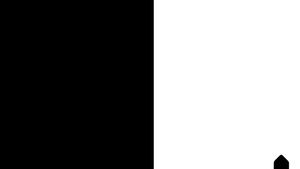} &
This image has been manipulated through splicing, merging content captured by different devices. &
The forensic evidence clearly indicates that this image is not uniform , two separate regions were detected. Analysis identifies one region as originating from a iPhone 6, and the other portion is attributed to a Nikon D90. The iPhone 6 region shows signs of downscaled to approximately 52\% of its original resolution. Distinct camera traces in separate spatial regions leave little doubt that this image was assembled from multiple photographs. \\
\timg{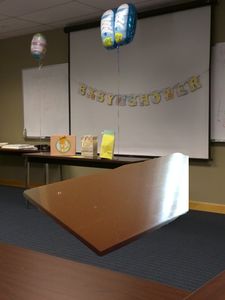} &
\timg{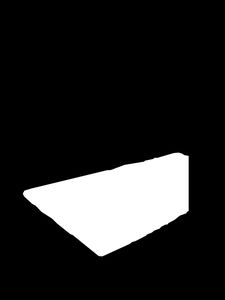} &
\timg{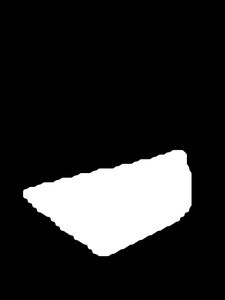} &
This image has been spliced , it contains content from two different cameras. &
Forensic analysis reveals clear evidence of tampering, with two distinct regions identified in the image. One region was captured by a iPhone 6s, while the other originates from a iPhone X. The fact that two different cameras produced content within the same image is strong evidence of splicing. \\
\bottomrule
\end{tabular}
\end{table}

\begin{table}[p]
\centering
\caption{Qualitative examples from FKG-50K: \textbf{Splicing}. The ground-truth mask marks the donor region composited from a different camera source. (cont.)}
\smallerr
\setlength{\tabcolsep}{2pt}
\renewcommand{\arraystretch}{1.1}
\begin{tabular}{
	>{\centering\arraybackslash}m{17mm}
	>{\centering\arraybackslash}m{17mm}
	>{\centering\arraybackslash}m{17mm}
	>{\noindent\justifying\arraybackslash}m{20mm}
	>{\noindent\justifying\arraybackslash}m{48mm}
}
\toprule
\textbf{Image} & \textbf{GT Mask} & \textbf{Pred Mask} & \textbf{Decision} & \textbf{Justification} \\
\midrule
\rowcolor{gray!30}
\timg{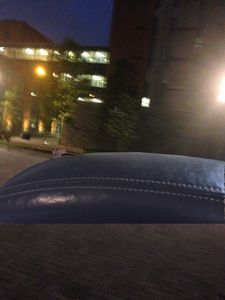} &
\timg{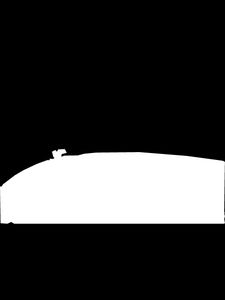} &
\timg{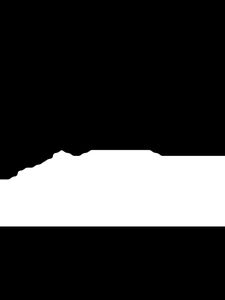} &
Image splicing detected: portions of this image originate from two separate cameras. &
The image shows strong signs of manipulation, with the content separating into two forensically different regions. The first identified region is consistent with a iPhone 5, and the second region traces back to a iPhone 6. Finding content from two separate cameras within a single image confirms that splicing has occurred. \\
\timg{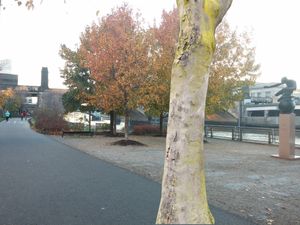} &
\timg{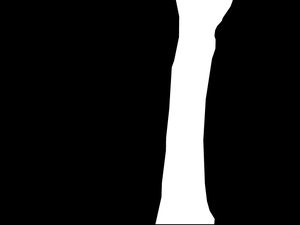} &
\timg{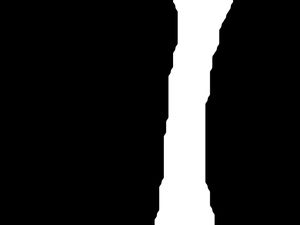} &
This image is a composite, combining photographs from two different camera sources. &
Analysis of the image reveals two regions with markedly different forensic characteristics. Part of the image matches the forensic signature of a iPhone X, whereas the remaining region was captured by a iPhone 6s. Since different camera sensors leave distinct forensic fingerprints, their co-occurrence within one image is a clear indicator of splicing. \\
\rowcolor{gray!30}
\timg{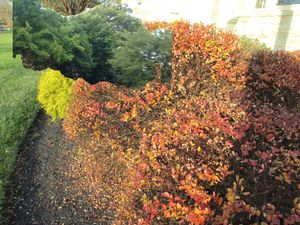} &
\timg{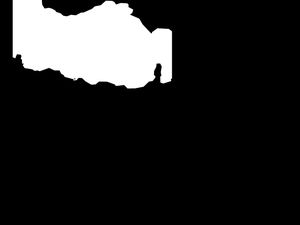} &
\timg{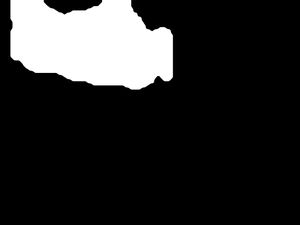} &
Splicing is present in this image , two forensically distinct camera sources were identified. &
Examination of this image uncovers two areas with distinctly different origins. One portion of the image was taken with a iPhone 5, while the rest matches a Sony A6000. The presence of two camera-specific forensic signatures in different parts of the image is a hallmark of image splicing. \\
\timg{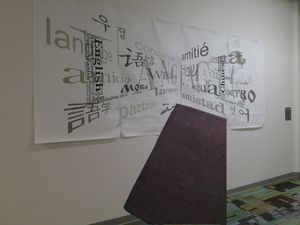} &
\timg{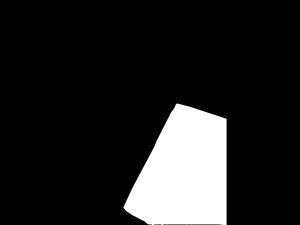} &
\timg{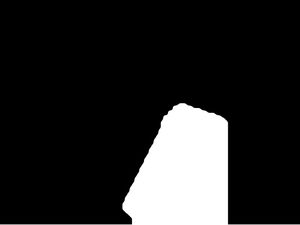} &
This image has been manipulated through splicing, merging content captured by different devices. &
The forensic evidence clearly indicates that this image is not uniform , two separate regions were detected. Analysis identifies one region as originating from a Nikon D3300, and the other portion is attributed to a iPhone 6s. Distinct camera traces in separate spatial regions leave little doubt that this image was assembled from multiple photographs. \\
\rowcolor{gray!30}
\timg{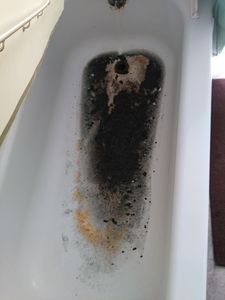} &
\timg{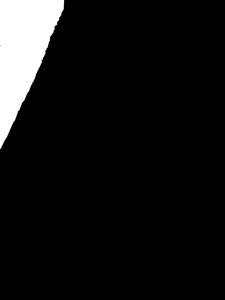} &
\timg{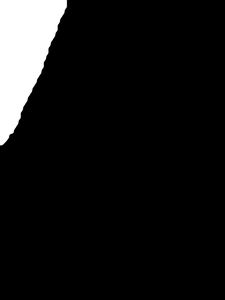} &
This image has been spliced , it contains content from two different cameras. &
Forensic analysis reveals clear evidence of tampering, with two distinct regions identified in the image. One region was captured by a Nikon D3300, while the other originates from a iPhone X. The fact that two different cameras produced content within the same image is strong evidence of splicing. \\
\timg{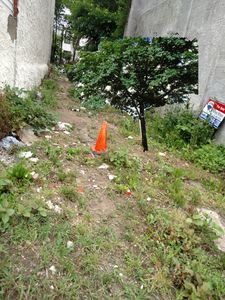} &
\timg{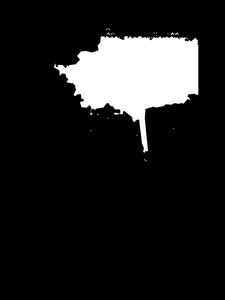} &
\timg{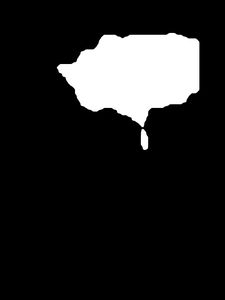} &
Image splicing detected: portions of this image originate from two separate cameras. &
The image shows strong signs of manipulation, with the content separating into two forensically different regions. The first identified region is consistent with a iPhone 5s, and the second region traces back to a Nikon D3300. Finding content from two separate cameras within a single image confirms that splicing has occurred. \\
\bottomrule
\end{tabular}
\end{table}

\begin{table}[p]
\centering
\caption{Qualitative examples from FKG-50K: \textbf{Splicing}. The ground-truth mask marks the donor region composited from a different camera source. (cont.)}
\smallerr
\setlength{\tabcolsep}{2pt}
\renewcommand{\arraystretch}{1.1}
\begin{tabular}{
	>{\centering\arraybackslash}m{17mm}
	>{\centering\arraybackslash}m{17mm}
	>{\centering\arraybackslash}m{17mm}
	>{\noindent\justifying\arraybackslash}m{20mm}
	>{\noindent\justifying\arraybackslash}m{48mm}
}
\toprule
\textbf{Image} & \textbf{GT Mask} & \textbf{Pred Mask} & \textbf{Decision} & \textbf{Justification} \\
\midrule
\rowcolor{gray!30}
\timg{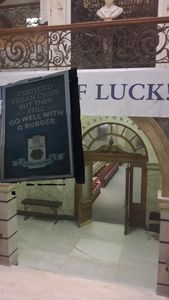} &
\timg{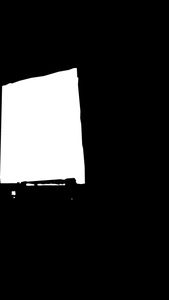} &
\timg{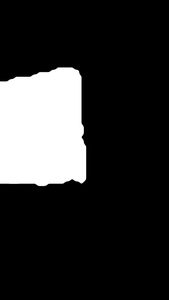} &
This image is a composite, combining photographs from two different camera sources. &
Analysis of the image reveals two regions with markedly different forensic characteristics. Part of the image matches the forensic signature of a Nikon D3300, whereas the remaining region was captured by a iPhone 6. Since different camera sensors leave distinct forensic fingerprints, their co-occurrence within one image is a clear indicator of splicing. \\
\timg{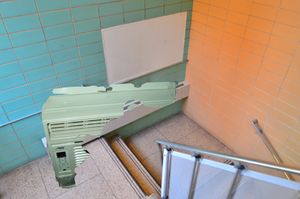} &
\timg{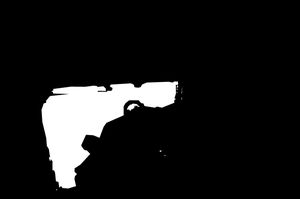} &
\timg{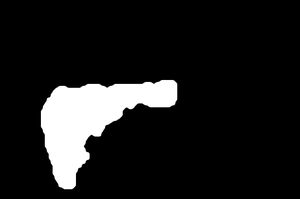} &
Splicing is present in this image , two forensically distinct camera sources were identified. &
Examination of this image uncovers two areas with distinctly different origins. One portion of the image was taken with a Apple Iphone 7 Plus, while the rest matches a Nikon D7000. The presence of two camera-specific forensic signatures in different parts of the image is a hallmark of image splicing. \\
\rowcolor{gray!30}
\timg{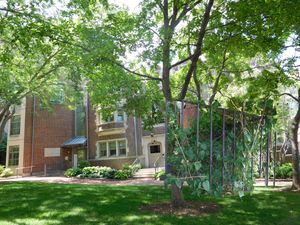} &
\timg{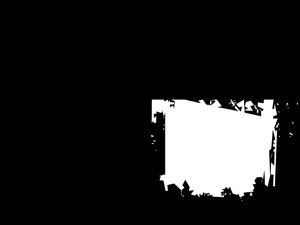} &
\timg{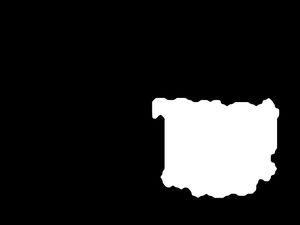} &
This image has been manipulated through splicing, merging content captured by different devices. &
The forensic evidence clearly indicates that this image is not uniform , two separate regions were detected. Analysis identifies one region as originating from a Nikon D3300, and the other portion is attributed to a Sony A7 III. Distinct camera traces in separate spatial regions leave little doubt that this image was assembled from multiple photographs. \\
\timg{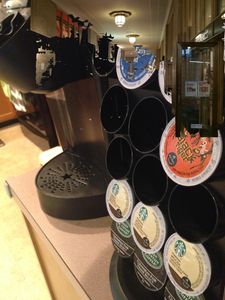} &
\timg{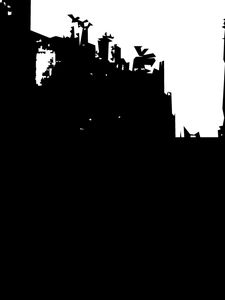} &
\timg{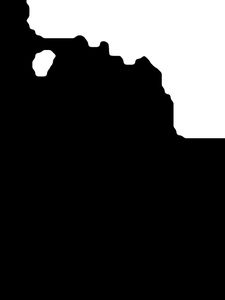} &
This image has been spliced , it contains content from two different cameras. &
Forensic analysis reveals clear evidence of tampering, with two distinct regions identified in the image. One region was captured by a iPhone 5s, while the other originates from a Sony A7 III. The Sony A7 III region shows signs of downscaled to approximately 50\% of its original resolution. The fact that two different cameras produced content within the same image is strong evidence of splicing. \\
\rowcolor{gray!30}
\timg{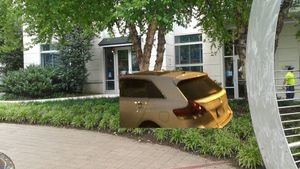} &
\timg{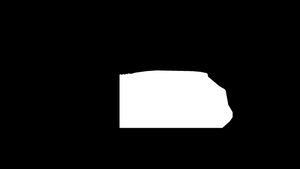} &
\timg{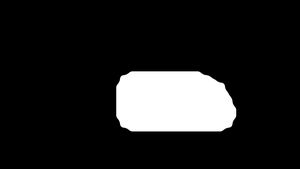} &
Image splicing detected: portions of this image originate from two separate cameras. &
The image shows strong signs of manipulation, with the content separating into two forensically different regions. The first identified region is consistent with a Nikon D90, and the second region traces back to a Sony A7 III. Finding content from two separate cameras within a single image confirms that splicing has occurred. \\
\timg{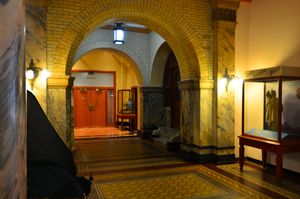} &
\timg{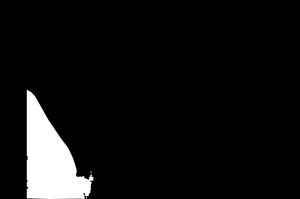} &
\timg{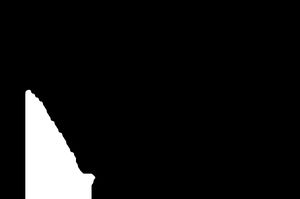} &
This image is a composite, combining photographs from two different camera sources. &
Analysis of the image reveals two regions with markedly different forensic characteristics. Part of the image matches the forensic signature of a Nikon D7000, whereas the remaining region was captured by a iPhone 5s. Since different camera sensors leave distinct forensic fingerprints, their co-occurrence within one image is a clear indicator of splicing. \\
\bottomrule
\end{tabular}
\end{table}

\begin{table}[p]
\centering
\caption{Qualitative examples from FKG-50K: \textbf{Splicing}. The ground-truth mask marks the donor region composited from a different camera source. (cont.)}
\smallerr
\setlength{\tabcolsep}{2pt}
\renewcommand{\arraystretch}{1.1}
\begin{tabular}{
	>{\centering\arraybackslash}m{17mm}
	>{\centering\arraybackslash}m{17mm}
	>{\centering\arraybackslash}m{17mm}
	>{\noindent\justifying\arraybackslash}m{20mm}
	>{\noindent\justifying\arraybackslash}m{48mm}
}
\toprule
\textbf{Image} & \textbf{GT Mask} & \textbf{Pred Mask} & \textbf{Decision} & \textbf{Justification} \\
\midrule
\rowcolor{gray!30}
\timg{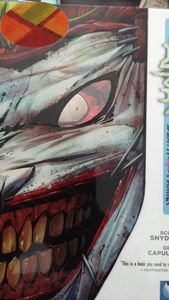} &
\timg{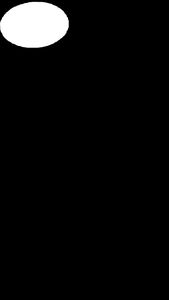} &
\timg{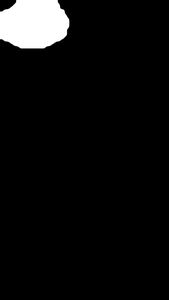} &
Splicing is present in this image , two forensically distinct camera sources were identified. &
Examination of this image uncovers two areas with distinctly different origins. One portion of the image was taken with a iPhone X, while the rest matches a Sony A7 III. The Sony A7 III region shows signs of upscaled to approximately 131\% of its original resolution. The presence of two camera-specific forensic signatures in different parts of the image is a hallmark of image splicing. \\
\timg{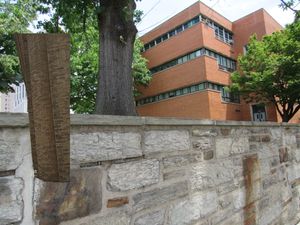} &
\timg{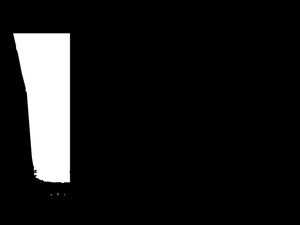} &
\timg{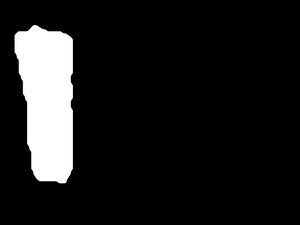} &
This image has been manipulated through splicing, merging content captured by different devices. &
The forensic evidence clearly indicates that this image is not uniform , two separate regions were detected. Analysis identifies one region as originating from a Sony A6000, and the other portion is attributed to a iPhone 6. Distinct camera traces in separate spatial regions leave little doubt that this image was assembled from multiple photographs. \\
\rowcolor{gray!30}
\timg{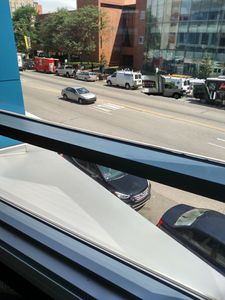} &
\timg{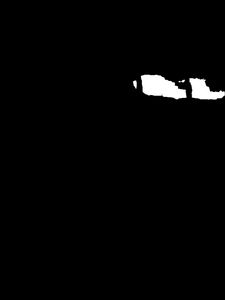} &
\timg{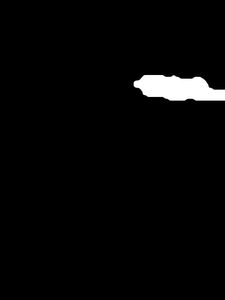} &
This image has been spliced , it contains content from two different cameras. &
Forensic analysis reveals clear evidence of tampering, with two distinct regions identified in the image. One region was captured by a iPhone X, while the other originates from a iPhone 5s. The iPhone X region shows signs of upscaled to approximately 127\% of its original resolution. The fact that two different cameras produced content within the same image is strong evidence of splicing. \\
\timg{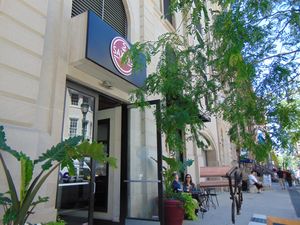} &
\timg{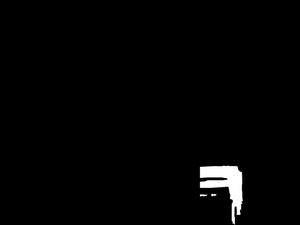} &
\timg{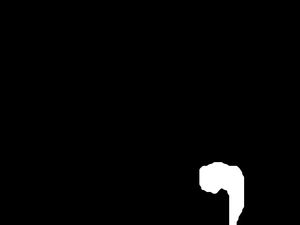} &
Image splicing detected: portions of this image originate from two separate cameras. &
The image shows strong signs of manipulation, with the content separating into two forensically different regions. The first identified region is consistent with a Apple Iphone 7, and the second region traces back to a Nikon D3300. The Apple Iphone 7 region shows signs of upscaled to approximately 120\% of its original resolution. Finding content from two separate cameras within a single image confirms that splicing has occurred. \\
\rowcolor{gray!30}
\timg{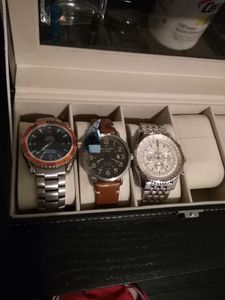} &
\timg{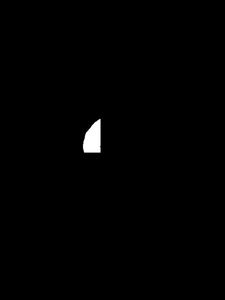} &
\timg{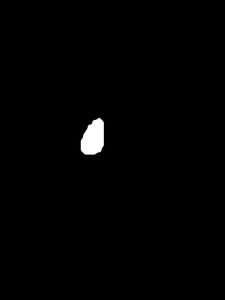} &
This image is a composite, combining photographs from two different camera sources. &
Analysis of the image reveals two regions with markedly different forensic characteristics. Part of the image matches the forensic signature of a iPhone X, whereas the remaining region was captured by a Sony A7 III. The iPhone X region shows signs of upscaled to approximately 122\% of its original resolution. Since different camera sensors leave distinct forensic fingerprints, their co-occurrence within one image is a clear indicator of splicing. \\
\timg{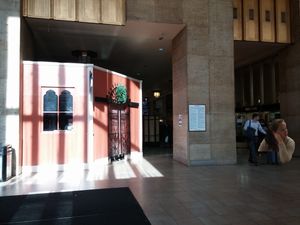} &
\timg{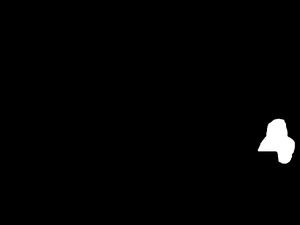} &
\timg{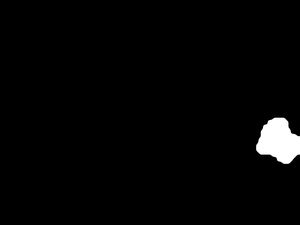} &
Splicing is present in this image , two forensically distinct camera sources were identified. &
Examination of this image uncovers two areas with distinctly different origins. One portion of the image was taken with a iPhone X, while the rest matches a iPhone 5s. The presence of two camera-specific forensic signatures in different parts of the image is a hallmark of image splicing. \\
\bottomrule
\end{tabular}
\end{table}

\begin{table}[p]
\centering
\caption{Qualitative examples from FKG-50K: \textbf{Splicing}. The ground-truth mask marks the donor region composited from a different camera source. (cont.)}
\smallerr
\setlength{\tabcolsep}{2pt}
\renewcommand{\arraystretch}{1.1}
\begin{tabular}{
	>{\centering\arraybackslash}m{17mm}
	>{\centering\arraybackslash}m{17mm}
	>{\centering\arraybackslash}m{17mm}
	>{\noindent\justifying\arraybackslash}m{20mm}
	>{\noindent\justifying\arraybackslash}m{48mm}
}
\toprule
\textbf{Image} & \textbf{GT Mask} & \textbf{Pred Mask} & \textbf{Decision} & \textbf{Justification} \\
\midrule
\rowcolor{gray!30}
\timg{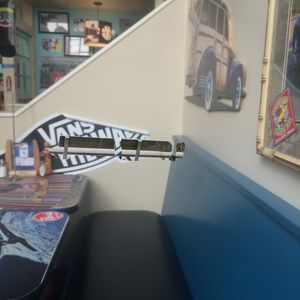} &
\timg{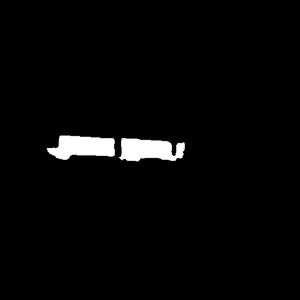} &
\timg{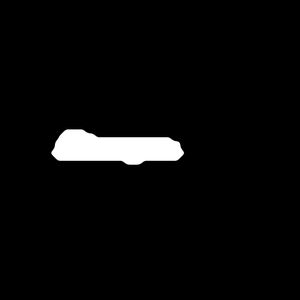} &
This image has been manipulated through splicing, merging content captured by different devices. &
The forensic evidence clearly indicates that this image is not uniform , two separate regions were detected. Analysis identifies one region as originating from a iPhone 5s, and the other portion is attributed to a iPhone X. The iPhone X region shows signs of upscaled to approximately 122\% of its original resolution. Distinct camera traces in separate spatial regions leave little doubt that this image was assembled from multiple photographs. \\
\bottomrule
\end{tabular}
\end{table}

\begin{table}[p]
\centering
\caption{Qualitative examples from FKG-50K: \textbf{Traditional Editing} (blur, noise, sharpening, resampling). The ground-truth mask marks the locally edited region.}
\label{tab:qual_traditional_editing}
\smallerr
\setlength{\tabcolsep}{2pt}
\renewcommand{\arraystretch}{1.1}
\begin{tabular}{
	>{\centering\arraybackslash}m{17mm}
	>{\centering\arraybackslash}m{17mm}
	>{\centering\arraybackslash}m{17mm}
	>{\noindent\justifying\arraybackslash}m{20mm}
	>{\noindent\justifying\arraybackslash}m{48mm}
}
\toprule
\textbf{Image} & \textbf{GT Mask} & \textbf{Pred Mask} & \textbf{Decision} & \textbf{Justification} \\
\midrule
\rowcolor{gray!30}
\timg{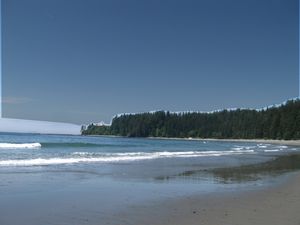} &
\timg{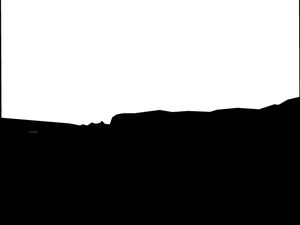} &
\timg{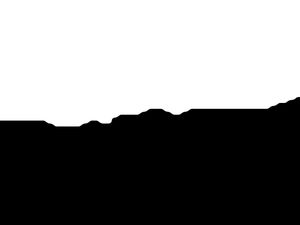} &
This image has been locally edited , the same camera captured it, but part of the image was post-processed differently. &
Forensic analysis reveals clear evidence of tampering, with two distinct regions identified in the image. Both regions originate from the same camera, a Sony A6000. However, the regions differ in their compression characteristics (quality approximately 90 vs. 92). Both regions share the same camera source but exhibit different processing traces, which is characteristic of traditional image editing. \\
\timg{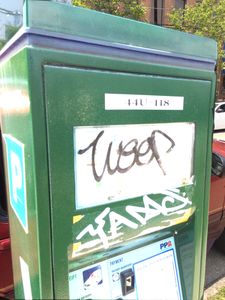} &
\timg{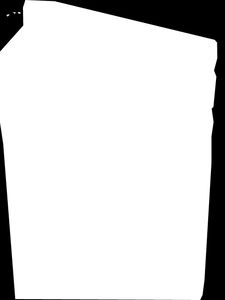} &
\timg{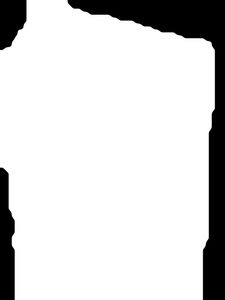} &
Traditional editing detected: a region of this image shows different processing characteristics than the rest. &
The image shows strong signs of manipulation, with the content separating into two forensically different regions. The entire image was captured by a single device, a iPhone 5s. However, one region shows signs of sharpening, while the other appears unprocessed. The divergent forensic signatures across regions from the same camera are consistent with localized content manipulation. \\
\rowcolor{gray!30}
\timg{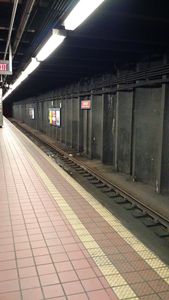} &
\timg{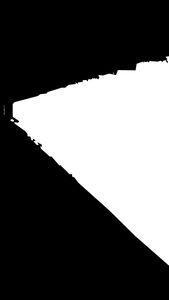} &
\timg{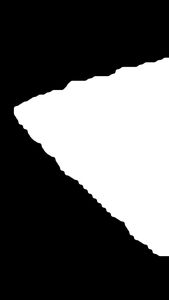} &
This image has undergone localized manipulation , while it originates from a single camera, part of it has been altered. &
Analysis of the image reveals two regions with markedly different forensic characteristics. Both portions of the image trace back to the same iPhone 5s. However, the forensic traces in each region show subtle but measurable differences. When different parts of a single-camera image show inconsistent processing artifacts, it typically indicates that selective edits were applied. \\
\timg{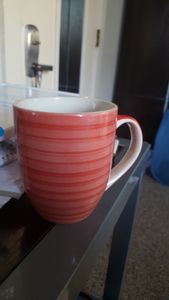} &
\timg{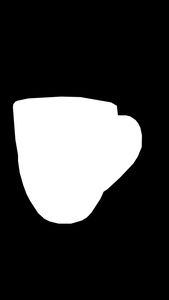} &
\timg{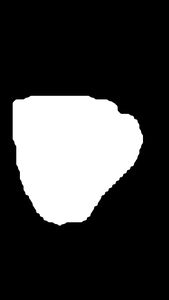} &
Local content manipulation found: forensic traces differ across regions despite sharing the same camera source. &
Examination of this image uncovers two areas with distinctly different origins. Forensic analysis attributes the full image to one camera source, a Sony A7 III. However, the regions differ in their compression characteristics (quality approximately 92 vs. 68). Matching camera sources with mismatched processing characteristics point to traditional editing operations applied to a specific area. \\
\rowcolor{gray!30}
\timg{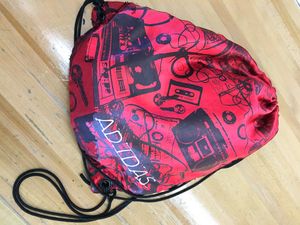} &
\timg{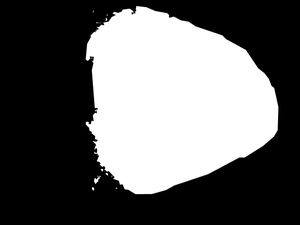} &
\timg{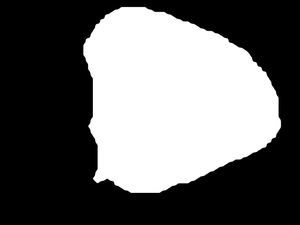} &
Part of this image has been traditionally edited, introducing forensic inconsistencies within an otherwise uniform photograph. &
The forensic evidence clearly indicates that this image is not uniform , two separate regions were detected. The underlying image was uniformly captured by a iPhone 6s. However, the regions differ in their compression characteristics (quality approximately 94 vs. 91). The forensic inconsistency between regions , despite originating from the same device , is a telltale sign of localized image manipulation. \\
\timg{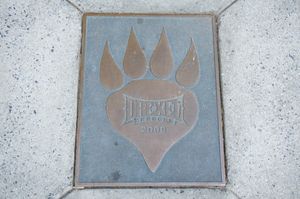} &
\timg{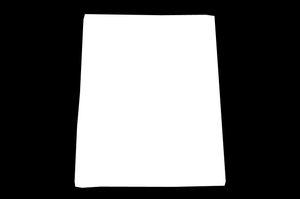} &
\timg{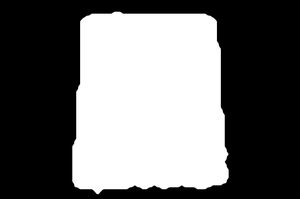} &
This image has been locally edited , the same camera captured it, but part of the image was post-processed differently. &
Forensic analysis reveals clear evidence of tampering, with two distinct regions identified in the image. Both regions originate from the same camera, a Nikon D3200. However, the regions differ in their compression characteristics (quality approximately 82 vs. 90). Both regions share the same camera source but exhibit different processing traces, which is characteristic of traditional image editing. \\
\bottomrule
\end{tabular}
\end{table}

\begin{table}[p]
\centering
\caption{Qualitative examples from FKG-50K: \textbf{Traditional Editing} (blur, noise, sharpening, resampling). The ground-truth mask marks the locally edited region. (cont.)}
\smallerr
\setlength{\tabcolsep}{2pt}
\renewcommand{\arraystretch}{1.1}
\begin{tabular}{
	>{\centering\arraybackslash}m{17mm}
	>{\centering\arraybackslash}m{17mm}
	>{\centering\arraybackslash}m{17mm}
	>{\noindent\justifying\arraybackslash}m{20mm}
	>{\noindent\justifying\arraybackslash}m{48mm}
}
\toprule
\textbf{Image} & \textbf{GT Mask} & \textbf{Pred Mask} & \textbf{Decision} & \textbf{Justification} \\
\midrule
\rowcolor{gray!30}
\timg{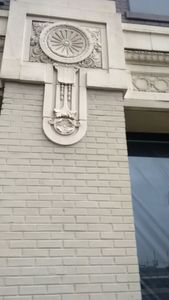} &
\timg{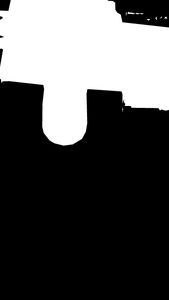} &
\timg{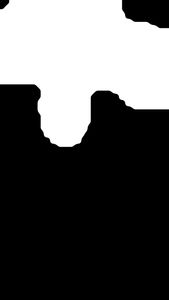} &
Traditional editing detected: a region of this image shows different processing characteristics than the rest. &
The image shows strong signs of manipulation, with the content separating into two forensically different regions. The entire image was captured by a single device, a Sony A7 III. However, the forensic traces in each region show subtle but measurable differences. The divergent forensic signatures across regions from the same camera are consistent with localized content manipulation. \\
\timg{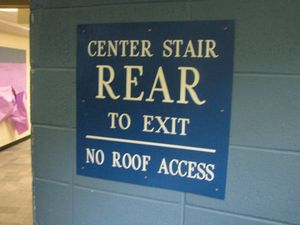} &
\timg{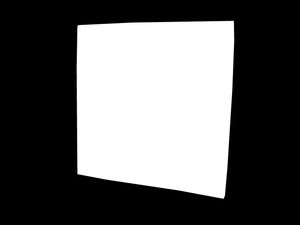} &
\timg{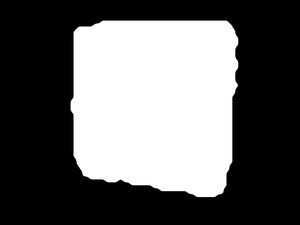} &
This image has undergone localized manipulation , while it originates from a single camera, part of it has been altered. &
Analysis of the image reveals two regions with markedly different forensic characteristics. Both portions of the image trace back to the same iPhone 6. However, the forensic traces in each region show subtle but measurable differences. When different parts of a single-camera image show inconsistent processing artifacts, it typically indicates that selective edits were applied. \\
\rowcolor{gray!30}
\timg{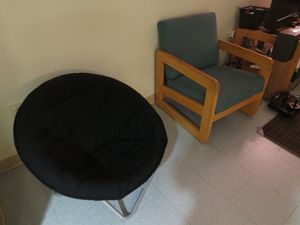} &
\timg{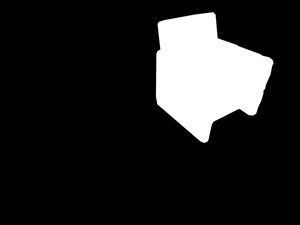} &
\timg{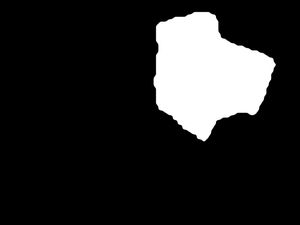} &
Local content manipulation found: forensic traces differ across regions despite sharing the same camera source. &
Examination of this image uncovers two areas with distinctly different origins. Forensic analysis attributes the full image to one camera source, a iPhone 6s. However, the regions differ in their compression characteristics (quality approximately 92 vs. 88). Matching camera sources with mismatched processing characteristics point to traditional editing operations applied to a specific area. \\
\timg{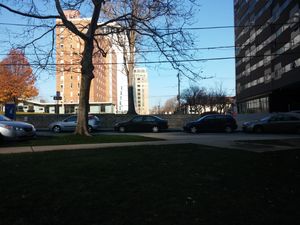} &
\timg{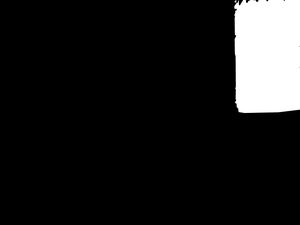} &
\timg{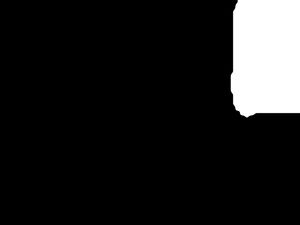} &
Part of this image has been traditionally edited, introducing forensic inconsistencies within an otherwise uniform photograph. &
The forensic evidence clearly indicates that this image is not uniform , two separate regions were detected. The underlying image was uniformly captured by a iPhone 5. However, the regions differ in their compression characteristics (quality approximately 88 vs. 91). The forensic inconsistency between regions , despite originating from the same device , is a telltale sign of localized image manipulation. \\
\rowcolor{gray!30}
\timg{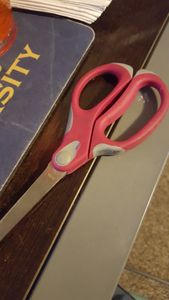} &
\timg{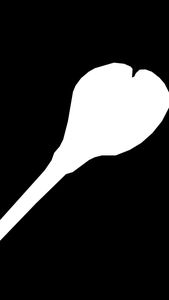} &
\timg{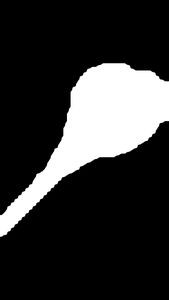} &
This image has been locally edited , the same camera captured it, but part of the image was post-processed differently. &
Forensic analysis reveals clear evidence of tampering, with two distinct regions identified in the image. Both regions originate from the same camera, a Sony A7 III. However, the regions differ in their compression characteristics (quality approximately 58 vs. 95). Both regions share the same camera source but exhibit different processing traces, which is characteristic of traditional image editing. \\
\timg{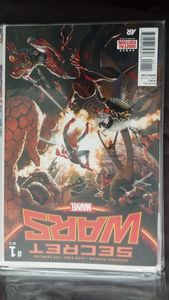} &
\timg{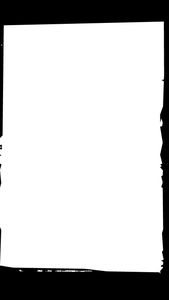} &
\timg{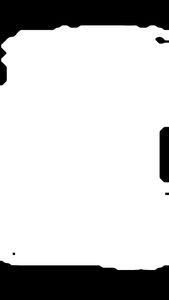} &
Traditional editing detected: a region of this image shows different processing characteristics than the rest. &
The image shows strong signs of manipulation, with the content separating into two forensically different regions. The entire image was captured by a single device, a iPhone X. However, the forensic traces in each region show subtle but measurable differences. The divergent forensic signatures across regions from the same camera are consistent with localized content manipulation. \\
\bottomrule
\end{tabular}
\end{table}

\begin{table}[p]
\centering
\caption{Qualitative examples from FKG-50K: \textbf{Traditional Editing} (blur, noise, sharpening, resampling). The ground-truth mask marks the locally edited region. (cont.)}
\smallerr
\setlength{\tabcolsep}{2pt}
\renewcommand{\arraystretch}{1.1}
\begin{tabular}{
	>{\centering\arraybackslash}m{17mm}
	>{\centering\arraybackslash}m{17mm}
	>{\centering\arraybackslash}m{17mm}
	>{\noindent\justifying\arraybackslash}m{20mm}
	>{\noindent\justifying\arraybackslash}m{48mm}
}
\toprule
\textbf{Image} & \textbf{GT Mask} & \textbf{Pred Mask} & \textbf{Decision} & \textbf{Justification} \\
\midrule
\rowcolor{gray!30}
\timg{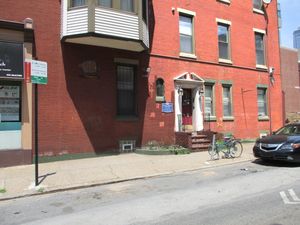} &
\timg{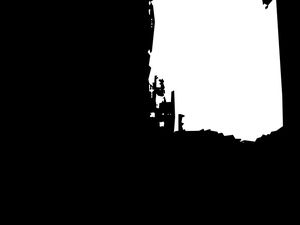} &
\timg{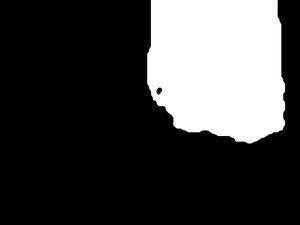} &
This image has undergone localized manipulation , while it originates from a single camera, part of it has been altered. &
Analysis of the image reveals two regions with markedly different forensic characteristics. Both portions of the image trace back to the same iPhone 6s. However, one region shows signs of sharpening, while the other appears unprocessed. When different parts of a single-camera image show inconsistent processing artifacts, it typically indicates that selective edits were applied. \\
\timg{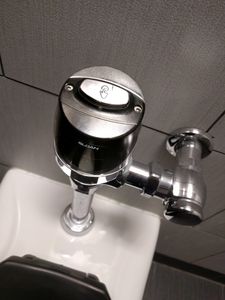} &
\timg{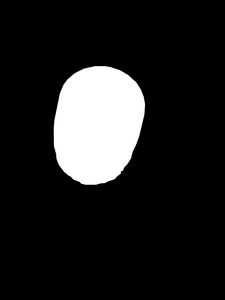} &
\timg{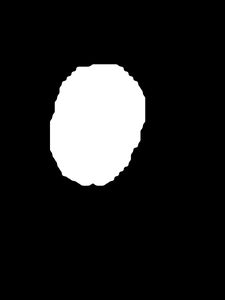} &
Local content manipulation found: forensic traces differ across regions despite sharing the same camera source. &
Examination of this image uncovers two areas with distinctly different origins. Forensic analysis attributes the full image to one camera source, a iPhone 5s. However, the forensic traces in each region show subtle but measurable differences. Matching camera sources with mismatched processing characteristics point to traditional editing operations applied to a specific area. \\
\rowcolor{gray!30}
\timg{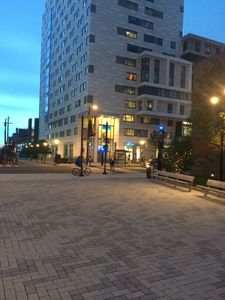} &
\timg{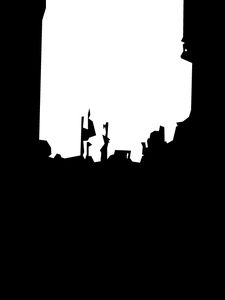} &
\timg{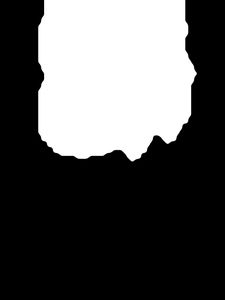} &
Part of this image has been traditionally edited, introducing forensic inconsistencies within an otherwise uniform photograph. &
The forensic evidence clearly indicates that this image is not uniform , two separate regions were detected. The underlying image was uniformly captured by a iPhone 5s. However, the forensic traces in each region show subtle but measurable differences. The forensic inconsistency between regions , despite originating from the same device , is a telltale sign of localized image manipulation. \\
\timg{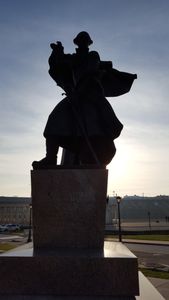} &
\timg{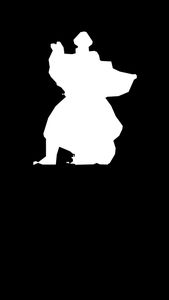} &
\timg{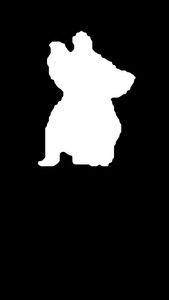} &
This image has been locally edited , the same camera captured it, but part of the image was post-processed differently. &
Forensic analysis reveals clear evidence of tampering, with two distinct regions identified in the image. Both regions originate from the same camera, a Sony A7 III. However, the regions differ in their compression characteristics (quality approximately 90 vs. 87). Both regions share the same camera source but exhibit different processing traces, which is characteristic of traditional image editing. \\
\rowcolor{gray!30}
\timg{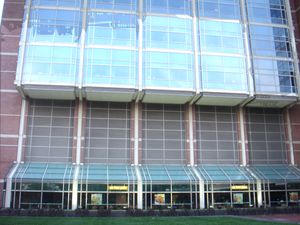} &
\timg{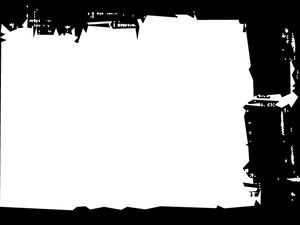} &
\timg{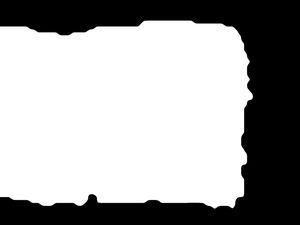} &
Traditional editing detected: a region of this image shows different processing characteristics than the rest. &
The image shows strong signs of manipulation, with the content separating into two forensically different regions. The entire image was captured by a single device, a Sony A6000. However, the forensic traces in each region show subtle but measurable differences. The divergent forensic signatures across regions from the same camera are consistent with localized content manipulation. \\
\timg{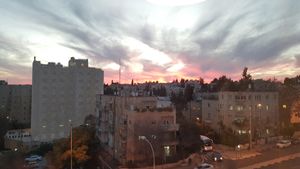} &
\timg{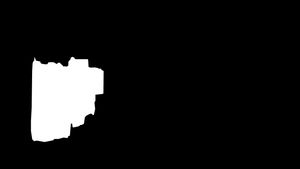} &
\timg{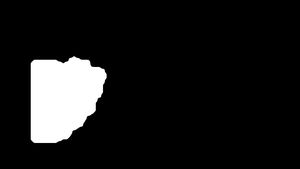} &
This image has undergone localized manipulation , while it originates from a single camera, part of it has been altered. &
Analysis of the image reveals two regions with markedly different forensic characteristics. Both portions of the image trace back to the same Sony A7 III. However, one region shows signs of additive noise, while the other does not. When different parts of a single-camera image show inconsistent processing artifacts, it typically indicates that selective edits were applied. \\
\bottomrule
\end{tabular}
\end{table}

\begin{table}[p]
\centering
\caption{Qualitative examples from FKG-50K: \textbf{Traditional Editing} (blur, noise, sharpening, resampling). The ground-truth mask marks the locally edited region. (cont.)}
\smallerr
\setlength{\tabcolsep}{2pt}
\renewcommand{\arraystretch}{1.1}
\begin{tabular}{
	>{\centering\arraybackslash}m{17mm}
	>{\centering\arraybackslash}m{17mm}
	>{\centering\arraybackslash}m{17mm}
	>{\noindent\justifying\arraybackslash}m{20mm}
	>{\noindent\justifying\arraybackslash}m{48mm}
}
\toprule
\textbf{Image} & \textbf{GT Mask} & \textbf{Pred Mask} & \textbf{Decision} & \textbf{Justification} \\
\midrule
\rowcolor{gray!30}
\timg{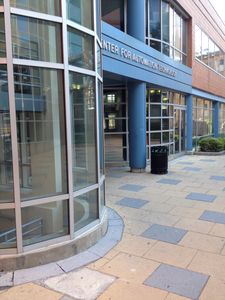} &
\timg{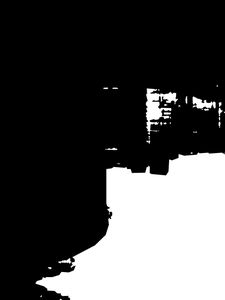} &
\timg{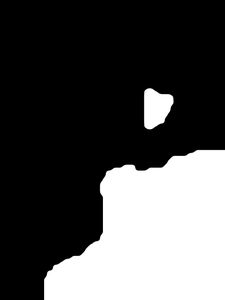} &
Local content manipulation found: forensic traces differ across regions despite sharing the same camera source. &
Examination of this image uncovers two areas with distinctly different origins. Forensic analysis attributes the full image to one camera source, a iPhone 5. However, the regions differ in their compression characteristics (quality approximately 92 vs. 94). Matching camera sources with mismatched processing characteristics point to traditional editing operations applied to a specific area. \\
\timg{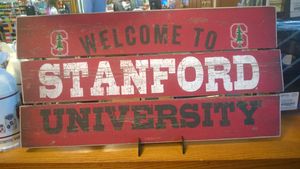} &
\timg{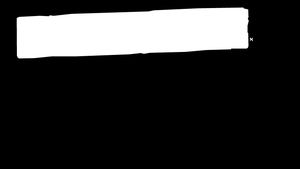} &
\timg{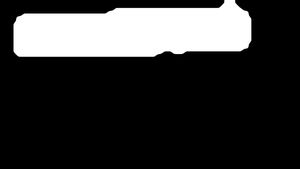} &
Part of this image has been traditionally edited, introducing forensic inconsistencies within an otherwise uniform photograph. &
The forensic evidence clearly indicates that this image is not uniform , two separate regions were detected. The underlying image was uniformly captured by a Sony A7 III. However, the regions differ in their compression characteristics (quality approximately 91 vs. 71). The forensic inconsistency between regions , despite originating from the same device , is a telltale sign of localized image manipulation. \\
\rowcolor{gray!30}
\timg{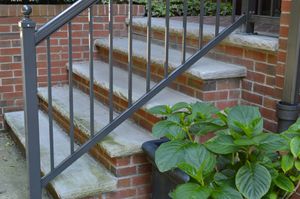} &
\timg{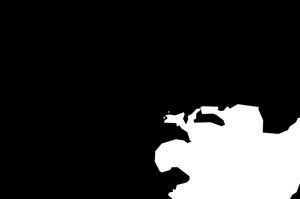} &
\timg{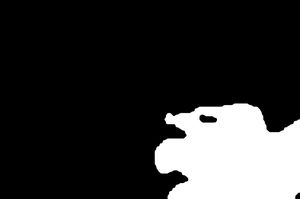} &
\red{This image has been partially edited using AI , a portion was generated by StyleGAN while the rest is an authentic photograph.} &
\red{Forensic analysis reveals clear evidence of tampering, with two distinct regions identified in the image. One region was produced by the StyleGAN AI model with moderate confidence, while the rest is an authentic photograph from a Nikon D3200. JPEG quality ranges from approximately 87 to 89 across regions. The mixture of AI-generated and camera-captured content within the same image indicates localized AI inpainting or editing.} \\
\timg{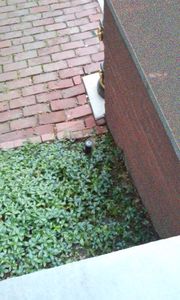} &
\timg{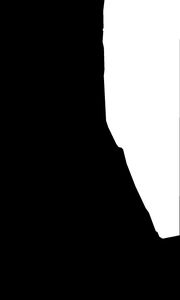} &
\timg{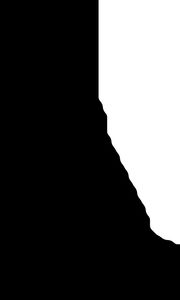} &
\red{This image has been spliced , it contains content from two different cameras.} &
\red{The image shows strong signs of manipulation, with the content separating into two forensically different regions. One region was captured by a iPhone 5s, while the other originates from a Sony A7 III. The Sony A7 III region shows signs of additive noise. The fact that two different cameras produced content within the same image is strong evidence of splicing.} \\
\rowcolor{gray!30}
\timg{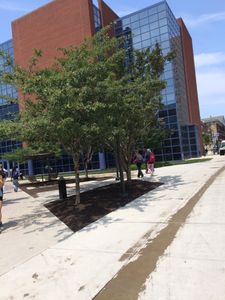} &
\timg{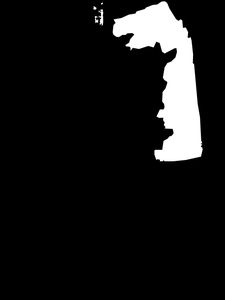} &
\timg{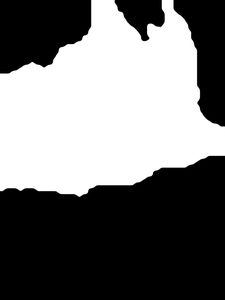} &
This image has been locally edited , the same camera captured it, but part of the image was post-processed differently. &
Analysis of the image reveals two regions with markedly different forensic characteristics. Both regions originate from the same camera, a iPhone 5s. However, the forensic traces in each region show subtle but measurable differences. Both regions share the same camera source but exhibit different processing traces, which is characteristic of traditional image editing. \\
\timg{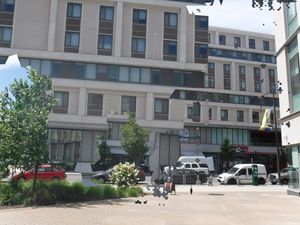} &
\timg{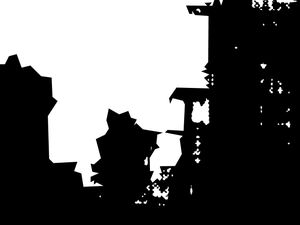} &
\timg{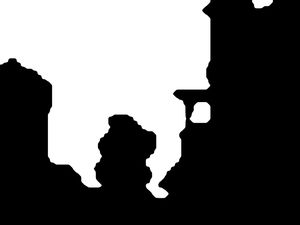} &
\red{Image splicing detected: portions of this image originate from two separate cameras.} &
\red{Examination of this image uncovers two areas with distinctly different origins. The first identified region is consistent with a Apple Iphone 7 Plus, and the second region traces back to a Nikon D3400. Finding content from two separate cameras within a single image confirms that splicing has occurred.} \\
\bottomrule
\end{tabular}
\end{table}

\begin{table}[p]
\centering
\caption{Qualitative examples from FKG-50K: \textbf{Traditional Editing} (blur, noise, sharpening, resampling). The ground-truth mask marks the locally edited region. (cont.)}
\smallerr
\setlength{\tabcolsep}{2pt}
\renewcommand{\arraystretch}{1.1}
\begin{tabular}{
	>{\centering\arraybackslash}m{17mm}
	>{\centering\arraybackslash}m{17mm}
	>{\centering\arraybackslash}m{17mm}
	>{\noindent\justifying\arraybackslash}m{20mm}
	>{\noindent\justifying\arraybackslash}m{48mm}
}
\toprule
\textbf{Image} & \textbf{GT Mask} & \textbf{Pred Mask} & \textbf{Decision} & \textbf{Justification} \\
\midrule
\rowcolor{gray!30}
\timg{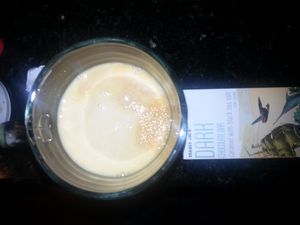} &
\timg{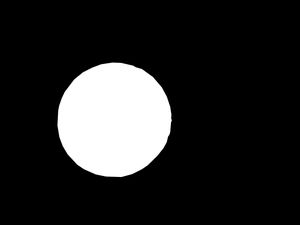} &
\timg{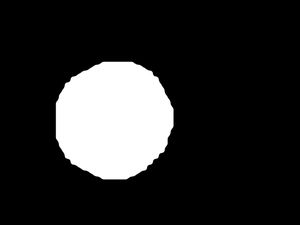} &
\red{This image is a composite, combining photographs from two different camera sources.} &
\red{The forensic evidence clearly indicates that this image is not uniform , two separate regions were detected. Part of the image matches the forensic signature of a Sony A6000, whereas the remaining region was captured by a Sony A7 III. Since different camera sensors leave distinct forensic fingerprints, their co-occurrence within one image is a clear indicator of splicing.} \\
\bottomrule
\end{tabular}
\end{table}

\begin{table}[p]
\centering
\caption{Qualitative examples from FKG-50K: \textbf{AI Editing} (inpainting-based object replacement, insertion, and removal). The ground-truth mask marks the AI-modified region.}
\label{tab:qual_ai_editing}
\smallerr
\setlength{\tabcolsep}{2pt}
\renewcommand{\arraystretch}{1.1}
\begin{tabular}{
	>{\centering\arraybackslash}m{17mm}
	>{\centering\arraybackslash}m{17mm}
	>{\centering\arraybackslash}m{17mm}
	>{\noindent\justifying\arraybackslash}m{20mm}
	>{\noindent\justifying\arraybackslash}m{48mm}
}
\toprule
\textbf{Image} & \textbf{GT Mask} & \textbf{Pred Mask} & \textbf{Decision} & \textbf{Justification} \\
\midrule
\rowcolor{gray!30}
\timg{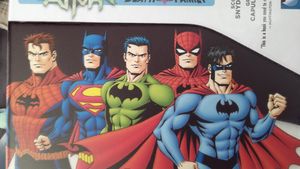} &
\timg{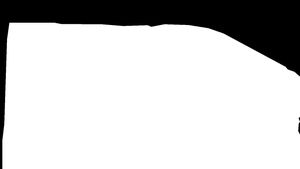} &
\timg{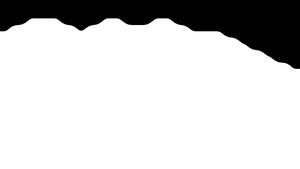} &
AI-assisted editing detected: part of this image was created by FLUX.1, inserted into a real photograph. &
Forensic analysis reveals two distinct regions. Part was generated by FLUX.1 (high confidence), while the rest is a genuine photo from a Sony A7 III. The Sony A7 III region was downscaled to ${\sim}$ 50\% of original size. JPEG quality ranges from ${\sim}$92 to 94. Both synthetic and authentic traces in different regions point to targeted AI manipulation. \\
\timg{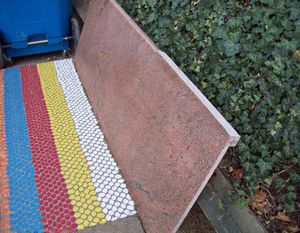} &
\timg{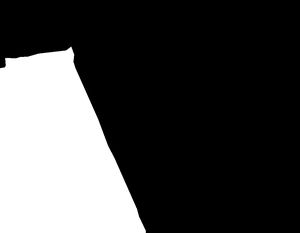} &
\timg{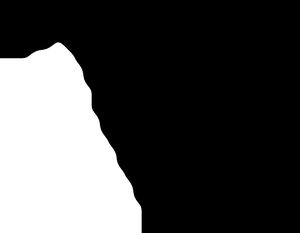} &
This image combines AI-generated content from Stable Diffusion XL with an authentic photograph. &
The image separates into two forensically different regions. The manipulated region was synthesized by Stable Diffusion XL (moderate confidence), while the host was captured by a iPhone 6s. The AI-generated region was downscaled to ${\sim}$ 56\% of original size. JPEG quality ranges from ${\sim}$92 to 93. The coexistence of machine-generated and camera-originated content is characteristic of AI-assisted editing. \\
\rowcolor{gray!30}
\timg{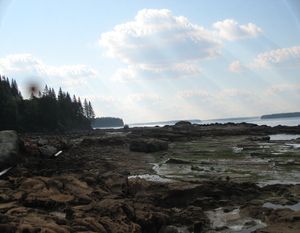} &
\timg{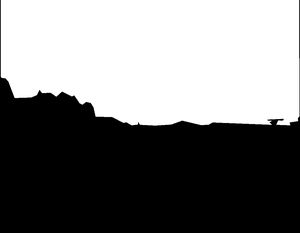} &
\timg{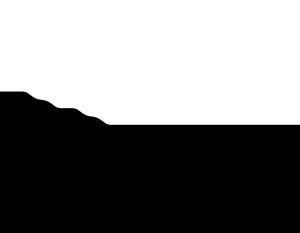} &
Localized AI editing found , Midjourney v6 was used to generate or modify a portion of this otherwise authentic image. &
Two regions with markedly different forensic characteristics were found. A portion was created using Midjourney v6 (moderate confidence), while the surrounding content originates from a Sony A6000. JPEG quality ranges from ${\sim}$91 to 93. This combination of artificial and natural signatures is typical of AI inpainting. \\
\timg{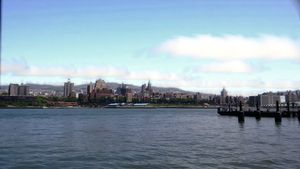} &
\timg{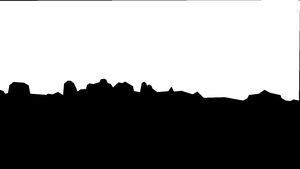} &
\timg{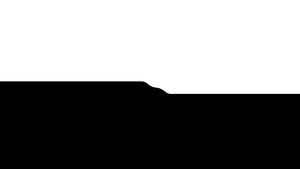} &
Part of this image has been synthesized by FLUX.1 and blended into a genuine photograph. &
Two areas with distinctly different origins were found. One exhibits FLUX.1-generated content (high confidence), whereas the unaltered portion was photographed with a iPhone 6. The iPhone 6 region was downscaled to ${\sim}$ 52\% of original size. JPEG quality ranges from ${\sim}$92 to 95. The contrast between AI-synthesized and naturally captured regions reveals generative editing was applied. \\
\rowcolor{gray!30}
\timg{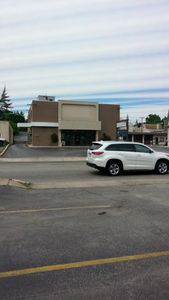} &
\timg{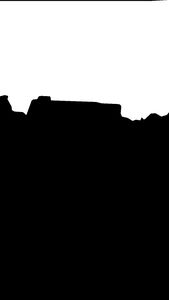} &
\timg{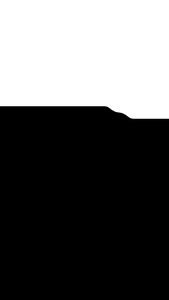} &
This image has been partially edited using AI , a portion was generated by FLUX.1 while the rest is an authentic photograph. &
Two separate regions were detected. One was produced by FLUX.1 (high), while the rest is an authentic photograph from a iPhone 6. The iPhone 6 region was downscaled to ${\sim}$ 57\% of original size. JPEG quality ranges from ${\sim}$90 to 91. The mixture of AI-generated and camera-captured content indicates localized AI editing. \\
\timg{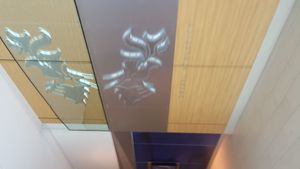} &
\timg{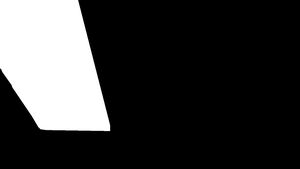} &
\timg{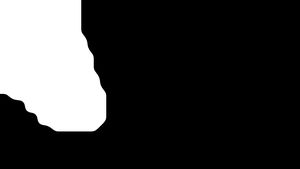} &
AI-assisted editing detected: part of this image was created by FLUX.1, inserted into a real photograph. &
Forensic analysis reveals two distinct regions. Part was generated by FLUX.1 (high confidence), while the rest is a genuine photo from a Sony A7 III. The Sony A7 III region was downscaled to ${\sim}$ 52\% of original size. JPEG quality ranges from ${\sim}$90 to 93. Both synthetic and authentic traces in different regions point to targeted AI manipulation. \\
\bottomrule
\end{tabular}
\end{table}

\begin{table}[p]
\centering
\caption{Qualitative examples from FKG-50K: \textbf{AI Editing} (inpainting-based object replacement, insertion, and removal). The ground-truth mask marks the AI-modified region. (cont.)}
\smallerr
\setlength{\tabcolsep}{2pt}
\renewcommand{\arraystretch}{1.1}
\begin{tabular}{
	>{\centering\arraybackslash}m{17mm}
	>{\centering\arraybackslash}m{17mm}
	>{\centering\arraybackslash}m{17mm}
	>{\noindent\justifying\arraybackslash}m{20mm}
	>{\noindent\justifying\arraybackslash}m{48mm}
}
\toprule
\textbf{Image} & \textbf{GT Mask} & \textbf{Pred Mask} & \textbf{Decision} & \textbf{Justification} \\
\midrule
\rowcolor{gray!30}
\timg{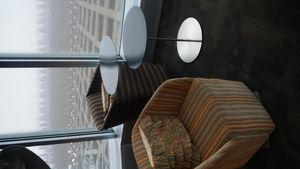} &
\timg{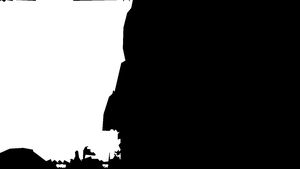} &
\timg{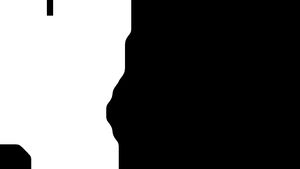} &
This image combines AI-generated content from FLUX.1 with an authentic photograph. &
The image separates into two forensically different regions. The manipulated region was synthesized by FLUX.1 (high confidence), while the host was captured by a Sony A7 III. JPEG quality ranges from ${\sim}$90 to 92. The coexistence of machine-generated and camera-originated content is characteristic of AI-assisted editing. \\
\timg{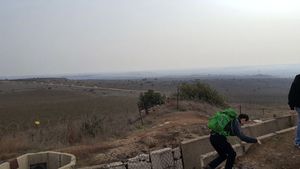} &
\timg{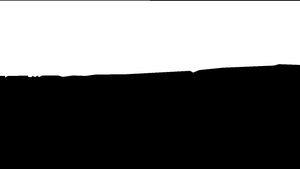} &
\timg{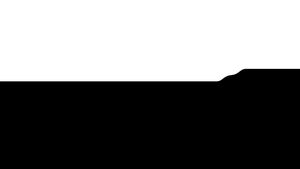} &
Localized AI editing found , FLUX.1 was used to generate or modify a portion of this otherwise authentic image. &
Two regions with markedly different forensic characteristics were found. A portion was created using FLUX.1 (high confidence), while the surrounding content originates from a iPhone 6. The iPhone 6 region was downscaled to ${\sim}$ 51\% of original size. JPEG quality ranges from ${\sim}$91 to 92. This combination of artificial and natural signatures is typical of AI inpainting. \\
\rowcolor{gray!30}
\timg{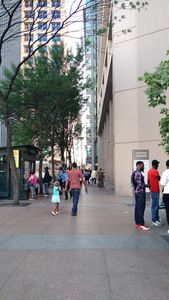} &
\timg{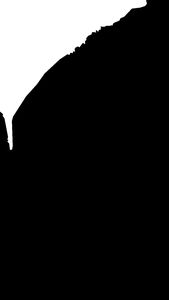} &
\timg{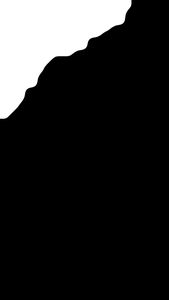} &
Part of this image has been synthesized by Midjourney v6 and blended into a genuine photograph. &
Two areas with distinctly different origins were found. One exhibits Midjourney v6-generated content (low confidence), whereas the unaltered portion was photographed with a Sony A7 III. The Sony A7 III region was downscaled to ${\sim}$ 51\% of original size. JPEG quality ranges from ${\sim}$89 to 91. The contrast between AI-synthesized and naturally captured regions reveals generative editing was applied. \\
\timg{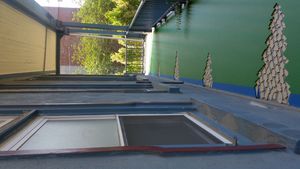} &
\timg{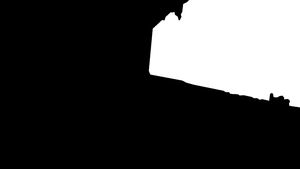} &
\timg{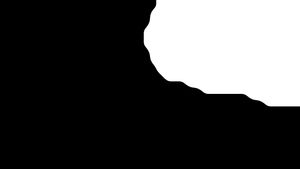} &
This image has been partially edited using AI , a portion was generated by FLUX.1 while the rest is an authentic photograph. &
Two separate regions were detected. One was produced by FLUX.1 (high), while the rest is an authentic photograph from a iPhone 6. The iPhone 6 region was downscaled to ${\sim}$ 51\% of original size. JPEG quality ranges from ${\sim}$91 to 93. The mixture of AI-generated and camera-captured content indicates localized AI editing. \\
\rowcolor{gray!30}
\timg{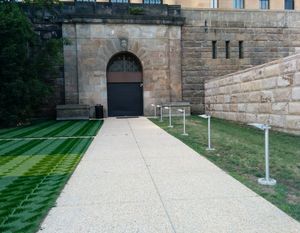} &
\timg{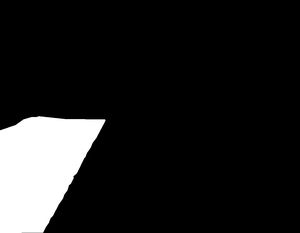} &
\timg{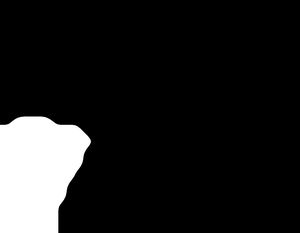} &
AI-assisted editing detected: part of this image was created by FLUX.1, inserted into a real photograph. &
Forensic analysis reveals two distinct regions. Part was generated by FLUX.1 (high confidence), while the rest is a genuine photo from a iPhone 6. The iPhone 6 region was downscaled to ${\sim}$ 52\% of original size. JPEG quality ranges from ${\sim}$91 to 92. Both synthetic and authentic traces in different regions point to targeted AI manipulation. \\
\timg{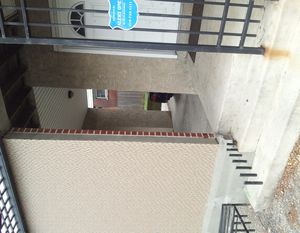} &
\timg{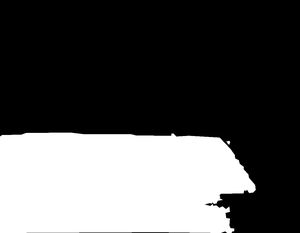} &
\timg{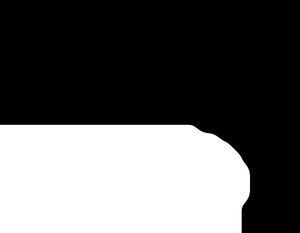} &
This image combines AI-generated content from FLUX.1 with an authentic photograph. &
The image separates into two forensically different regions. The manipulated region was synthesized by FLUX.1 (moderate confidence), while the host was captured by a iPhone 6. The iPhone 6 region was downscaled to ${\sim}$ 51\% of original size. JPEG quality ranges from ${\sim}$90 to 92. The coexistence of machine-generated and camera-originated content is characteristic of AI-assisted editing. \\
\bottomrule
\end{tabular}
\end{table}

\begin{table}[p]
\centering
\caption{Qualitative examples from FKG-50K: \textbf{AI Editing} (inpainting-based object replacement, insertion, and removal). The ground-truth mask marks the AI-modified region. (cont.)}
\smallerr
\setlength{\tabcolsep}{2pt}
\renewcommand{\arraystretch}{1.1}
\begin{tabular}{
	>{\centering\arraybackslash}m{17mm}
	>{\centering\arraybackslash}m{17mm}
	>{\centering\arraybackslash}m{17mm}
	>{\noindent\justifying\arraybackslash}m{20mm}
	>{\noindent\justifying\arraybackslash}m{48mm}
}
\toprule
\textbf{Image} & \textbf{GT Mask} & \textbf{Pred Mask} & \textbf{Decision} & \textbf{Justification} \\
\midrule
\rowcolor{gray!30}
\timg{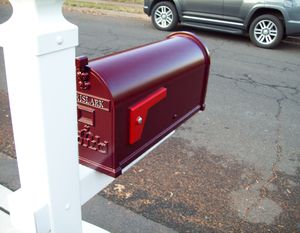} &
\timg{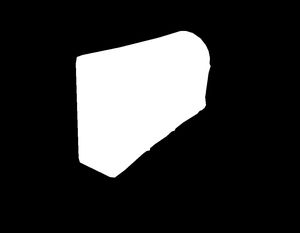} &
\timg{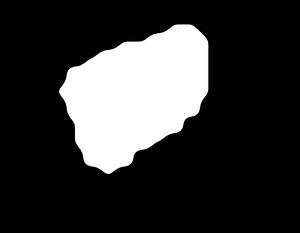} &
Localized AI editing found , Stable Diffusion XL was used to generate or modify a portion of this otherwise authentic image. &
Two regions with markedly different forensic characteristics were found. A portion was created using Stable Diffusion XL (moderate confidence), while the surrounding content originates from a Canon Eos 6d. JPEG quality ranges from ${\sim}$91 to 93. This combination of artificial and natural signatures is typical of AI inpainting. \\
\timg{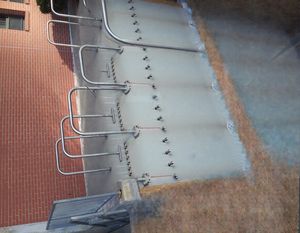} &
\timg{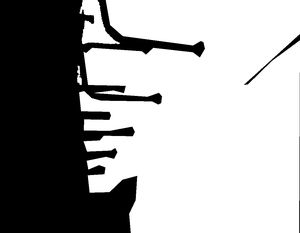} &
\timg{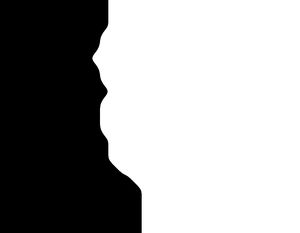} &
Part of this image has been synthesized by Stable Diffusion 3 Medium and blended into a genuine photograph. &
Two areas with distinctly different origins were found. One exhibits Stable Diffusion 3 Medium-generated content (high confidence), whereas the unaltered portion was photographed with a Canon EOS 5D Mark III. JPEG quality ranges from ${\sim}$90 to 92. The contrast between AI-synthesized and naturally captured regions reveals generative editing was applied. \\
\rowcolor{gray!30}
\timg{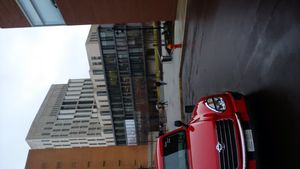} &
\timg{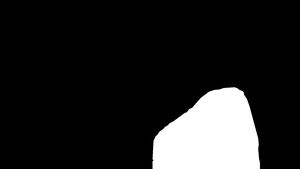} &
\timg{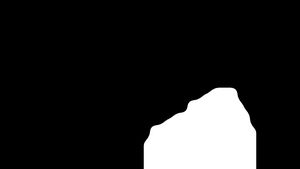} &
This image has been partially edited using AI , a portion was generated by FLUX.1 while the rest is an authentic photograph. &
Two separate regions were detected. One was produced by FLUX.1 (moderate), while the rest is an authentic photograph from a Sony A7 III. The Sony A7 III region was downscaled to ${\sim}$ 54\% of original size. Both regions share JPEG quality ${\sim}$ 92. The mixture of AI-generated and camera-captured content indicates localized AI editing. \\
\timg{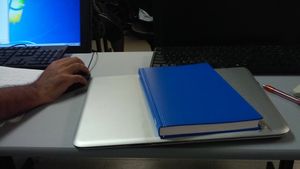} &
\timg{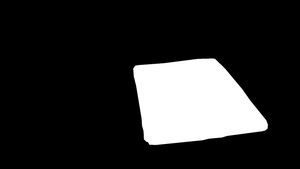} &
\timg{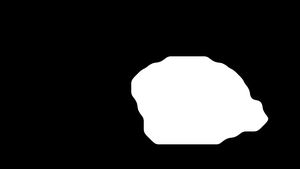} &
AI-assisted editing detected: part of this image was created by FLUX.1, inserted into a real photograph. &
Forensic analysis reveals two distinct regions. Part was generated by FLUX.1 (high confidence), while the rest is a genuine photo from a iPhone 5s. The iPhone 5s region was downscaled to ${\sim}$ 59\% of original size. JPEG quality ranges from ${\sim}$90 to 92. Both synthetic and authentic traces in different regions point to targeted AI manipulation. \\
\rowcolor{gray!30}
\timg{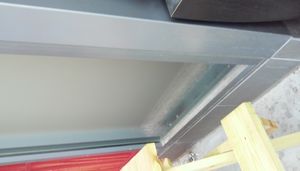} &
\timg{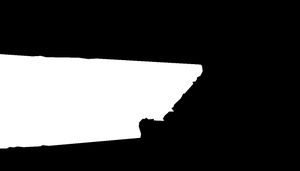} &
\timg{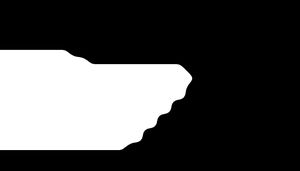} &
This image combines AI-generated content from Stable Diffusion XL with an authentic photograph. &
The image separates into two forensically different regions. The manipulated region was synthesized by Stable Diffusion XL (high confidence), while the host was captured by a iPhone 5s. The iPhone 5s region was downscaled to ${\sim}$ 57\% of original size. Both regions share JPEG quality ${\sim}$ 92. The coexistence of machine-generated and camera-originated content is characteristic of AI-assisted editing. \\
\timg{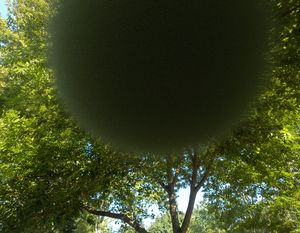} &
\timg{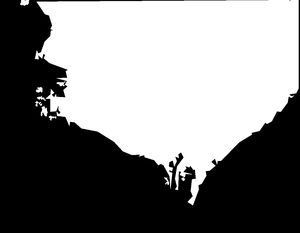} &
\timg{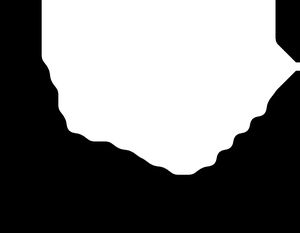} &
Localized AI editing found , ProGAN was used to generate or modify a portion of this otherwise authentic image. &
Two regions with markedly different forensic characteristics were found. A portion was created using ProGAN (moderate confidence), while the surrounding content originates from a Sony A7 III. JPEG quality ranges from ${\sim}$91 to 94. This combination of artificial and natural signatures is typical of AI inpainting. \\
\bottomrule
\end{tabular}
\end{table}

\begin{table}[p]
\centering
\caption{Qualitative examples from FKG-50K: \textbf{AI Editing} (inpainting-based object replacement, insertion, and removal). The ground-truth mask marks the AI-modified region. (cont.)}
\smallerr
\setlength{\tabcolsep}{2pt}
\renewcommand{\arraystretch}{1.1}
\begin{tabular}{
	>{\centering\arraybackslash}m{17mm}
	>{\centering\arraybackslash}m{17mm}
	>{\centering\arraybackslash}m{17mm}
	>{\noindent\justifying\arraybackslash}m{20mm}
	>{\noindent\justifying\arraybackslash}m{48mm}
}
\toprule
\textbf{Image} & \textbf{GT Mask} & \textbf{Pred Mask} & \textbf{Decision} & \textbf{Justification} \\
\midrule
\rowcolor{gray!30}
\timg{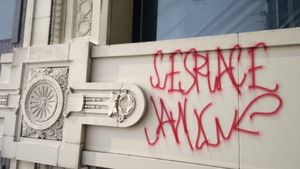} &
\timg{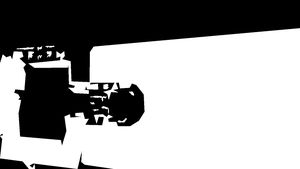} &
\timg{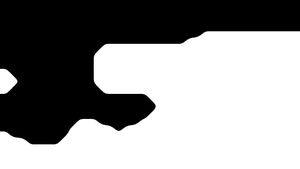} &
Part of this image has been synthesized by FLUX.1 and blended into a genuine photograph. &
Two areas with distinctly different origins were found. One exhibits FLUX.1-generated content (high confidence), whereas the unaltered portion was photographed with a Sony A7 III. The Sony A7 III region was downscaled to ${\sim}$ 54\% of original size. JPEG quality ranges from ${\sim}$91 to 92. The contrast between AI-synthesized and naturally captured regions reveals generative editing was applied. \\
\timg{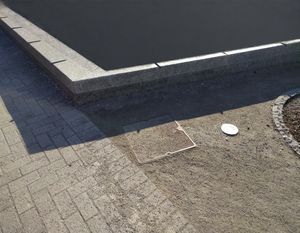} &
\timg{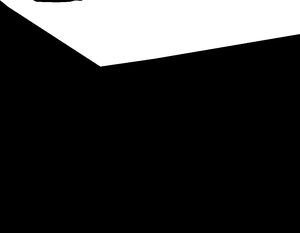} &
\timg{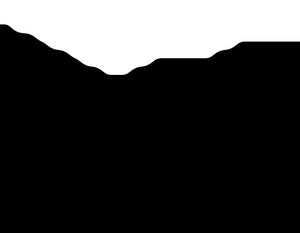} &
This image has been partially edited using AI , a portion was generated by FLUX.1 while the rest is an authentic photograph. &
Two separate regions were detected. One was produced by FLUX.1 (high), while the rest is an authentic photograph from a Sony A7 III. Both regions share JPEG quality ${\sim}$ 92. The mixture of AI-generated and camera-captured content indicates localized AI editing. \\
\rowcolor{gray!30}
\timg{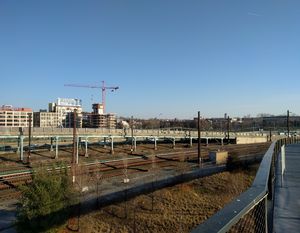} &
\timg{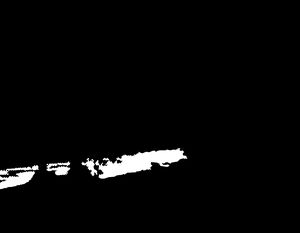} &
\timg{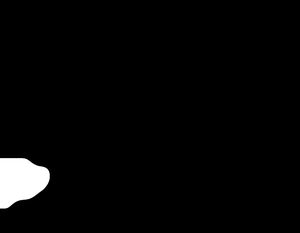} &
AI-assisted editing detected: part of this image was created by Midjourney v6, inserted into a real photograph. &
Forensic analysis reveals two distinct regions. Part was generated by Midjourney v6 (low confidence), while the rest is a genuine photo from a Sony A7 III. Both regions share JPEG quality ${\sim}$ 93. Both synthetic and authentic traces in different regions point to targeted AI manipulation. \\
\timg{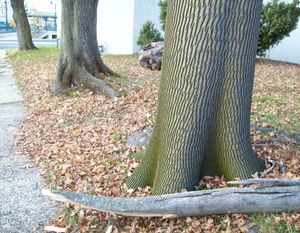} &
\timg{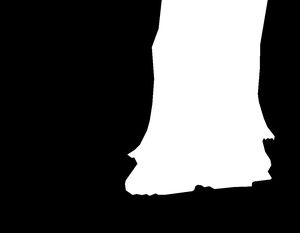} &
\timg{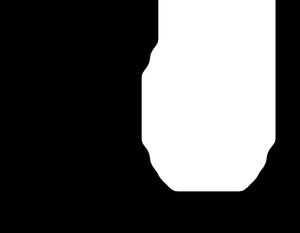} &
\red{This image is entirely AI-generated, most likely produced by Stable Diffusion XL.} &
\red{Although two regions were identified, both exhibit AI-generated characteristics rather than natural camera traces. The content is attributed to Stable Diffusion XL (high confidence). Saved at JPEG quality ${\sim}$ 92. No authentic camera-captured content was found.} \\
\rowcolor{gray!30}
\timg{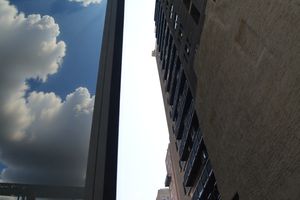} &
\timg{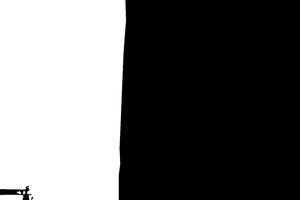} &
\timg{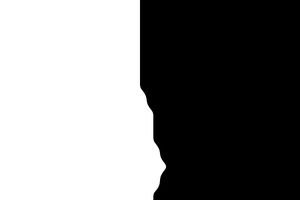} &
\red{This image is entirely AI-generated, most likely produced by FLUX.1.} &
\red{Although two regions were identified, both exhibit AI-generated characteristics rather than natural camera traces. The content is attributed to FLUX.1 (high confidence). Saved at JPEG quality ${\sim}$ 93. No authentic camera-captured content was found.} \\
\timg{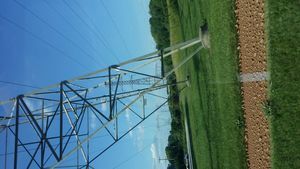} &
\timg{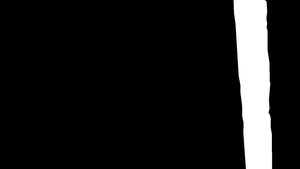} &
\timg{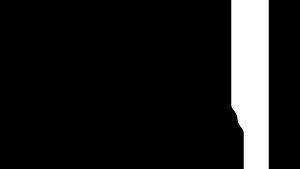} &
\red{Splicing is present in this image , two forensically distinct camera sources were identified.} &
\red{The image separates into two forensically different regions. One portion was taken with a iPhone 6, while the rest matches a iPhone 5s. The iPhone 6 region was downscaled to ${\sim}$ 51\% of original size. Two camera-specific forensic signatures in different parts is a hallmark of splicing.} \\
\bottomrule
\end{tabular}
\end{table}

\begin{table}[p]
\centering
\caption{Qualitative examples from FKG-50K: \textbf{AI Editing} (inpainting-based object replacement, insertion, and removal). The ground-truth mask marks the AI-modified region. (cont.)}
\smallerr
\setlength{\tabcolsep}{2pt}
\renewcommand{\arraystretch}{1.1}
\begin{tabular}{
	>{\centering\arraybackslash}m{17mm}
	>{\centering\arraybackslash}m{17mm}
	>{\centering\arraybackslash}m{17mm}
	>{\noindent\justifying\arraybackslash}m{20mm}
	>{\noindent\justifying\arraybackslash}m{48mm}
}
\toprule
\textbf{Image} & \textbf{GT Mask} & \textbf{Pred Mask} & \textbf{Decision} & \textbf{Justification} \\
\midrule
\rowcolor{gray!30}
\timg{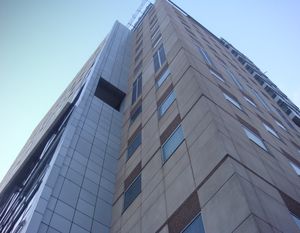} &
\timg{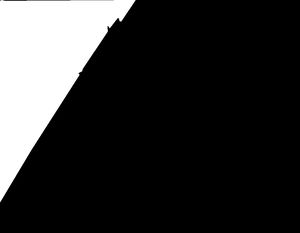} &
\timg{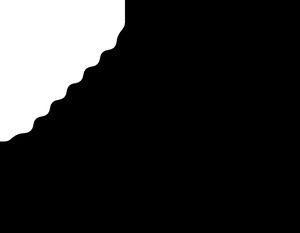} &
\red{This image is entirely AI-generated, most likely produced by Stable Diffusion XL.} &
\red{Although two regions were identified, both exhibit AI-generated characteristics rather than natural camera traces. The content is attributed to Stable Diffusion XL (high confidence). Saved at JPEG quality ${\sim}$ 90. No authentic camera-captured content was found.} \\
\bottomrule
\end{tabular}
\end{table}

\begin{table}[p]
\centering
\caption{Qualitative examples from FKG-50K: \textbf{Full Synthesis}. These images are entirely AI-generated. The predicted mask is blank (all zeros) because the system correctly identifies a single region with consistent source and processing throughout, rather than multiple regions with differing sources.}
\label{tab:qual_full_synthesis}
\smallerr
\setlength{\tabcolsep}{2pt}
\renewcommand{\arraystretch}{1.1}
\begin{tabular}{
	>{\centering\arraybackslash}m{17mm}
	>{\centering\arraybackslash}m{17mm}
	>{\noindent\justifying\arraybackslash}m{24mm}
	>{\noindent\justifying\arraybackslash}m{55mm}
}
\toprule
\textbf{Image} & \textbf{Pred Mask} & \textbf{Decision} & \textbf{Justification} \\
\midrule
\rowcolor{gray!30}
\timg{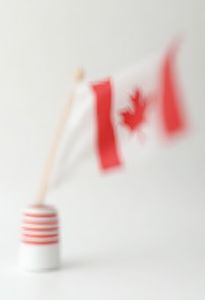} &
\timg{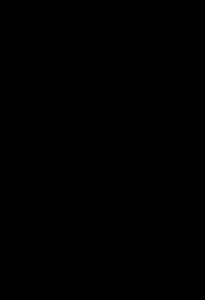} &
This is a fully synthetic image produced by an AI model. &
Forensic analysis reveals no camera-specific traces anywhere in the image. Instead, the characteristics are consistent with FLUX.1 (high confidence), indicating the entire image is synthetically generated. No JPEG compression was detected. \\
\timg{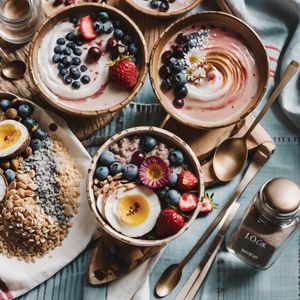} &
\timg{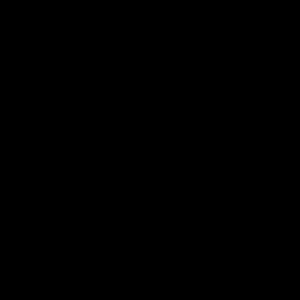} &
Forensic analysis identifies this as an AI-generated image, not a real photograph. &
The image lacks the forensic fingerprints left by physical camera sensors. Analysis points to Stable Diffusion XL as the source (high confidence), confirming full synthesis. No JPEG compression was detected. \\
\rowcolor{gray!30}
\timg{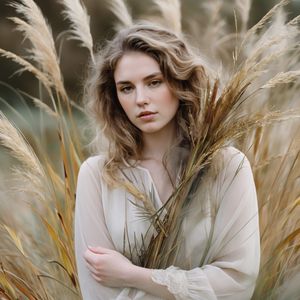} &
\timg{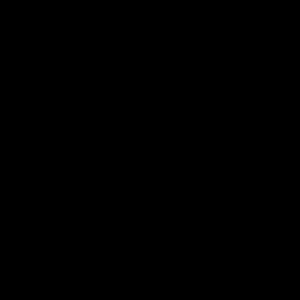} &
The image is synthetic , it was produced entirely by artificial intelligence. &
The entire image carries forensic characteristics typical of AI-generated content rather than real photography. With high confidence, the image was produced by Stable Diffusion XL. No JPEG compression was detected. \\
\timg{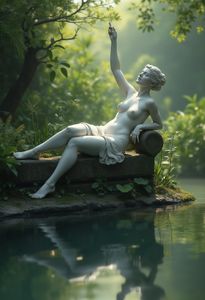} &
\timg{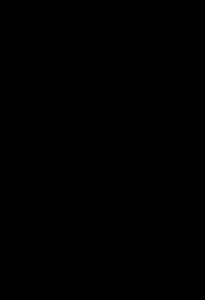} &
This image is entirely AI-generated. &
Every region of the image exhibits patterns associated with AI generation rather than optical capture. The content is attributed to FLUX.1 with high confidence. No JPEG compression was detected. \\
\rowcolor{gray!30}
\timg{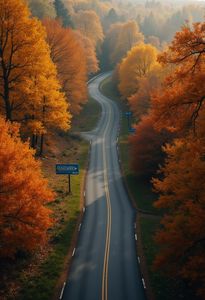} &
\timg{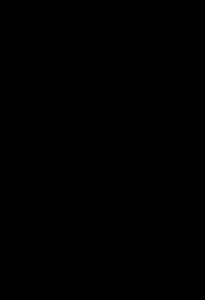} &
No real camera captured this image , it was generated by AI. &
Rather than the sensor noise and optical artifacts typical of real photographs, this image displays the hallmarks of AI generation. It is identified as having been created by FLUX.1 with high confidence. No JPEG compression was detected. \\
\timg{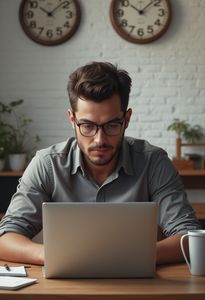} &
\timg{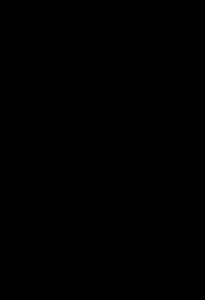} &
This image was created by an AI image generator, not captured by a camera. &
Forensic analysis reveals no camera-specific traces anywhere in the image. Instead, the characteristics are consistent with FLUX.1 (high confidence), indicating the entire image is synthetically generated. No JPEG compression was detected. \\
\bottomrule
\end{tabular}
\end{table}

\begin{table}[p]
\centering
\caption{Qualitative examples from FKG-50K: \textbf{Full Synthesis}. These images are entirely AI-generated. The predicted mask is blank (all zeros) because the system correctly identifies a single region with consistent source and processing throughout, rather than multiple regions with differing sources. (cont.)}
\smallerr
\setlength{\tabcolsep}{2pt}
\renewcommand{\arraystretch}{1.1}
\begin{tabular}{
	>{\centering\arraybackslash}m{17mm}
	>{\centering\arraybackslash}m{17mm}
	>{\noindent\justifying\arraybackslash}m{24mm}
	>{\noindent\justifying\arraybackslash}m{55mm}
}
\toprule
\textbf{Image} & \textbf{Pred Mask} & \textbf{Decision} & \textbf{Justification} \\
\midrule
\rowcolor{gray!30}
\timg{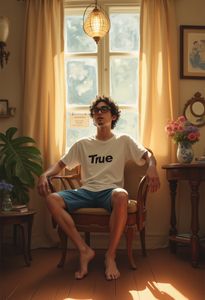} &
\timg{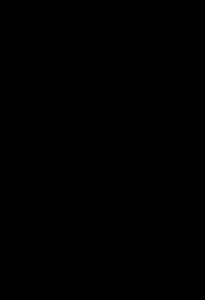} &
Analysis confirms this image is AI-generated rather than photographed. &
The image lacks the forensic fingerprints left by physical camera sensors. Analysis points to FLUX.1 as the source (high confidence), confirming full synthesis. No JPEG compression was detected. \\
\timg{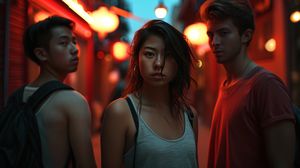} &
\timg{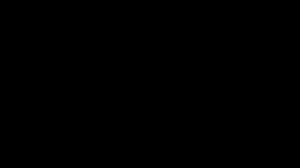} &
This is a fully synthetic image produced by an AI model. &
The entire image carries forensic characteristics typical of AI-generated content rather than real photography. With high confidence, the image was produced by Midjourney v6. No JPEG compression was detected. \\
\rowcolor{gray!30}
\timg{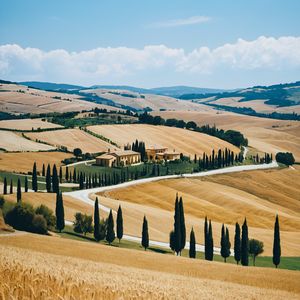} &
\timg{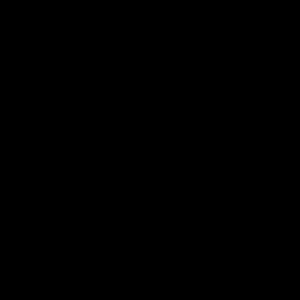} &
Forensic analysis identifies this as an AI-generated image, not a real photograph. &
Every region of the image exhibits patterns associated with AI generation rather than optical capture. The content is attributed to Stable Diffusion XL with high confidence. No JPEG compression was detected. \\
\timg{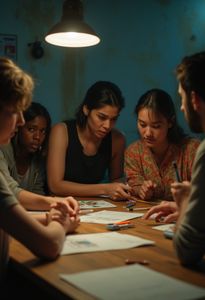} &
\timg{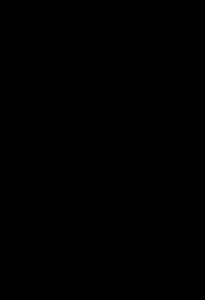} &
The image is synthetic , it was produced entirely by artificial intelligence. &
Rather than the sensor noise and optical artifacts typical of real photographs, this image displays the hallmarks of AI generation. It is identified as having been created by FLUX.1 with high confidence. No JPEG compression was detected. \\
\rowcolor{gray!30}
\timg{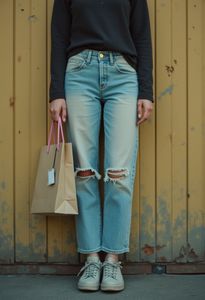} &
\timg{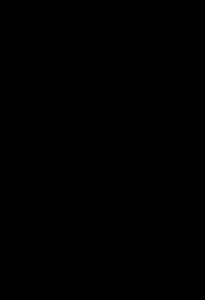} &
This image is entirely AI-generated. &
Forensic analysis reveals no camera-specific traces anywhere in the image. Instead, the characteristics are consistent with FLUX.1 (high confidence), indicating the entire image is synthetically generated. No JPEG compression was detected. \\
\timg{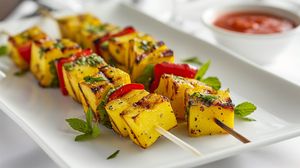} &
\timg{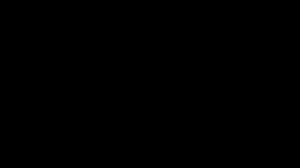} &
No real camera captured this image , it was generated by AI. &
The image lacks the forensic fingerprints left by physical camera sensors. Analysis points to Midjourney v6 as the source (high confidence), confirming full synthesis. No JPEG compression was detected. \\
\bottomrule
\end{tabular}
\end{table}

\begin{table}[p]
\centering
\caption{Qualitative examples from FKG-50K: \textbf{Full Synthesis}. These images are entirely AI-generated. The predicted mask is blank (all zeros) because the system correctly identifies a single region with consistent source and processing throughout, rather than multiple regions with differing sources. (cont.)}
\smallerr
\setlength{\tabcolsep}{2pt}
\renewcommand{\arraystretch}{1.1}
\begin{tabular}{
	>{\centering\arraybackslash}m{17mm}
	>{\centering\arraybackslash}m{17mm}
	>{\noindent\justifying\arraybackslash}m{24mm}
	>{\noindent\justifying\arraybackslash}m{55mm}
}
\toprule
\textbf{Image} & \textbf{Pred Mask} & \textbf{Decision} & \textbf{Justification} \\
\midrule
\rowcolor{gray!30}
\timg{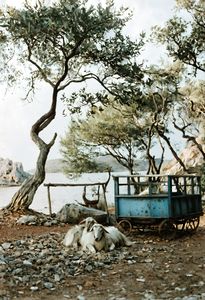} &
\timg{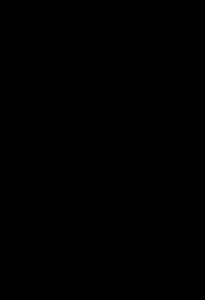} &
This image was created by an AI image generator, not captured by a camera. &
The entire image carries forensic characteristics typical of AI-generated content rather than real photography. With high confidence, the image was produced by Stable Diffusion XL. No JPEG compression was detected. \\
\timg{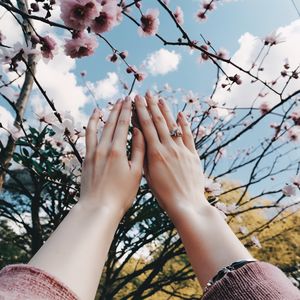} &
\timg{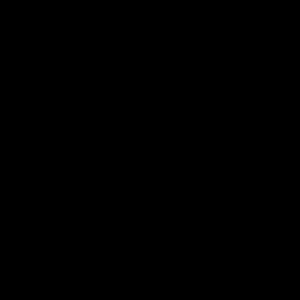} &
Analysis confirms this image is AI-generated rather than photographed. &
Every region of the image exhibits patterns associated with AI generation rather than optical capture. The content is attributed to Stable Diffusion XL with high confidence. No JPEG compression was detected. \\
\rowcolor{gray!30}
\timg{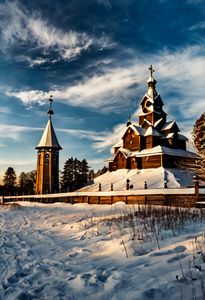} &
\timg{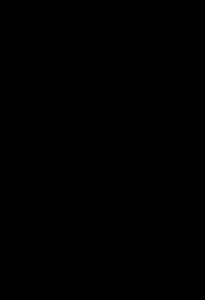} &
This is a fully synthetic image produced by an AI model. &
Rather than the sensor noise and optical artifacts typical of real photographs, this image displays the hallmarks of AI generation. It is identified as having been created by Stable Diffusion XL with high confidence. No JPEG compression was detected. \\
\timg{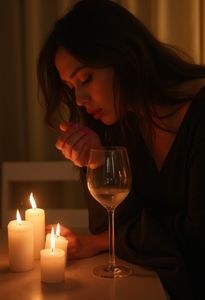} &
\timg{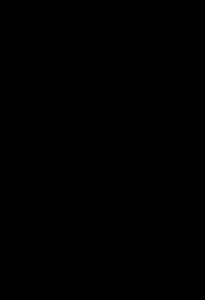} &
Forensic analysis identifies this as an AI-generated image, not a real photograph. &
Forensic analysis reveals no camera-specific traces anywhere in the image. Instead, the characteristics are consistent with FLUX.1 (high confidence), indicating the entire image is synthetically generated. No JPEG compression was detected. \\
\rowcolor{gray!30}
\timg{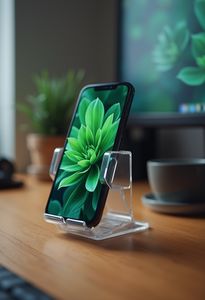} &
\timg{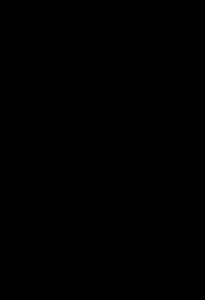} &
The image is synthetic , it was produced entirely by artificial intelligence. &
The image lacks the forensic fingerprints left by physical camera sensors. Analysis points to FLUX.1 as the source (high confidence), confirming full synthesis. No JPEG compression was detected. \\
\timg{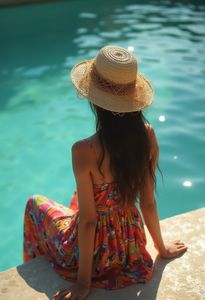} &
\timg{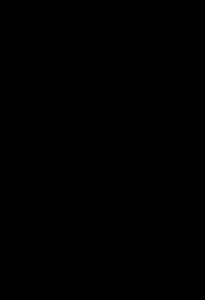} &
This image is entirely AI-generated. &
The entire image carries forensic characteristics typical of AI-generated content rather than real photography. With high confidence, the image was produced by FLUX.1. No JPEG compression was detected. \\
\bottomrule
\end{tabular}
\end{table}

\begin{table}[p]
\centering
\caption{Qualitative examples from FKG-50K: \textbf{Full Synthesis}. These images are entirely AI-generated. The predicted mask is blank (all zeros) because the system correctly identifies a single region with consistent source and processing throughout, rather than multiple regions with differing sources. (cont.)}
\smallerr
\setlength{\tabcolsep}{2pt}
\renewcommand{\arraystretch}{1.1}
\begin{tabular}{
	>{\centering\arraybackslash}m{17mm}
	>{\centering\arraybackslash}m{17mm}
	>{\noindent\justifying\arraybackslash}m{24mm}
	>{\noindent\justifying\arraybackslash}m{55mm}
}
\toprule
\textbf{Image} & \textbf{Pred Mask} & \textbf{Decision} & \textbf{Justification} \\
\midrule
\rowcolor{gray!30}
\timg{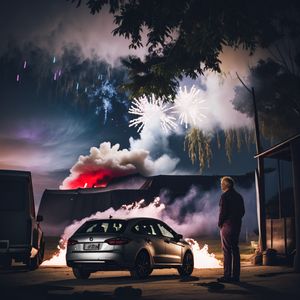} &
\timg{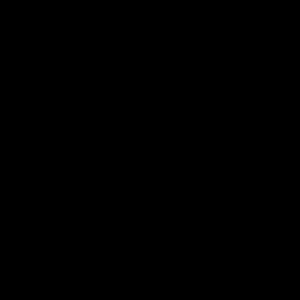} &
No real camera captured this image , it was generated by AI. &
Every region of the image exhibits patterns associated with AI generation rather than optical capture. The content is attributed to Stable Diffusion XL with high confidence. No JPEG compression was detected. \\
\timg{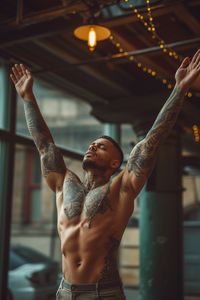} &
\timg{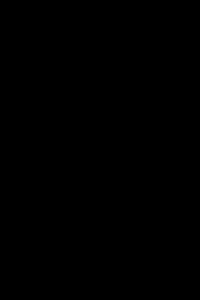} &
This image was created by an AI image generator, not captured by a camera. &
Rather than the sensor noise and optical artifacts typical of real photographs, this image displays the hallmarks of AI generation. It is identified as having been created by Midjourney v6 with high confidence. No JPEG compression was detected. \\
\rowcolor{gray!30}
\timg{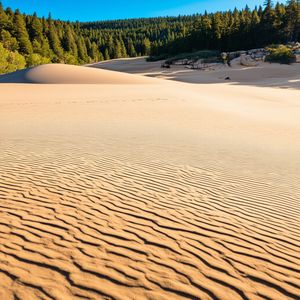} &
\timg{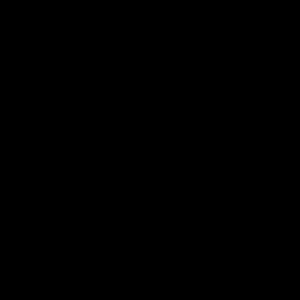} &
Analysis confirms this image is AI-generated rather than photographed. &
Forensic analysis reveals no camera-specific traces anywhere in the image. Instead, the characteristics are consistent with Stable Diffusion 3 (high confidence), indicating the entire image is synthetically generated. No JPEG compression was detected. \\
\timg{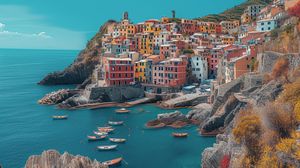} &
\timg{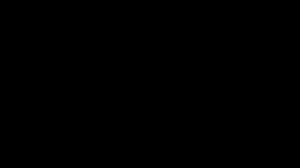} &
This is a fully synthetic image produced by an AI model. &
The image lacks the forensic fingerprints left by physical camera sensors. Analysis points to Midjourney v6 as the source (high confidence), confirming full synthesis. No JPEG compression was detected. \\
\rowcolor{gray!30}
\timg{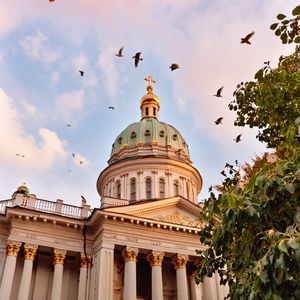} &
\timg{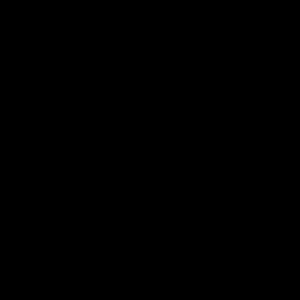} &
Forensic analysis identifies this as an AI-generated image, not a real photograph. &
The entire image carries forensic characteristics typical of AI-generated content rather than real photography. With high confidence, the image was produced by Stable Diffusion XL. No JPEG compression was detected. \\
\timg{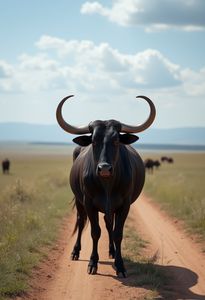} &
\timg{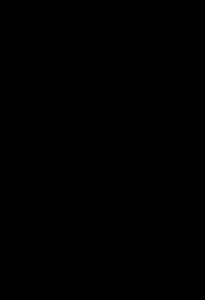} &
The image is synthetic , it was produced entirely by artificial intelligence. &
Every region of the image exhibits patterns associated with AI generation rather than optical capture. The content is attributed to FLUX.1 with high confidence. No JPEG compression was detected. \\
\bottomrule
\end{tabular}
\end{table}

\begin{table}[p]
\centering
\caption{Qualitative examples from FKG-50K: \textbf{Full Synthesis}. These images are entirely AI-generated. The predicted mask is blank (all zeros) because the system correctly identifies a single region with consistent source and processing throughout, rather than multiple regions with differing sources. (cont.)}
\smallerr
\setlength{\tabcolsep}{2pt}
\renewcommand{\arraystretch}{1.1}
\begin{tabular}{
	>{\centering\arraybackslash}m{17mm}
	>{\centering\arraybackslash}m{17mm}
	>{\noindent\justifying\arraybackslash}m{24mm}
	>{\noindent\justifying\arraybackslash}m{55mm}
}
\toprule
\textbf{Image} & \textbf{Pred Mask} & \textbf{Decision} & \textbf{Justification} \\
\midrule
\rowcolor{gray!30}
\timg{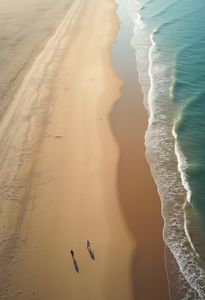} &
\timg{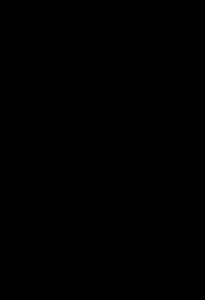} &
This image is entirely AI-generated. &
Rather than the sensor noise and optical artifacts typical of real photographs, this image displays the hallmarks of AI generation. It is identified as having been created by FLUX.1 with high confidence. No JPEG compression was detected. \\
\bottomrule
\end{tabular}
\end{table}

\begin{table}[p]
\centering
\caption{Qualitative examples from \textbf{DSO-1} (external dataset). DSO-1 contains splicing forgeries created by compositing regions from different camera sources. The ground-truth mask marks the spliced region.}
\label{tab:qual_dso-1}
\smallerr
\setlength{\tabcolsep}{2pt}
\renewcommand{\arraystretch}{1.1}
\begin{tabular}{
	>{\centering\arraybackslash}m{17mm}
	>{\centering\arraybackslash}m{17mm}
	>{\centering\arraybackslash}m{17mm}
	>{\noindent\justifying\arraybackslash}m{20mm}
	>{\noindent\justifying\arraybackslash}m{48mm}
}
\toprule
\textbf{Image} & \textbf{GT Mask} & \textbf{Pred Mask} & \textbf{Decision} & \textbf{Justification} \\
\midrule
\rowcolor{gray!30}
\timg{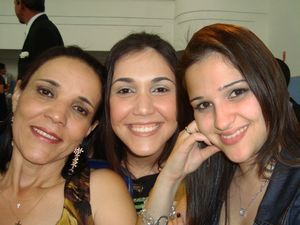} &
\timg{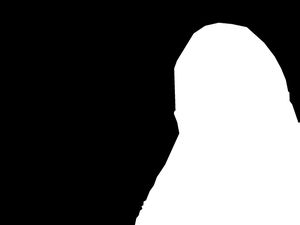} &
\timg{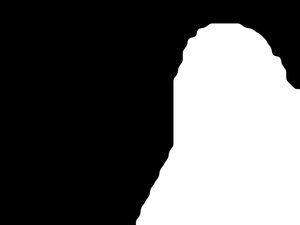} &
This image has been spliced , it contains content from two different cameras. &
Forensic analysis identifies two distinct regions in this image. One region is consistent with a Sony A6000, while the other traces back to a Hasselblad L1D-20c. The presence of two different camera signatures within one image is strong evidence of splicing. The image has been JPEG compressed at quality 89. \\
\timg{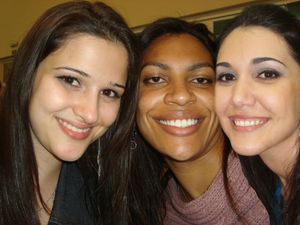} &
\timg{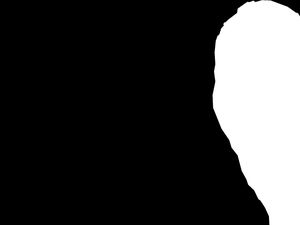} &
\timg{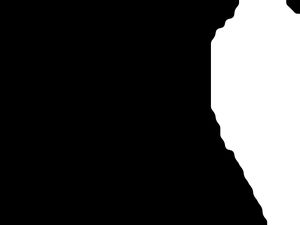} &
Image tampering detected: part of this image has been altered through traditional editing techniques. &
Forensic analysis detects two distinct regions in this image, suggesting localized manipulation. The dominant region is consistent with a Sony A6000. The edited area shows altered forensic characteristics compared to the rest of the image. This region also shows signs of being resized to approximately 114\% of its original resolution. The image has been JPEG compressed at quality 86. \\
\rowcolor{gray!30}
\timg{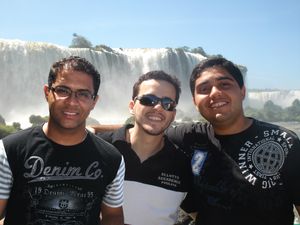} &
\timg{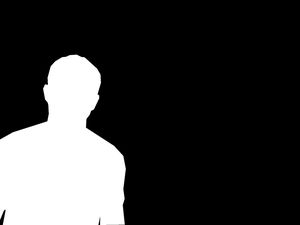} &
\timg{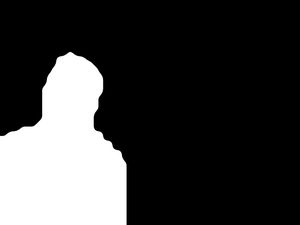} &
This image is a composite, combining photographs from two different camera sources. &
Forensic analysis identifies two distinct regions in this image. One region is consistent with a Canon EOS Rebel T3i, while the other originates from a different camera. The presence of two different camera signatures within one image is strong evidence of splicing. The image has been JPEG compressed at quality 90. \\
\timg{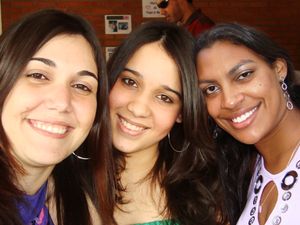} &
\timg{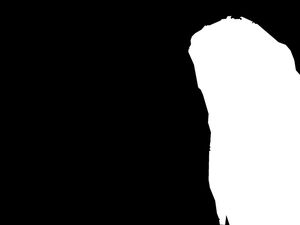} &
\timg{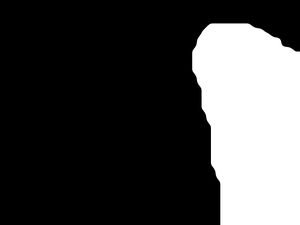} &
Splicing is present in this image , two forensically distinct camera sources were identified. &
Forensic analysis identifies two distinct regions in this image. One region is consistent with a Sony A6000, while the other traces back to a Hasselblad L1D-20c. The presence of two different camera signatures within one image is strong evidence of splicing. The image has been JPEG compressed at quality 87. \\
\rowcolor{gray!30}
\timg{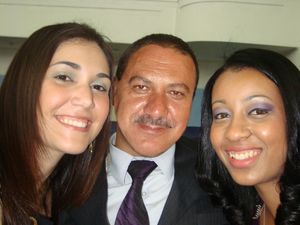} &
\timg{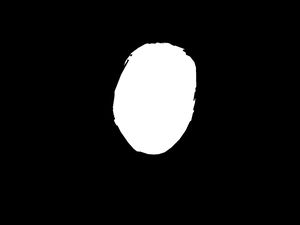} &
\timg{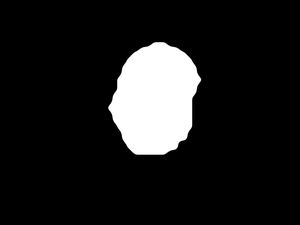} &
This image has been manipulated through splicing, merging content captured by different devices. &
Forensic analysis identifies two distinct regions in this image. One region is consistent with a Sony A6000, while the other traces back to a Hasselblad L1D-20c. The presence of two different camera signatures within one image is strong evidence of splicing. This region also shows signs of being resized to approximately 77\% of its original resolution. The image has been JPEG compressed at quality 89. \\
\timg{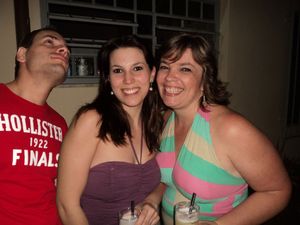} &
\timg{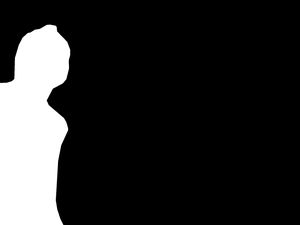} &
\timg{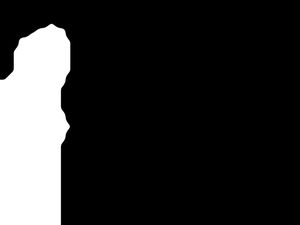} &
This image has been spliced , it contains content from two different cameras. &
Forensic analysis identifies two distinct regions in this image. One region appears to come from a digital camera, while the other traces back to a Canon EOS 5D Mark IV. The presence of two different camera signatures within one image is strong evidence of splicing. The image has been JPEG compressed at quality 86. \\
\bottomrule
\end{tabular}
\end{table}

\begin{table}[p]
\centering
\caption{Qualitative examples from \textbf{DSO-1} (external dataset). DSO-1 contains splicing forgeries created by compositing regions from different camera sources. The ground-truth mask marks the spliced region. (cont.)}
\smallerr
\setlength{\tabcolsep}{2pt}
\renewcommand{\arraystretch}{1.1}
\begin{tabular}{
	>{\centering\arraybackslash}m{17mm}
	>{\centering\arraybackslash}m{17mm}
	>{\centering\arraybackslash}m{17mm}
	>{\noindent\justifying\arraybackslash}m{20mm}
	>{\noindent\justifying\arraybackslash}m{48mm}
}
\toprule
\textbf{Image} & \textbf{GT Mask} & \textbf{Pred Mask} & \textbf{Decision} & \textbf{Justification} \\
\midrule
\rowcolor{gray!30}
\timg{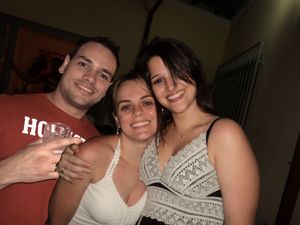} &
\timg{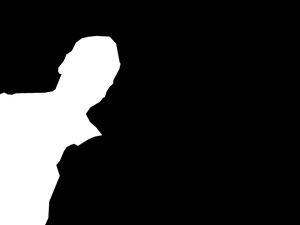} &
\timg{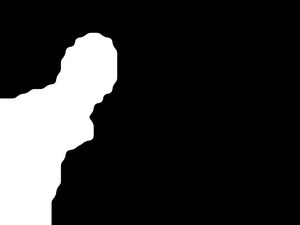} &
Image splicing detected: portions of this image originate from two separate cameras. &
Forensic analysis identifies two distinct regions in this image. One region is consistent with a Canon EOS 5D Mark IV, while the other originates from a different camera. The presence of two different camera signatures within one image is strong evidence of splicing. The image has been JPEG compressed at quality 92. \\
\timg{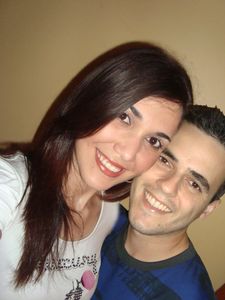} &
\timg{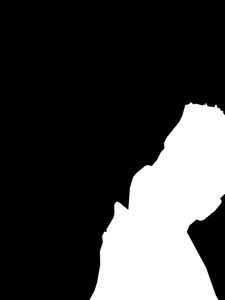} &
\timg{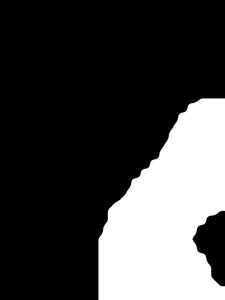} &
This image is a composite, combining photographs from two different camera sources. &
Forensic analysis identifies two distinct regions in this image. One region is consistent with a Sony A6000, while the other originates from a different camera. The presence of two different camera signatures within one image is strong evidence of splicing. This region also shows signs of being resized to approximately 80\% of its original resolution. The image has been JPEG compressed at quality 92. \\
\rowcolor{gray!30}
\timg{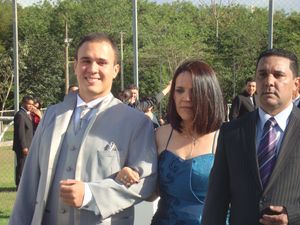} &
\timg{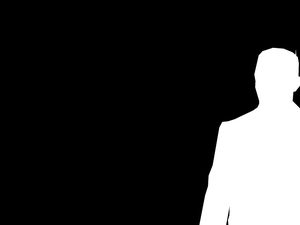} &
\timg{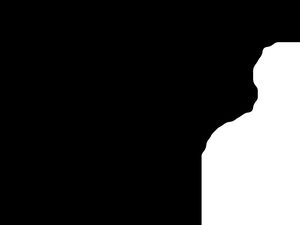} &
This image shows signs of manipulation , local edits have been applied to alter the original content. &
Forensic analysis detects two distinct regions in this image, suggesting localized manipulation. Both regions are attributed to a Sony A6000, but they exhibit different forensic characteristics, consistent with local editing or retouching. This region also shows signs of being resized to approximately 114\% of its original resolution. The image has been JPEG compressed at quality 88. \\
\timg{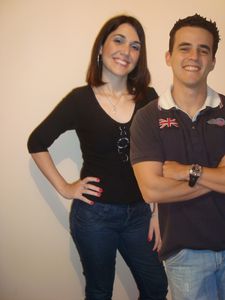} &
\timg{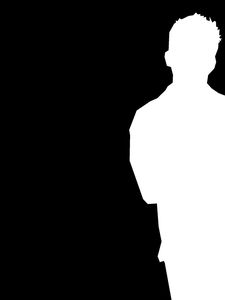} &
\timg{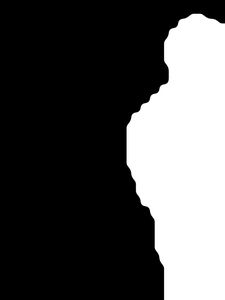} &
This image has been manipulated through splicing, merging content captured by different devices. &
Forensic analysis identifies two distinct regions in this image. One region appears to come from a digital camera, while the other traces back to a Hasselblad L1D-20c. The presence of two different camera signatures within one image is strong evidence of splicing. This region also shows signs of being resized to approximately 63\% of its original resolution. The image has been JPEG compressed at quality 92. \\
\rowcolor{gray!30}
\timg{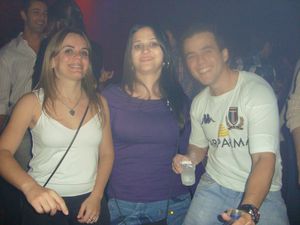} &
\timg{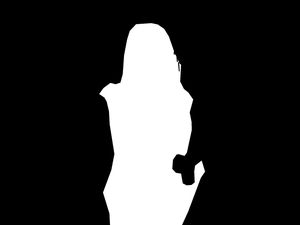} &
\timg{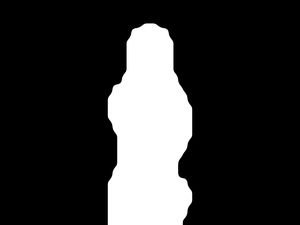} &
This image has been locally edited , forensic analysis detects manipulation within the image. &
Forensic analysis detects two distinct regions in this image, suggesting localized manipulation. Both regions are attributed to a Sony A6000, but they exhibit different forensic characteristics, consistent with local editing or retouching. The image has been JPEG compressed at quality 89. \\
\timg{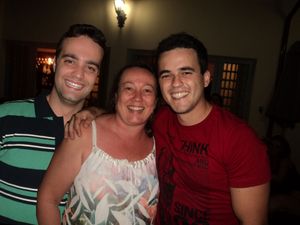} &
\timg{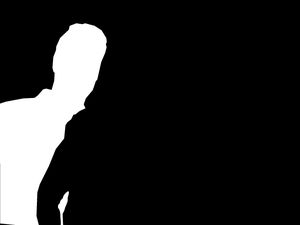} &
\timg{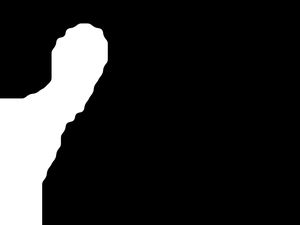} &
Image splicing detected: portions of this image originate from two separate cameras. &
Forensic analysis identifies two distinct regions in this image. One region is consistent with a Canon EOS 5D Mark IV, while the other originates from a different camera. The presence of two different camera signatures within one image is strong evidence of splicing. The image has been JPEG compressed at quality 91. \\
\bottomrule
\end{tabular}
\end{table}

\begin{table}[p]
\centering
\caption{Qualitative examples from \textbf{DSO-1} (external dataset). DSO-1 contains splicing forgeries created by compositing regions from different camera sources. The ground-truth mask marks the spliced region. (cont.)}
\smallerr
\setlength{\tabcolsep}{2pt}
\renewcommand{\arraystretch}{1.1}
\begin{tabular}{
	>{\centering\arraybackslash}m{17mm}
	>{\centering\arraybackslash}m{17mm}
	>{\centering\arraybackslash}m{17mm}
	>{\noindent\justifying\arraybackslash}m{20mm}
	>{\noindent\justifying\arraybackslash}m{48mm}
}
\toprule
\textbf{Image} & \textbf{GT Mask} & \textbf{Pred Mask} & \textbf{Decision} & \textbf{Justification} \\
\midrule
\rowcolor{gray!30}
\timg{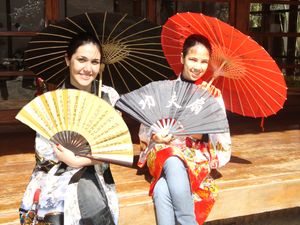} &
\timg{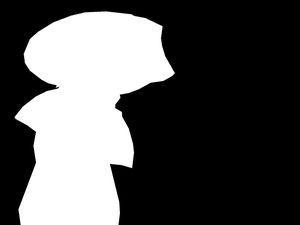} &
\timg{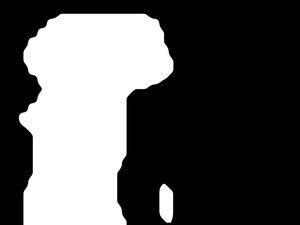} &
This image is a composite, combining photographs from two different camera sources. &
Forensic analysis identifies two distinct regions in this image. One region is consistent with a Sony A6000, while the other originates from a different camera. The presence of two different camera signatures within one image is strong evidence of splicing. This region also shows signs of being resized to approximately 91\% of its original resolution. The image has been JPEG compressed at quality 86. \\
\timg{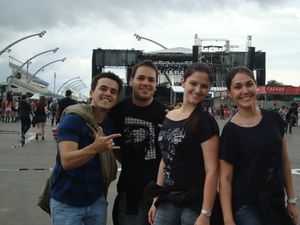} &
\timg{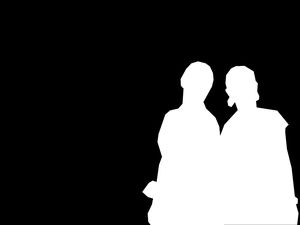} &
\timg{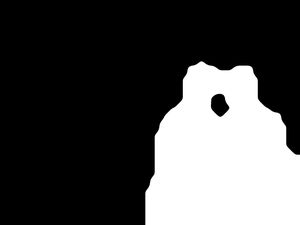} &
Splicing is present in this image , two forensically distinct camera sources were identified. &
Forensic analysis identifies two distinct regions in this image. One region is consistent with a Sony A6000, while the other originates from a different camera. The presence of two different camera signatures within one image is strong evidence of splicing. This region also shows signs of being resized to approximately 67\% of its original resolution. The image has been JPEG compressed at quality 88. \\
\rowcolor{gray!30}
\timg{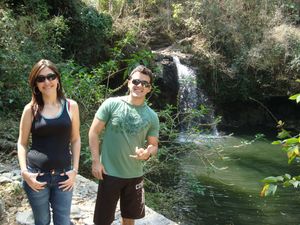} &
\timg{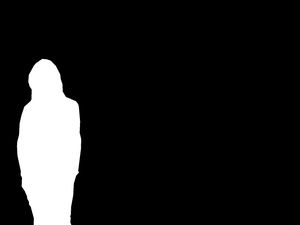} &
\timg{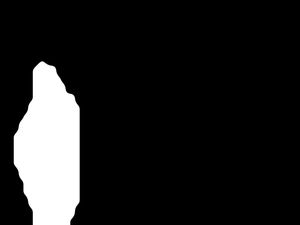} &
Tampering detected: the image has been modified through localized editing operations. &
Forensic analysis detects two distinct regions in this image, suggesting localized manipulation. Both regions are attributed to a Sony A6000, but they exhibit different forensic characteristics, consistent with local editing or retouching. The image has been JPEG compressed at quality 87. \\
\timg{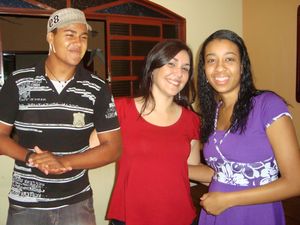} &
\timg{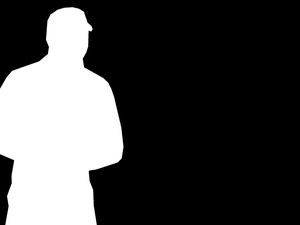} &
\timg{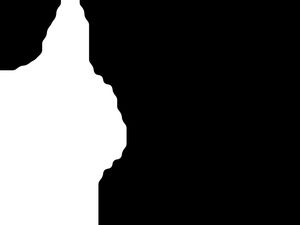} &
This image has been spliced , it contains content from two different cameras. &
Forensic analysis identifies two distinct regions in this image. One region is consistent with a Sony A6000, while the other originates from a different camera. The presence of two different camera signatures within one image is strong evidence of splicing. The image has been JPEG compressed at quality 88. \\
\rowcolor{gray!30}
\timg{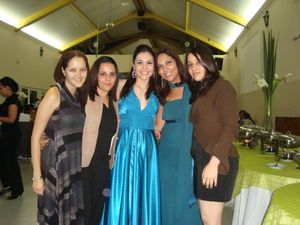} &
\timg{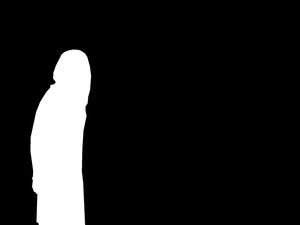} &
\timg{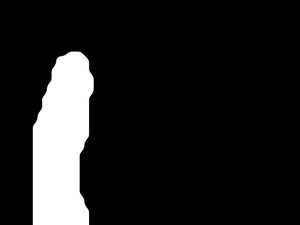} &
Image splicing detected: portions of this image originate from two separate cameras. &
Forensic analysis identifies two distinct regions in this image. One region appears to come from a digital camera, while the other traces back to a Sony A6000. The presence of two different camera signatures within one image is strong evidence of splicing. This region also shows signs of being resized to approximately 114\% of its original resolution. The image has been JPEG compressed at quality 89. \\
\timg{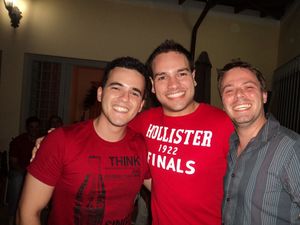} &
\timg{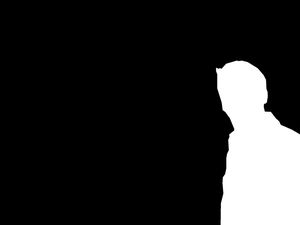} &
\timg{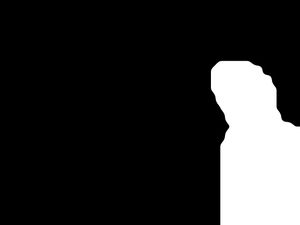} &
This image is a composite, combining photographs from two different camera sources. &
Forensic analysis identifies two distinct regions in this image. One region appears to come from a digital camera, while the other traces back to a Canon EOS 5D Mark IV. The presence of two different camera signatures within one image is strong evidence of splicing. The image has been JPEG compressed at quality 89. \\
\bottomrule
\end{tabular}
\end{table}

\begin{table}[p]
\centering
\caption{Qualitative examples from \textbf{DSO-1} (external dataset). DSO-1 contains splicing forgeries created by compositing regions from different camera sources. The ground-truth mask marks the spliced region. (cont.)}
\smallerr
\setlength{\tabcolsep}{2pt}
\renewcommand{\arraystretch}{1.1}
\begin{tabular}{
	>{\centering\arraybackslash}m{17mm}
	>{\centering\arraybackslash}m{17mm}
	>{\centering\arraybackslash}m{17mm}
	>{\noindent\justifying\arraybackslash}m{20mm}
	>{\noindent\justifying\arraybackslash}m{48mm}
}
\toprule
\textbf{Image} & \textbf{GT Mask} & \textbf{Pred Mask} & \textbf{Decision} & \textbf{Justification} \\
\midrule
\rowcolor{gray!30}
\timg{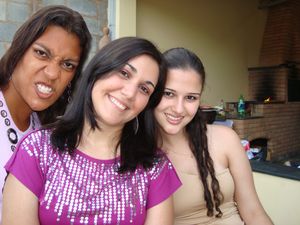} &
\timg{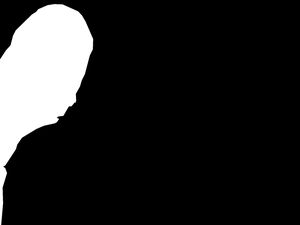} &
\timg{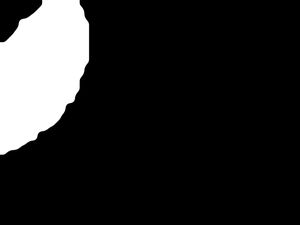} &
Splicing is present in this image , two forensically distinct camera sources were identified. &
Forensic analysis identifies two distinct regions in this image. One region is consistent with a Sony A6000, while the other originates from a different camera. The presence of two different camera signatures within one image is strong evidence of splicing. This region also shows signs of being resized to approximately 123\% of its original resolution. The image has been JPEG compressed at quality 88. \\
\timg{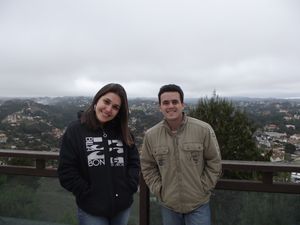} &
\timg{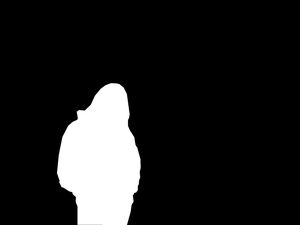} &
\timg{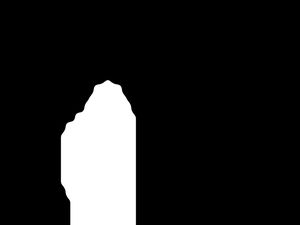} &
This image has been manipulated through splicing, merging content captured by different devices. &
Forensic analysis identifies two distinct regions in this image. One region is consistent with a iPhone 5s, while the other originates from a different camera. The presence of two different camera signatures within one image is strong evidence of splicing. This region also shows signs of being resized to approximately 58\% of its original resolution. The image has been JPEG compressed at quality 93. \\
\rowcolor{gray!30}
\timg{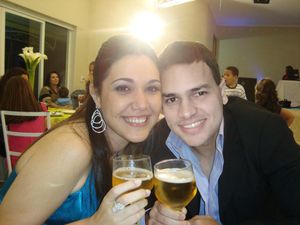} &
\timg{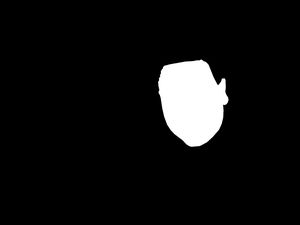} &
\timg{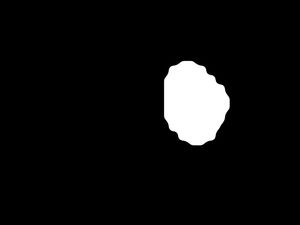} &
This image has been locally edited , forensic analysis detects manipulation within the image. &
Forensic analysis detects two distinct regions in this image, suggesting localized manipulation. Part of the image is consistent with a Sony A6000. The other area shows altered forensic characteristics, indicating local manipulation. The image has been JPEG compressed at quality 91. \\
\timg{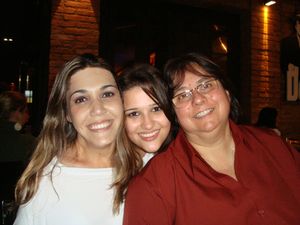} &
\timg{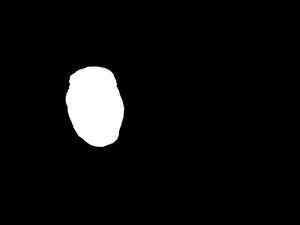} &
\timg{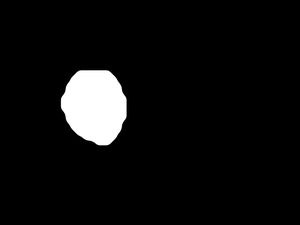} &
Image tampering detected: part of this image has been altered through traditional editing techniques. &
Forensic analysis detects two distinct regions in this image, suggesting localized manipulation. Both regions are attributed to a Sony A6000, but they exhibit different forensic characteristics, consistent with local editing or retouching. The image has been JPEG compressed at quality 90. \\
\rowcolor{gray!30}
\timg{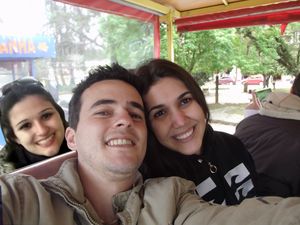} &
\timg{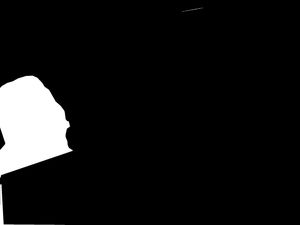} &
\timg{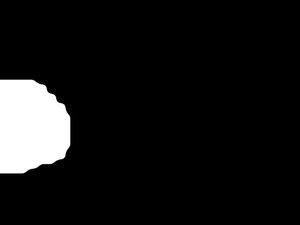} &
This image is a composite, combining photographs from two different camera sources. &
Forensic analysis identifies two distinct regions in this image. One region appears to come from a digital camera, while the other originates from a different camera. The presence of two different camera signatures within one image is strong evidence of splicing. This region also shows signs of being resized to approximately 109\% of its original resolution. The image has been JPEG compressed at quality 94. \\
\timg{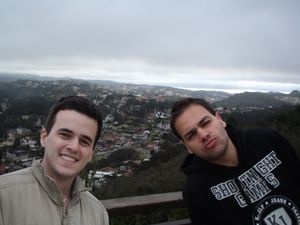} &
\timg{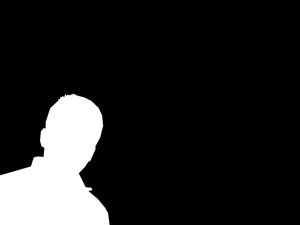} &
\timg{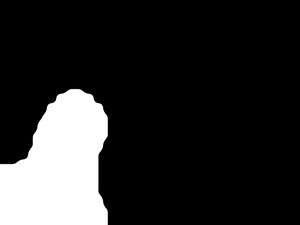} &
Splicing is present in this image , two forensically distinct camera sources were identified. &
Forensic analysis identifies two distinct regions in this image. One region appears to come from a digital camera, while the other originates from a different camera. The presence of two different camera signatures within one image is strong evidence of splicing. This region also shows signs of being resized to approximately 60\% of its original resolution. The image has been JPEG compressed at quality 89. \\
\bottomrule
\end{tabular}
\end{table}

\begin{table}[p]
\centering
\caption{Qualitative examples from \textbf{DSO-1} (external dataset). DSO-1 contains splicing forgeries created by compositing regions from different camera sources. The ground-truth mask marks the spliced region. (cont.)}
\smallerr
\setlength{\tabcolsep}{2pt}
\renewcommand{\arraystretch}{1.1}
\begin{tabular}{
	>{\centering\arraybackslash}m{17mm}
	>{\centering\arraybackslash}m{17mm}
	>{\centering\arraybackslash}m{17mm}
	>{\noindent\justifying\arraybackslash}m{20mm}
	>{\noindent\justifying\arraybackslash}m{48mm}
}
\toprule
\textbf{Image} & \textbf{GT Mask} & \textbf{Pred Mask} & \textbf{Decision} & \textbf{Justification} \\
\midrule
\rowcolor{gray!30}
\timg{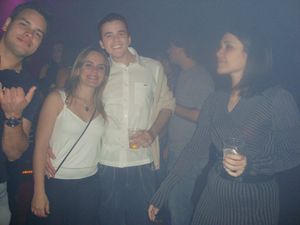} &
\timg{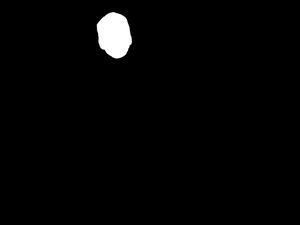} &
\timg{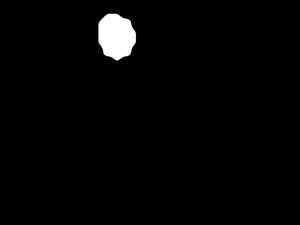} &
This image has been manipulated through splicing, merging content captured by different devices. &
Forensic analysis identifies two distinct regions in this image. One region appears to come from a digital camera, while the other traces back to a Canon EOS 600D. The presence of two different camera signatures within one image is strong evidence of splicing. The image has been JPEG compressed at quality 84. \\
\bottomrule
\end{tabular}
\end{table}

\begin{table}[p]
\centering
\caption{Qualitative examples from \textbf{CASIAv2} (external dataset). CASIAv2 contains splicing and copy-move forgeries. The ground-truth mask marks the manipulated region.}
\label{tab:qual_casiav2}
\smallerr
\setlength{\tabcolsep}{2pt}
\renewcommand{\arraystretch}{1.1}
\begin{tabular}{
	>{\centering\arraybackslash}m{17mm}
	>{\centering\arraybackslash}m{17mm}
	>{\centering\arraybackslash}m{17mm}
	>{\noindent\justifying\arraybackslash}m{20mm}
	>{\noindent\justifying\arraybackslash}m{48mm}
}
\toprule
\textbf{Image} & \textbf{GT Mask} & \textbf{Pred Mask} & \textbf{Decision} & \textbf{Justification} \\
\midrule
\rowcolor{gray!30}
\timg{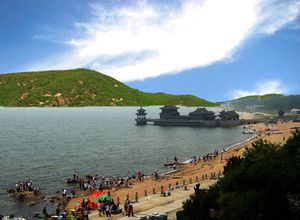} &
\timg{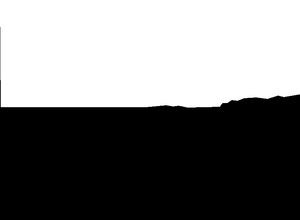} &
\timg{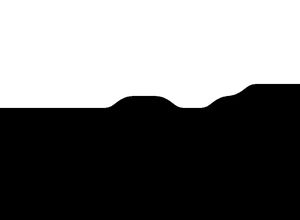} &
This image has been spliced , it contains content from two different cameras. &
Forensic analysis identifies two distinct regions in this image. One region is consistent with a Sony A6000, while the other originates from a different camera. The presence of two different camera signatures within one image is strong evidence of splicing. The image has been JPEG compressed at quality 92. \\
\timg{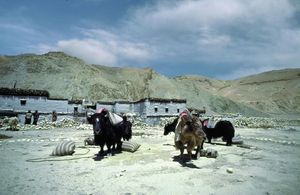} &
\timg{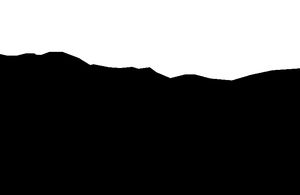} &
\timg{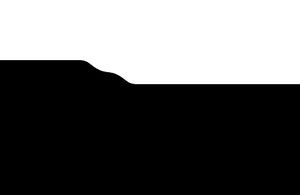} &
Image splicing detected: portions of this image originate from two separate cameras. &
Forensic analysis identifies two distinct regions in this image. One region is consistent with a DJI Mavic Pro, while the other traces back to a Nikon D3300. The presence of two different camera signatures within one image is strong evidence of splicing. The image has been JPEG compressed at quality 92. \\
\rowcolor{gray!30}
\timg{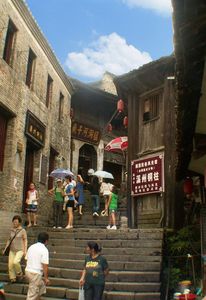} &
\timg{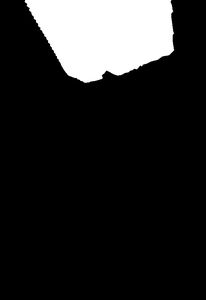} &
\timg{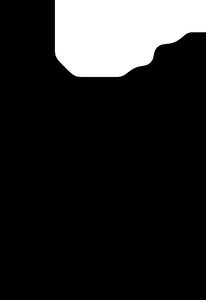} &
This image is a composite, combining photographs from two different camera sources. &
Forensic analysis identifies two distinct regions in this image. One region appears to come from a digital camera, while the other originates from a different camera. The presence of two different camera signatures within one image is strong evidence of splicing. This region also shows signs of being resized to approximately 129\% of its original resolution. The image has been JPEG compressed at quality 72. \\
\timg{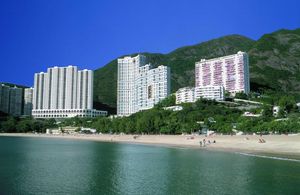} &
\timg{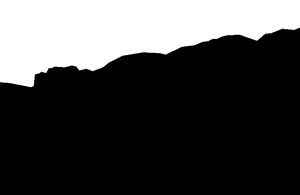} &
\timg{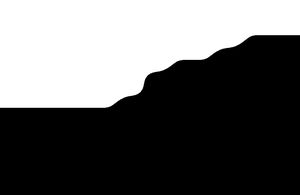} &
Splicing is present in this image , two forensically distinct camera sources were identified. &
Forensic analysis identifies two distinct regions in this image. One region is consistent with a Canon EOS Rebel T3i, while the other originates from a different camera. The presence of two different camera signatures within one image is strong evidence of splicing. The image has been JPEG compressed at quality 88. \\
\rowcolor{gray!30}
\timg{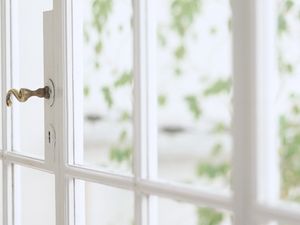} &
\timg{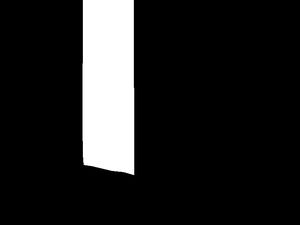} &
\timg{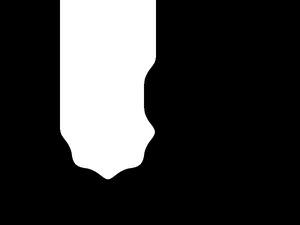} &
This image has been manipulated through splicing, merging content captured by different devices. &
Forensic analysis identifies two distinct regions in this image. One region is consistent with a Nikon D3400, while the other originates from a different camera. The presence of two different camera signatures within one image is strong evidence of splicing. This region also shows signs of being resized to approximately 75\% of its original resolution. The image has been JPEG compressed at quality 93. \\
\timg{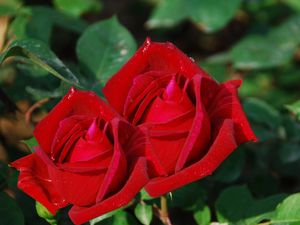} &
\timg{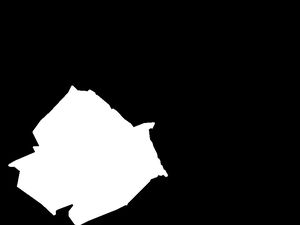} &
\timg{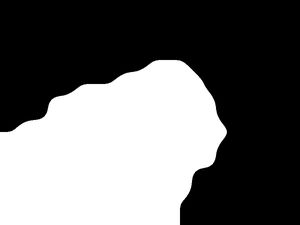} &
This image has been spliced , it contains content from two different cameras. &
Forensic analysis identifies two distinct regions in this image. One region is consistent with a Sony A6000, while the other traces back to a Nikon D7500. The presence of two different camera signatures within one image is strong evidence of splicing. The image has been JPEG compressed at quality 84. \\
\bottomrule
\end{tabular}
\end{table}

\begin{table}[p]
\centering
\caption{Qualitative examples from \textbf{CASIAv2} (external dataset). CASIAv2 contains splicing and copy-move forgeries. The ground-truth mask marks the manipulated region. (cont.)}
\smallerr
\setlength{\tabcolsep}{2pt}
\renewcommand{\arraystretch}{1.1}
\begin{tabular}{
	>{\centering\arraybackslash}m{17mm}
	>{\centering\arraybackslash}m{17mm}
	>{\centering\arraybackslash}m{17mm}
	>{\noindent\justifying\arraybackslash}m{20mm}
	>{\noindent\justifying\arraybackslash}m{48mm}
}
\toprule
\textbf{Image} & \textbf{GT Mask} & \textbf{Pred Mask} & \textbf{Decision} & \textbf{Justification} \\
\midrule
\rowcolor{gray!30}
\timg{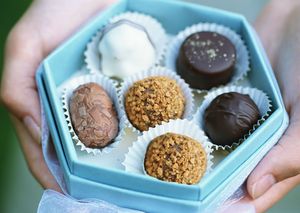} &
\timg{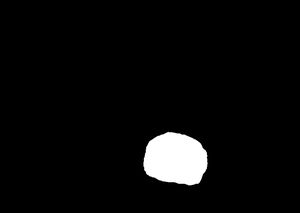} &
\timg{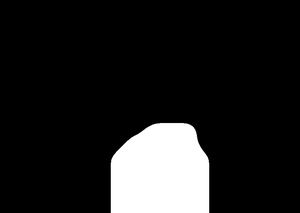} &
Image splicing detected: portions of this image originate from two separate cameras. &
Forensic analysis identifies two distinct regions in this image. One region appears to come from a digital camera, while the other originates from a different camera. The presence of two different camera signatures within one image is strong evidence of splicing. The image has been JPEG compressed at quality 90. \\
\timg{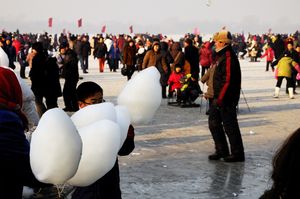} &
\timg{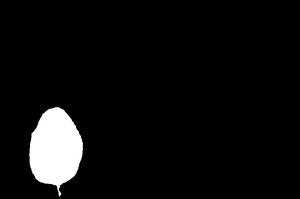} &
\timg{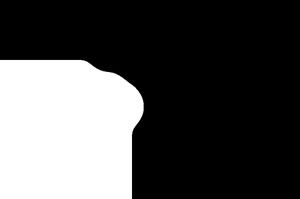} &
This image is a composite, combining photographs from two different camera sources. &
Forensic analysis identifies two distinct regions in this image. One region appears to come from a digital camera, while the other originates from a different camera. The presence of two different camera signatures within one image is strong evidence of splicing. The image has been JPEG compressed at quality 92. \\
\rowcolor{gray!30}
\timg{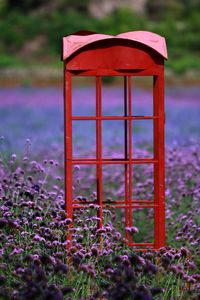} &
\timg{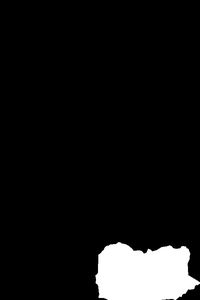} &
\timg{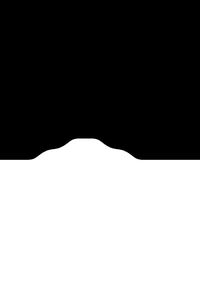} &
Splicing is present in this image , two forensically distinct camera sources were identified. &
Forensic analysis identifies two distinct regions in this image. One region appears to come from a digital camera, while the other originates from a different camera. The presence of two different camera signatures within one image is strong evidence of splicing. The image has been JPEG compressed at quality 84. \\
\timg{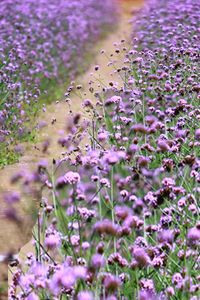} &
\timg{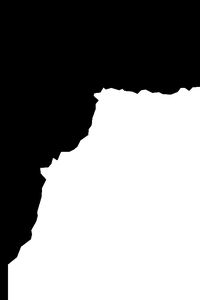} &
\timg{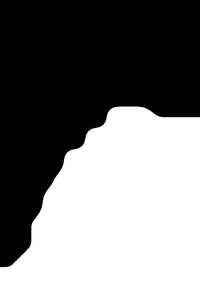} &
Forensic analysis reveals that this image blends authentic camera content with AI-generated regions. &
Forensic analysis reveals two distinct regions in this image. One region appears to be a real photograph from a digital camera. The other region shows characteristics of AI-generated content, attributed to GigaGAN with moderate confidence. The combination of real and AI-generated content indicates AI-based editing. The image has been JPEG compressed at quality 87. \\
\rowcolor{gray!30}
\timg{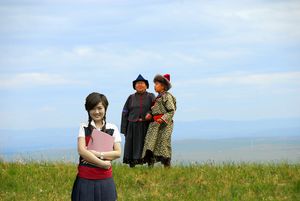} &
\timg{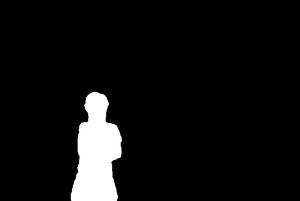} &
\timg{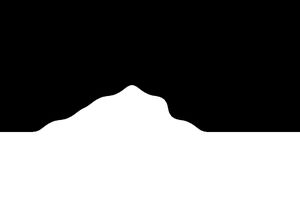} &
This image has been spliced , it contains content from two different cameras. &
Forensic analysis identifies two distinct regions in this image. One region is consistent with a Sony A6000, while the other originates from a different camera. The presence of two different camera signatures within one image is strong evidence of splicing. The image has been JPEG compressed at quality 91. \\
\timg{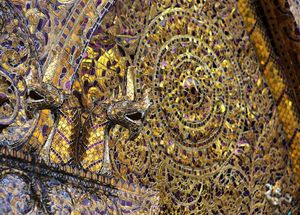} &
\timg{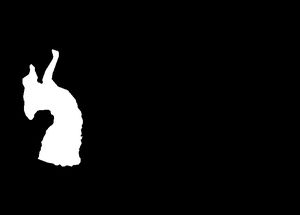} &
\timg{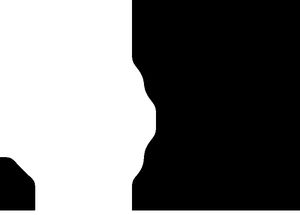} &
Image splicing detected: portions of this image originate from two separate cameras. &
Forensic analysis identifies two distinct regions in this image. One region is consistent with a Canon EOS Rebel T3i, while the other traces back to a Sony A6000. The presence of two different camera signatures within one image is strong evidence of splicing. The image has been JPEG compressed at quality 84. \\
\bottomrule
\end{tabular}
\end{table}

\begin{table}[p]
\centering
\caption{Qualitative examples from \textbf{CASIAv2} (external dataset). CASIAv2 contains splicing and copy-move forgeries. The ground-truth mask marks the manipulated region. (cont.)}
\smallerr
\setlength{\tabcolsep}{2pt}
\renewcommand{\arraystretch}{1.1}
\begin{tabular}{
	>{\centering\arraybackslash}m{17mm}
	>{\centering\arraybackslash}m{17mm}
	>{\centering\arraybackslash}m{17mm}
	>{\noindent\justifying\arraybackslash}m{20mm}
	>{\noindent\justifying\arraybackslash}m{48mm}
}
\toprule
\textbf{Image} & \textbf{GT Mask} & \textbf{Pred Mask} & \textbf{Decision} & \textbf{Justification} \\
\midrule
\rowcolor{gray!30}
\timg{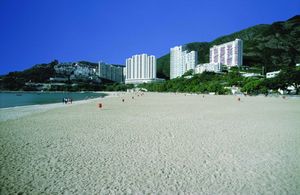} &
\timg{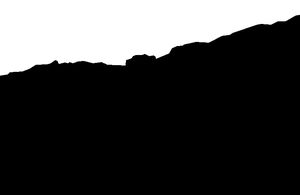} &
\timg{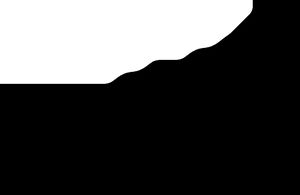} &
Forensic analysis reveals localized editing in this image, though both regions originate from the same type of camera. &
Forensic analysis detects two distinct regions in this image, suggesting localized manipulation. The dominant region is consistent with a Canon EOS Rebel T3i. The edited area shows altered forensic characteristics compared to the rest of the image. The image has been JPEG compressed at quality 89. \\
\timg{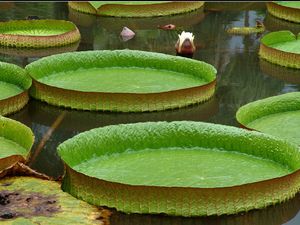} &
\timg{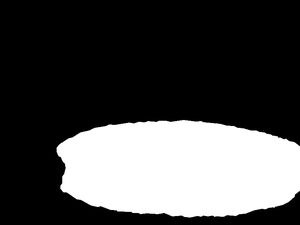} &
\timg{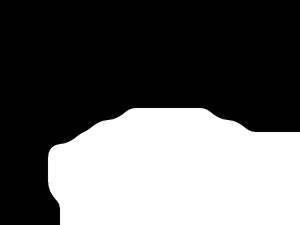} &
Part of this image has been altered with AI generation tools while the rest remains from the original photograph. &
Forensic analysis reveals two distinct regions in this image. One region is consistent with a real photograph taken by a DJI Mavic Pro. The other region shows characteristics of AI-generated content, attributed to Tam Trans with moderate confidence. The combination of real and AI-generated content indicates AI-based editing. The image has been JPEG compressed at quality 86. \\
\rowcolor{gray!30}
\timg{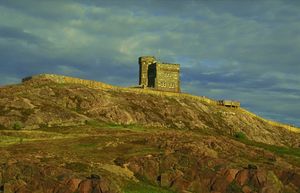} &
\timg{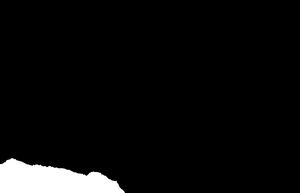} &
\timg{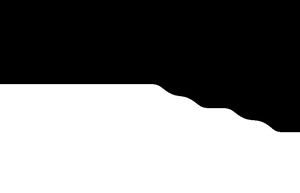} &
This image has been manipulated through splicing, merging content captured by different devices. &
Forensic analysis identifies two distinct regions in this image. One region is consistent with a Canon EOS 5D Mark II, while the other originates from a different camera. The presence of two different camera signatures within one image is strong evidence of splicing. The image has been JPEG compressed at quality 90. \\
\timg{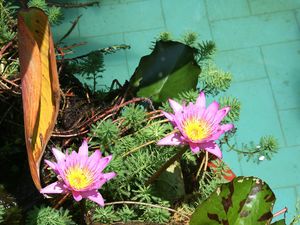} &
\timg{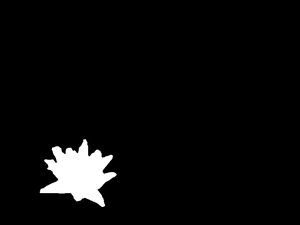} &
\timg{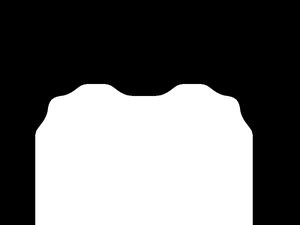} &
This image has been spliced , it contains content from two different cameras. &
Forensic analysis identifies two distinct regions in this image. One region is consistent with a Canon EOS Rebel T3i, while the other traces back to a Sony A6000. The presence of two different camera signatures within one image is strong evidence of splicing. The image has been JPEG compressed at quality 87. \\
\rowcolor{gray!30}
\timg{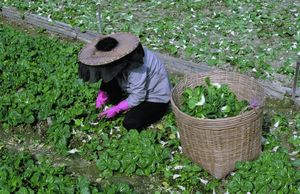} &
\timg{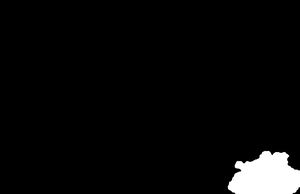} &
\timg{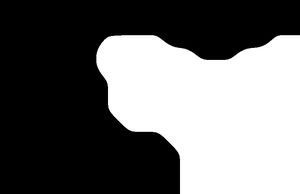} &
Image splicing detected: portions of this image originate from two separate cameras. &
Forensic analysis identifies two distinct regions in this image. One region is consistent with a Canon EOS Rebel T3i, while the other traces back to a Sony A6000. The presence of two different camera signatures within one image is strong evidence of splicing. The image has been JPEG compressed at quality 88. \\
\timg{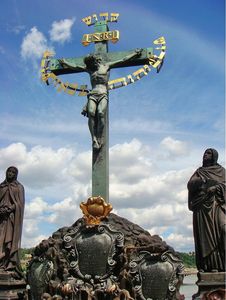} &
\timg{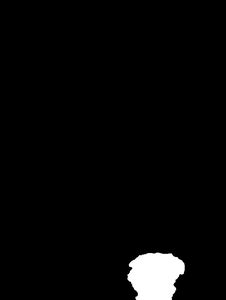} &
\timg{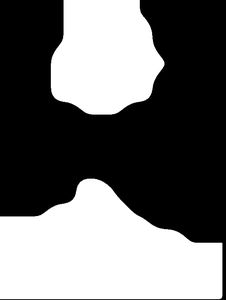} &
This image is a composite, combining photographs from two different camera sources. &
Forensic analysis identifies two distinct regions in this image. One region is consistent with a Sony A6000, while the other traces back to a Canon EOS 5D Mark II. The presence of two different camera signatures within one image is strong evidence of splicing. The image has been JPEG compressed at quality 91. \\
\bottomrule
\end{tabular}
\end{table}

\begin{table}[p]
\centering
\caption{Qualitative examples from \textbf{CASIAv2} (external dataset). CASIAv2 contains splicing and copy-move forgeries. The ground-truth mask marks the manipulated region. (cont.)}
\smallerr
\setlength{\tabcolsep}{2pt}
\renewcommand{\arraystretch}{1.1}
\begin{tabular}{
	>{\centering\arraybackslash}m{17mm}
	>{\centering\arraybackslash}m{17mm}
	>{\centering\arraybackslash}m{17mm}
	>{\noindent\justifying\arraybackslash}m{20mm}
	>{\noindent\justifying\arraybackslash}m{48mm}
}
\toprule
\textbf{Image} & \textbf{GT Mask} & \textbf{Pred Mask} & \textbf{Decision} & \textbf{Justification} \\
\midrule
\rowcolor{gray!30}
\timg{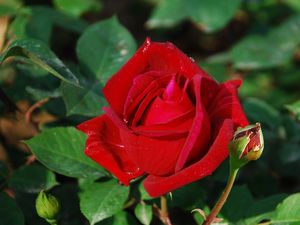} &
\timg{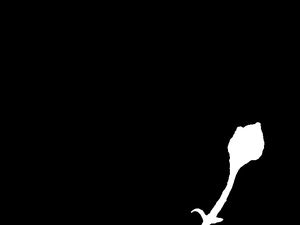} &
\timg{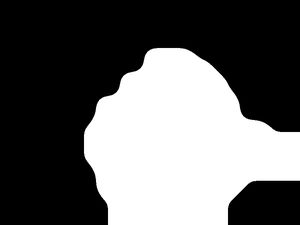} &
Splicing is present in this image , two forensically distinct camera sources were identified. &
Forensic analysis identifies two distinct regions in this image. One region is consistent with a Sony A6000, while the other originates from a different camera. The presence of two different camera signatures within one image is strong evidence of splicing. The image has been JPEG compressed at quality 88. \\
\timg{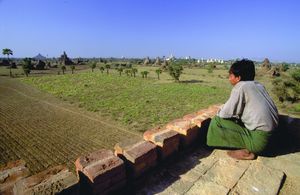} &
\timg{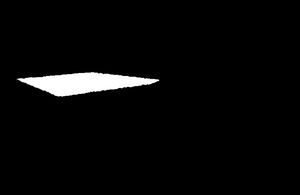} &
\timg{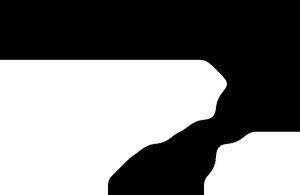} &
This image has been manipulated through splicing, merging content captured by different devices. &
Forensic analysis identifies two distinct regions in this image. One region appears to come from a digital camera, while the other traces back to a DJI Mavic Pro. The presence of two different camera signatures within one image is strong evidence of splicing. The image has been JPEG compressed at quality 89. \\
\bottomrule
\end{tabular}
\end{table}

\end{document}